\PassOptionsToPackage{table,xcdraw}{xcolor}

\documentclass[sigconf, nonacm]{acmart}

\AtBeginDocument{
  }

\setcopyright{none}
\copyrightyear{2027}
\acmYear{2027}
\acmDOI{XXXXXXX.XXXXXXX}
\acmConference[KDD '27]{Proceedings of the 33rd ACM SIGKDD Conference on Knowledge Discovery and Data Mining}{August 2027}{Washington, DC, USA}
\acmISBN{978-1-4503-XXXX-X/2027/08}

\usepackage{graphicx}
\usepackage{subcaption}
\usepackage{multirow}
\usepackage{colortbl}
\usepackage{placeins}
\usepackage{algorithm}
\usepackage{algpseudocode}
\usepackage{mathtools}
\usepackage{amsthm}
\usepackage{wrapfig}
\usepackage{enumitem}

\usepackage[capitalize,noabbrev]{cleveref}

\theoremstyle{plain}

\theoremstyle{definition}

\theoremstyle{remark}

\usepackage[textsize=tiny]{todonotes}

\begin{document}

\title{MOT-SR: Multi-Objective Tool-Augmented Scientific Equation Discovery with Large Language Models}

\settopmatter{authorsperrow=4}

\author{Boxiao Wang}
\orcid{0009-0008-2970-7575}
\affiliation{\department{National Key Laboratory of Cognition and Decision Intelligence for Complex Systems}
  \institution{Institute of Automation, Chinese Academy of Sciences}
  \city{Beijing}
  \country{China}}
\email{wangboxiao2026@ia.ac.cn}

\author{Runxiang Wang}
\affiliation{\institution{University of the Chinese Academy of Sciences
School of Advanced Interdisciplinary Sciences}
  \city{Beijing}
  \country{China}}
\email{wangrunxiang2023@ia.ac.cn}

\author{Kai Li}
\orcid{0000-0003-3840-3270}
\affiliation{\department{National Key Laboratory of Cognition and Decision Intelligence for Complex Systems}
  \institution{Institute of Automation, Chinese Academy of Sciences}
  \city{Beijing}
  \country{China}}
\email{kai.li@ia.ac.cn}

\author{Chongming Li}
\affiliation{\institution{School of Astronomy and Space Science, University of the Chinese Academy of Sciences}
  \city{Beijing}
  \country{China}}
\email{lichongming24@mails.ucas.ac.cn}

\author{Zhiwei Chen}
\affiliation{\department{Relativistic Astrophysics, Institute for Theoretical Physics}
  \institution{Goethe University Frankfurt}
  \city{Frankfurt am Main}
  \country{Germany}}
\email{zchen@itp.uni-frankfurt.de}

\author{Yifan Zhang}
\orcid{0000-0002-9190-3509}
\affiliation{\department{National Key Laboratory of Cognition and Decision Intelligence for Complex Systems}
  \institution{Institute of Automation, Chinese Academy of Sciences}
  \city{Beijing}
  \country{China}}
\email{yfzhang@nlpr.ia.ac.cn}

\author{Jian Cheng}
\orcid{0000-0003-1289-2758}
\affiliation{\department{National Lab of Pattern Recognition}
  \institution{Institute of Automation, Chinese Academy of Sciences}
  \city{Beijing}
  \country{China}}
\email{jcheng@nlpr.ia.ac.cn}

\renewcommand{\shortauthors}{Wang et al.}

\begin{abstract}
Discovering compact and interpretable equations from observational data is essential for understanding complex scientific systems. Symbolic regression (SR) provides a general computational framework for this task. While recent Large Language Model (LLM) based SR approaches show promise, they face two key limitations. First, they lack dedicated data analysis mechanisms for uncovering variable dependencies, which reduces the efficiency of equation discovery. Second, most methods rely on single-objective evaluation focused solely on fitting error. This neglect of structural complexity and generalization often causes models to converge prematurely to local optima, limiting their ability to explore the broader equation space. We propose Multi-Objective Tool-augmented Symbolic Regression (MOT-SR\footnote{Code is available at \url{https://github.com/wswbx/MOT-SR}.}), a unified framework that integrates external analytical tools
to extract structural priors and guide equation generation, while jointly optimizing for accuracy, complexity, and generalization via a multi-objective evaluation module that maintains a dynamic Pareto front. MOT-SR employs two collaborative LLM modules: a Meta Strategy Generator, which selects tools and synthesizes structural optimization strategies based on Pareto-optimal equations, and an Equation Generator, which produces new candidate equations accordingly. The system operates in a closed-loop manner, continuously refining both strategies and equation structures. Across 40 standard tasks, MOT-SR outperforms existing SR methods in accuracy, generalization, and efficiency. We further validate MOT-SR on extreme mass-ratio inspiral (EMRI) orbital modeling, an important problem in space-based gravitational-wave astronomy where small local errors can accumulate substantially over long-term evolution. The discovered interpretable correction achieves the lowest trajectory-level integration error on held-out configurations. These results demonstrate the potential of MOT-SR to enable reliable modeling of long-horizon scientific dynamics.

\end{abstract}


\maketitle

\section{Introduction}

Symbolic Regression (SR)~\citep{article} aims to discover underlying mathematical equations from data and has long been recognized as a key methodology in scientific discovery. It has been widely applied across disciplines, from identifying physical laws~\citep{10.1093/pnasnexus/pgae467,10.1007/978-3-031-29573-7_3} and modeling chemical systems~\citep{ICLR2025_a76b693f,DENG2023109010}, to analyzing  dynamic processes in biological or economic systems~\citep{Wahlquist2024,shi2024alphaforgeframeworkdynamicallycombine}. By generating compact and interpretable equations, SR enables both accurate prediction and deep insight into system behavior.

SR has long been recognized as an NP-hard problem~\citep{virgolin2022symbolicregressionnphard}, motivating diverse algorithmic developments. Early approaches based on genetic programming~\citep{doi:10.1126/science.1165893,cranmer2023interpretablemachinelearningscience} evolve equations via mutation and crossover. Reinforcement learning~\citep{petersen2021deepsymbolicregressionrecovering} models SR as a sequential decision-making process. Recently, Transformer-based models have enabled end-to-end learning from data to equations~\citep{biggio2021neuralsymbolicregressionscales,kamienny2022endtoendsymbolicregressiontransformers,zhang2025ragsr}. With the rise of large language models (LLMs), methods such as LLM-SR~\citep{shojaee2025llmsrscientificequationdiscovery} and LaSR~\citep{grayeli2024symbolicregressionlearnedconcept} leverage LLM's in-context learning capabilities and scientific priors to perform symbolic reasoning and equation generation.

Despite encouraging progress, existing LLM-based SR methods face two key limitations. First, they typically lack systematic analysis of variable dependencies and data distributions, relying instead on problem descriptions as context. This often results in poorly constrained search spaces, which limits both the efficiency and directionality of equation exploration. Second, most approaches adopt a single-objective evaluation, typically minimizing fitting error, while overlooking other critical factors such as equation complexity, generalization, and diversity. This can lead to overfitting and premature convergence to locally optimal solutions. 

\begin{figure}[t]
  \centering
\includegraphics[width=1\linewidth]{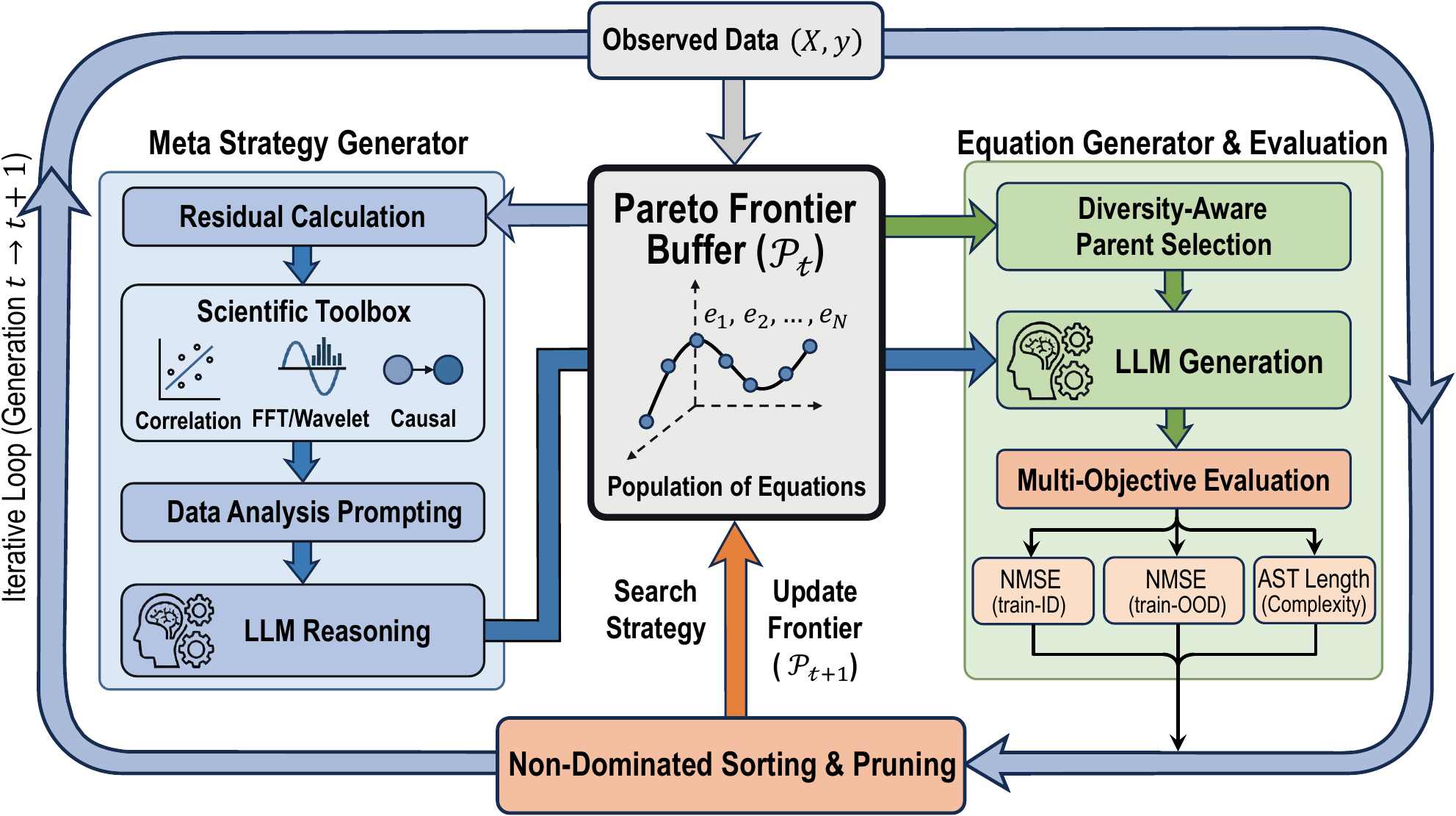}
\caption{\textbf{Overview of MOT-SR.} Given observed data $(X, y)$, MOT-SR maintains a Pareto-frontier buffer $\mathcal{P}_t$ of candidate equations and multi-objective scores. At iteration $t$, a Meta Strategy Generator analyzes residuals of Pareto-optimal candidates, applies scientific tools, and outputs search strategy. Following this search strategy, an Equation Generator \& Evaluation module performs diversity-aware parent selection, generates new equations, and evaluates them by $\mathrm{NMSE}$ on $D_{\mathrm{train}}^{\mathrm{ID}}$ and $D_{\mathrm{train}}^{\mathrm{OOD}}$ (training-derived in-domain (ID)/out-of-domain (OOD) split) and by AST length for complexity. Non-dominated sorting and pruning update the frontier to $\mathcal{P}_{t+1}$, forming a closed-loop scientific equation discovery process.
}

  \label{fig:main}
\end{figure}

To address these challenges, we propose Multi-Objective Tool-augmented Symbolic Regression (MOT-SR), a unified framework (Figure~\ref{fig:main}) inspired by the human scientific modeling process. Scientists typically begin by analyzing data using a variety of tools, evaluate candidate hypotheses from multiple perspectives, and iteratively refine their modeling direction accordingly. MOT-SR operationalizes this human-like modeling paradigm through two core mechanisms and a cooperative, LLM-driven evolution system.

At its core, MOT-SR first employs a tool-augmented analysis mechanism that invokes a suite of analytical tools to extract variable relationships from multiple complementary dimensions, such as linearity, periodicity, and causality. These insights are converted into interpretable priors and structural constraints that guide equation generation. In parallel, MOT-SR adopts a multi-objective evaluation mechanism that jointly assesses candidate equations across three dimensions: accuracy on the in-domain (ID) training subset $D_{\mathrm{train}}^{\mathrm{ID}}$, generalization on the out-of-domain (OOD) training subset $D_{\mathrm{train}}^{\mathrm{OOD}}$,
and structural complexity measured via abstract syntax tree (AST) length. A Pareto front is maintained using non-dominated sorting to preserve high-quality equations that represent optimal trade-offs.

To realize these mechanisms in an adaptive and iterative manner, MOT-SR incorporates two cooperating LLMs. Specifically, the Meta Strategy Generator analyzes residual patterns of the current Pareto-optimal candidates and leverages external scientific tools (e.g., correlation, FFT/wavelet, and causal discovery) to produces a data-driven search strategy. Guided by this strategy, the Equation Generator \& Evaluation module performs diversity-aware parent selection and generates new candidate equations, which are evaluated and integrated into the Pareto-frontier buffer via non-dominated sorting and pruning.

We evaluate MOT-SR on five benchmarks spanning physics, chemistry, biology, and materials science, using both the open-source LLaMA-3.1~\citep{kassianik2025llama31foundationaisecurityllmbase8btechnicalreport} and the commercial GPT-4o mini~\citep{openai_gpt4o_mini2024}. Across all tasks, MOT-SR consistently outperforms traditional SR methods and recent LLM-based baselines in terms of accuracy, generalization, and equation compactness.

Beyond benchmark evaluation, we apply MOT-SR to extreme mass-ratio inspiral (EMRI) orbital evolution, a challenging problem in space-based gravitational-wave astronomy where small local errors accumulate over many orbital cycles. Using 58 configurations for equation discovery and 30 held-out configurations for a posteriori evaluation, MOT-SR identifies a compact symbolic correction that achieves the lowest trajectory-level integration error—approximately three orders of magnitude lower than a neural residual baseline and 26.8 times lower than LLM-SR. These results demonstrate that MOT-SR can recover transferable correction structures that remain reliable across unseen physical configurations and long horizons, highlighting the potential of interpretable equation discovery for demanding scientific dynamical systems.

\section{Preliminaries}

In SR, the learning task typically starts with a dataset consisting of input-output pairs:
\[
D = \{ (\mathbf{x}_i, y_i) \}_{i=1}^n, \quad \mathbf{x}_i \in \mathbb{R}^d, \quad y_i \in \mathbb{R},
\]
where \( \mathbf{x}_i \) denotes a $d$-dimensional input vector and \( y_i \) is the corresponding scalar output. The goal is to discover an explicit analytical equation \( f(\cdot) \) such that the predicted outputs \( \hat{y}_i = f(\mathbf{x}_i) \) closely approximate the ground truth \( y_i \). To assess the quality of a candidate equation, the normalized mean squared error (NMSE) is defined as
\begin{equation}
    \mathrm{NMSE}(f,D)
    =
    \frac{1}{n}
    \sum_{i=1}^{n}
    \left(
        \frac{f(\mathbf{x}_i)-y_i}{\sigma_y}
    \right)^2.
    \label{eq:nmse}
\end{equation}
where \( \sigma_y \) is the standard deviation of the target values across the dataset \( D \). This metric reflects the equation’s predictive accuracy, normalized by the variance of the outputs. Beyond fitting accuracy, SR also values simplicity and generalization, seeking equations that are not only accurate but also compact and transferable to unseen domains.

Our work builds on LLM-SR, a framework that leverages LLMs to generate symbolic equations through iterative optimization. Its core pipeline includes:

\begin{itemize}
\item Equation skeleton generation: Structured prompts contain task-specific information (e.g., variable names, optimization goals, example equations), guiding the LLM to produce physically plausible equation skeletons.

\item Parameter optimization: The skeletons’ free parameters are optimized (e.g., via BFGS~\citep{fletcher1987_practical_methods} ) and scored using NMSE.

\item Feedback: High-quality equations are retained and reused as in-context examples, enabling iterative refinement through feedback-driven generation.
\end{itemize}

While promising, LLM-based SR methods face key limitations: they lack systematic modeling of variable dependencies, leading to structurally under-informed equations; and they rely on a single-objective evaluation focused solely on fitting error, neglecting factors such as complexity and generalization. These shortcomings diminish their effectiveness in solving more challenging tasks.

\section{Method}

To address the above limitations, we propose \textbf{MOT-SR} (Multi-Objective Tool-augmented Symbolic Regression), a unified framework that integrates external data analysis tools, multi-objective evaluation, and cooperative LLMs to enhance equation quality and search efficiency. MOT-SR first extracts structural priors by analyzing variable relationships with diverse analytical tools. It then introduces a multi-objective evaluation mechanism that jointly considers fitting error, equation complexity, and generalization, dynamically maintaining a Pareto front through non-dominated sorting. Finally, two complementary LLMs work in tandem: one generates structural refinement strategies, while the other synthesizes candidate equations accordingly, forming a closed-loop process that continuously guides and improves equation discovery.

\subsection{Tool-Augmented Variable Analysis}

This component constructs structural priors by quantifying diverse variable relationships using a suite of carefully designed analytical tools. Details are provided in Appendix~\ref{appendix:Tool Set}.

\textbf{Linear correlation tools} are essential for identifying dominant variables and constructing interpretable regression structures. MOT-SR integrates several complementary methods to assess linear dependencies from multiple statistical angles. The \textit{Pearson correlation coefficient}~\citep{a0dc553c-0830-3fe2-ab2a-eece0d66a7db} quantifies pairwise linear associations, especially effective for Gaussian-like data. \textit{Simple linear regression} and \textit{residual variance analysis}~\citep{Montgomery2013} evaluate predictive capacity and error stability. \textit{PCA-based explained variance}~\citep{jolliffe2002_principal_component_analysis} identifies the key directions of structural variance. These tools jointly establish a solid basis for linear trend detection. 

\textbf{Nonlinear dependency tools} are employed to capture complex interactions essential for modeling nonlinear systems. MOT-SR integrates three complementary methods: the \textit{Spearman rank correlation}~\citep{ca468a70-0be4-389a-b0b9-5dd1ff52b33f} measures monotonic associations based on rank, offering robustness to noise and non-Gaussian data; \textit{mutual information}~\citep{6773024} measures the overall statistical dependency between variables without assuming any parametric form; and \textit{mutual information regression }~\citep{pedregosa2011_scikitlearn} quantifies the marginal contribution of each variable conditioned on others. Together, these tools guide whether nonlinear or higher-order terms should be introduced. 

\textbf{Time-frequency analysis tools} help detect periodicity and transient dynamics, which frequently occur in oscillatory and multiscale systems. MOT-SR employs two complementary methods: \textit{Fast Fourier Transform }~\citep{Cooley:1965zz} identifies dominant global frequency components, while the \textit{wavelet transform energy spectrum}~\citep{daubechies1992_ten_lectures} captures localized, non-stationary fluctuations. These insights support the inclusion of periodic terms (e.g., $\sin$) in candidate equations. 

\textbf{Causal inference tools} are used to identify whether one variable may influence another in a predictive or explanatory sense. MOT-SR adopts \textit{Granger causality}~\citep{702ab909-8cb1-3c30-a5f1-ab4517d6cf1c} for detecting temporal causal influence in linear time-series, and \textit{Convergent Cross Mapping }~\citep{doi:10.1126/science.1227079} for identifying latent causality in nonlinear systems with potential delays. These methods provide structural signals that enhance the interpretability and explanatory power of generated equations. 

\textbf{Dynamic complexity tools} help assess intrinsic system richness and redundancy, guiding the pruning of over-specified components in equations. MOT-SR employs three methods: the \textit{Lyapunov exponent}~\citep{WOLF1985285}, which measures sensitivity to initial conditions and indicates chaotic behavior; the \textit{correlation dimension}~\citep{GRASSBERGER1983189}, which estimates the system’s effective degrees of freedom; and \textit{Dynamic Time Warping (DTW)}~\citep{1163055}, which evaluates time-shifted similarity between variable trajectories. Together, they offer structural cues for constructing compact and robust equations. 

\textbf{Distribution consistency tools} evaluate whether input variables behave uniformly across different input regions. MOT-SR uses the \textit{Kolmogorov–Smirnov (KS) test}~\citep{16e7f618-c06b-3d10-8705-1086b218d827} to detect distributional shifts between subdomains, informing the use of region-dependent structures to reflect local variations in the data distribution. 

Rather than introducing new tools, the key innovation of \textsc{MOT-SR} lies in enabling LLMs to autonomously invoke and coordinate these tools to extract dependency patterns among variables. These insights are distilled into concise guidance that informs variable selection and function composition, thereby improving responsiveness to data characteristics and enhancing the scientific plausibility of generated equations.
By adaptively combining outputs from heterogeneous analyses, for instance by linking correlation measures with periodicity detection, MOT-SR supports more targeted equation discovery.
The impact of this tool-augmented analysis on equation generation is substantiated by the case studies in Appendix~\ref{tool-case-study}.
Looking ahead, \textsc{MOT-SR} opens the possibility for LLMs to synthesize new tools, further expanding the scope of SR research.

\subsection{Multi-Objective Evaluation}  
To enhance search efficiency and model quality, MOT-SR adopts a multi-objective evaluation scheme that jointly considers predictive accuracy, generalization, and structural simplicity.
A Pareto-based selection strategy maintains a diverse set of non-dominated candidate equations, 
improving robustness and exploration of the solution space. We define the ID/OOD regions \textit{within} the training set $D_{\mathrm{train}}$ as a spatial split (Figure~\ref{fig:id_ood_2x2}). The detailed configuration of the ID-OOD partition for $D_\mathrm{train}$ is provided in Appendix~\ref{appendix:id-ood}.

\subsubsection*{Evaluation Metrics}

\begin{figure}[t]
  \centering

  \begin{subfigure}[t]{0.46\linewidth}
    \includegraphics[width=\linewidth]{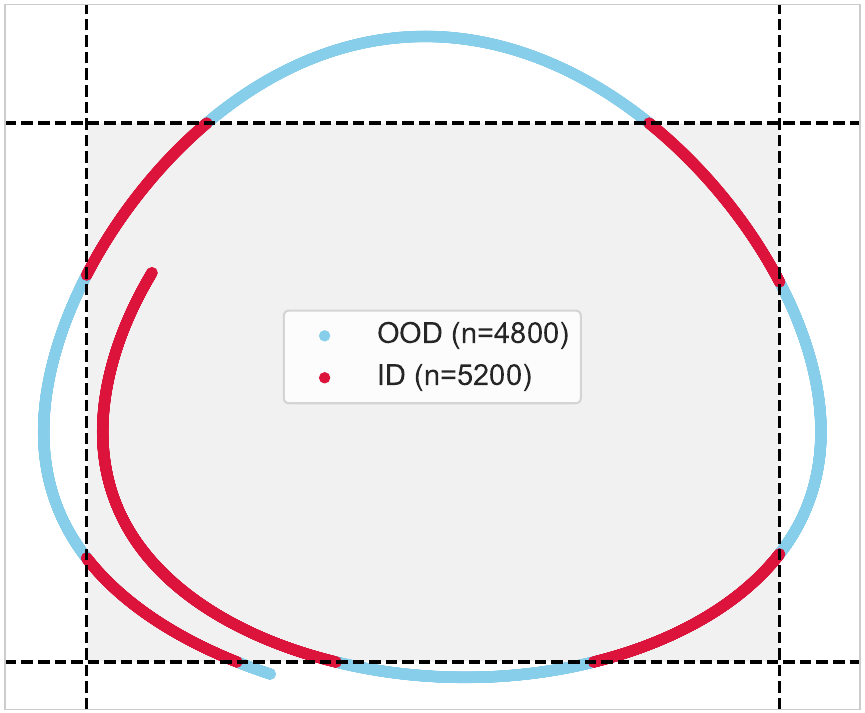}
  \end{subfigure}
  \hspace{0.025\textwidth} 
  \begin{subfigure}[t]{0.46\linewidth}
    \includegraphics[width=\linewidth]{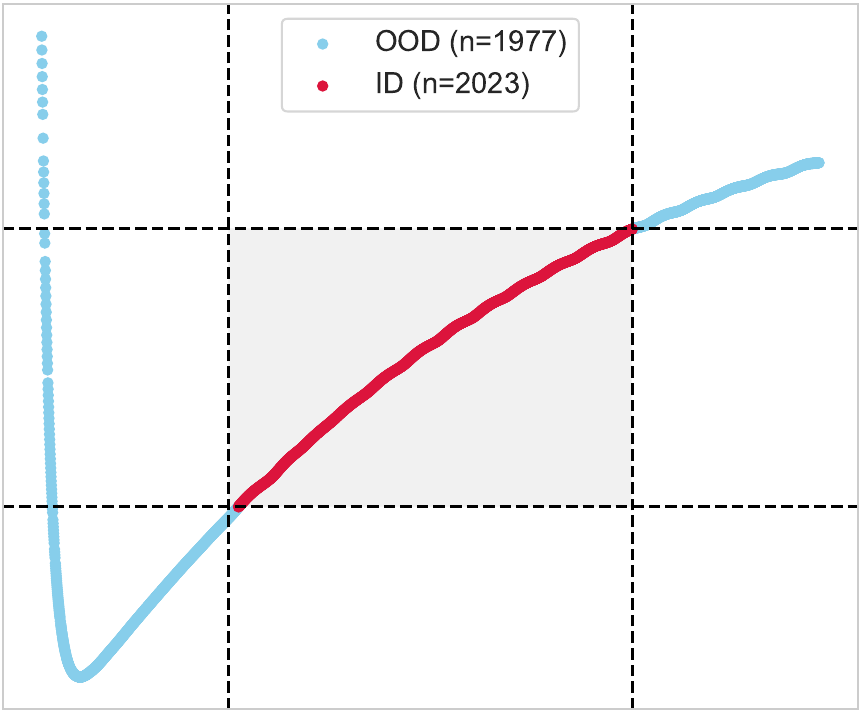}
  \end{subfigure}
\caption{Illustration of the training-derived ID/OOD split used in MOT-SR. The shaded region denotes $D_{\mathrm{train}}^{\mathrm{ID}}$ defined by per-dimension percentile bounds, and the white region denotes $D_{\mathrm{train}}^{\mathrm{OOD}}=D_{\mathrm{train}}\setminus D_{\mathrm{train}}^{\mathrm{ID}}$, shown on two representative tasks.}

  \label{fig:id_ood_2x2}
  \vspace{-1em}
\end{figure}

Unlike prior LLM-based SR methods that optimize only fitting error, \textsc{MOT-SR} employs a three-fold metric:

\textbf{(1) In-Domain (ID) Accuracy.}  
Measures interpolation performance within the central region of the input space. 
The in-domain subset $D_{\mathrm{train}}^{\mathrm{ID}}$ is selected from the middle intervals along each input dimension (Figure~\ref{fig:id_ood_2x2}). 
Accuracy is quantified using NMSE on $D_{\mathrm{train}}^{\mathrm{ID}}$, denoted as $\mathrm{NMSE}_{\mathrm{ID}}$.

\textbf{(2) Out-of-Domain (OOD) Generalization.}  
Measures extrapolation performance in the peripheral regions 
$D_{\mathrm{train}}^{\mathrm{OOD}} = D_{\mathrm{train}} \setminus D_{\mathrm{train}}^{\mathrm{ID}}$, which correspond to the white regions in Figure~\ref{fig:id_ood_2x2}. 
NMSE on this subset is denoted as $\mathrm{NMSE}_{\mathrm{OOD}}$ and reflects robustness to distributional shifts, a key requirement for scientific applications.
It is worth noting that $\mathrm{NMSE}_{\mathrm{OOD}}$ is calculated on the training-derived region $D_{\mathrm{train}}^{\mathrm{OOD}}$ and does not involve any held-out test data.

\textbf{(3) Equation Complexity.}
To promote interpretability, equation complexity is measured via the size of its abstract syntax tree (AST)~\citep{10.1145/1083142.1083143}. In \textsc{MOT-SR}, each equation is represented as a Python function and parsed into an AST, where each node corresponds to a variable, operator, or function call. The total node count provides a low-overhead estimate of structural complexity, guiding the search toward compact equations. Although Python AST length may disproportionately penalize composite function calls, we adopt it as a computationally inexpensive proxy for expression-tree complexity; its empirical ranking consistency and associated trade-offs are discussed in Appendix~\ref{app:ast_complexity}.

\subsubsection*{Pareto-Guided Multi-Objective Optimization}

Based on the three metrics, MOT-SR formulates equation discovery as a multi-objective optimization problem over a candidate set \( \mathcal{P} = \{f_1, f_2, \dots, f_M\} \), aiming to balance fitting accuracy, extrapolation, and structural simplicity:
\[
\resizebox{\columnwidth}{!}{$
f^*=\arg\min_{f\in\mathcal{P}}\big(\mathrm{NMSE}(f,D_{\mathrm{train}}^{\mathrm{ID}}),\ \mathrm{NMSE}(f,D_{\mathrm{train}}^{\mathrm{OOD}}),\ \mathrm{ASTLen}(f)\big)
$}
\]

A function \( f_a \) dominates \( f_b \) if it performs no worse across all objectives and better on at least one. The Pareto front \( \mathcal{P}^* \) consists of all non-dominated candidates in \( \mathcal{P} \).
To improve the quality of the front, MOT-SR filters out overly simple candidates with low equation complexity that may mislead the search by applying upper-bound thresholds on $\mathrm{NMSE}(f, D_{\mathrm{train}}^{\mathrm{ID}})$ and $\mathrm{NMSE}(f, D_{\mathrm{train}}^{\mathrm{OOD}})$. Remaining equations are ranked via non-dominated sorting to construct the current Pareto-optimal set. Details of the dynamic threshold construction are provided in Appendix~\ref{app:thresholding}.

\subsection{Cooperative LLMs for Equation Evolution}

To integrate variable-aware analysis with multi-objective optimization, \textsc{MOT-SR} employs a cooperative framework consisting of two LLMs: the \textit{Meta Strategy Generator} \( \pi_{\text{stg}} \) and the \textit{Equation Generator} \( \pi_{\text{eq}} \). The former extracts structural search strategies, while the latter generates candidate equations accordingly, jointly driving the population toward Pareto-optimality (Algorithm~\ref{alg:main}).

\paragraph{Meta Strategy Generator.}
$\pi_{\text{stg}}$ formulates equation search strategies by synthesizing variable-level insights and structural abstraction. At each iteration, given the current Pareto front \( \mathcal{P}_{t-1}^* \), \( \pi_{\text{stg}} \) autonomously selects a subset of tools \( \mathcal{A} \subseteq \texttt{Toolbox} \), applies them to the dataset $D_{\mathrm{train}}$, and summarizes key relationships as natural language descriptions \( \mathcal{R}_{\text{varrel}} \) to guide downstream generation.

In parallel, \( \pi_{\text{stg}} \) performs structural abstraction over \( \mathcal{P}_{t-1}^* \), producing complementary guidance \( \mathcal{R}_{\text{struct}} \) through: 1) \textit{Commonality Extraction}, which identifies frequent substructures that define prevailing symbolic motifs; 2) \textit{Disparity Analysis}, which diagnoses regional residual patterns to uncover structural weaknesses; and 3) \textit{Blind Spot Discovery}, which detects unexplored symbolic components to encourage diversity.
The final strategy \( \mathcal{S}_t = (\mathcal{R}_{\text{varrel}}, \mathcal{R}_{\text{struct}}) \) is passed to equation generator \( \pi_{\text{eq}} \) to steer the next round of generation.

\paragraph{Equation Generator.}

Upon receiving the strategy \( \mathcal{S}_t \) from $\pi_{\text{stg}}$, \( \pi_{\text{eq}} \) is responsible for synthesizing a new batch of candidate equations \( \mathcal{P}_t \). To promote structural diversity and prevent premature convergence, \textsc{MOT-SR} incorporates a \textit{structure-diversity–guided parent sampling module}, which explicitly prioritizes structurally distinct candidates during equation generation.

Specifically, each equation \( f \in \mathcal{P}_t^* \) on the Pareto front is parsed into an abstract syntax tree (AST), from which a set of symbolic subtrees \( S(f) \) is extracted. The pairwise structural dissimilarity between two equations is computed via a subtree-overlap metric defined as:
\[
\text{SyntaxDiv}(f_i, f_j) = -\frac{|S(f_i)\cap S(f_j)|}{|S(f_i)|}.
\]
This score reflects the relative uniqueness of \( f_i \)'s structure compared to \( f_j \). For each equation, we compute the average structural diversity score:
\begin{center}
\begin{math}
\text{Score}_{\text{div}}(f_i) = \frac{1}{N-1} \sum_{j \ne i} \text{SyntaxDiv}(f_i, f_j).
\end{math}
\end{center}
A softmax sampling procedure is then applied to construct a parent set \( \mathcal{P}_{\text{parent}} \), biased toward structurally diverse equations. These parents serve as in-context examples, which, together with the strategy \( \mathcal{S}_t \), are fed into \( \pi_{\text{eq}} \) to generate new equation candidates: $f' \sim \pi_{\text{eq}}(\mathcal{S}_t, \mathcal{P}_{\text{parent}})$.
This procedure facilitates the generation of structurally diverse and generalizable expressions, driving the continuous evolution and expansion of the equation population \( \mathcal{P} \).

\begin{table*}[!t]

    \centering
    \resizebox{\textwidth}{!}{

    \begin{tabular}{lcccccccc}
    \hline
    \hline
    \multirow{2}{*}{\textbf{Model}} & \multicolumn{2}{c}{\textbf{Oscillation 1}} & \multicolumn{2}{c}{\textbf{Oscillation 2}} & \multicolumn{2}{c}{\textbf{E. coli growth}} & \multicolumn{2}{c}{\textbf{Stress-Strain}} \\
    \cline{2-9}
     & {Acc\textsubscript{avg-0.001}(\%)↑} & {NMSE↓} & {Acc\textsubscript{avg-0.001}(\%)↑} & {NMSE↓} & {Acc\textsubscript{avg-0.1}(\%)↑} & {NMSE↓} & {Acc\textsubscript{avg-0.1}(\%)↑} & {NMSE↓} \\
    \hline
    GPlern & 0.11 & 0.0972 & 0.05 & 0.2000 & 0.76 & 1.0023 & 28.43 & 0.3496 \\
    PySR & 3.80 & 0.0003 & \underline{7.02} & \underline{0.0002} & \underline{2.80} & 0.4068 & 70.60 & 0.0347 \\
    RAG-SR  & \underline{39.47} & \underline{1.49e-6} & 0.43 & 0.0282 & 2.04 & \underline{0.2754} & \underline{76.28} & \underline{0.0282} \\
    uDSR & 1.78 & 0.0002 & 0.36 & 0.0856 & 1.12 & 0.5059 & 59.15 & 0.0639 \\
    \hline
    LaSR (Llama-3.1) & 2.79 & 0.7485 & 1.09 & 0.0310 & \underline{3.44} & 0.1349 & 71.84 & \underline{0.0320} \\
    LLM-SR (Llama-3.1) & \underline{12.67} & 2.55e-5 & 8.20 & 4.70e-5 & 1.36 & 0.5815 & \underline{76.21} & 0.0333 \\
    LLM-SR (4o-mini) & 11.12 & \underline{2.07e-5} & \underline{8.66} & \underline{4.51e-5} & 3.24 & \underline{0.0863} & 71.28 & 0.0491 \\
    \hline
    MOT-SR (Llama-3.1) & {\cellcolor{gray!20}\textbf{100.00}} & {\cellcolor{gray!20}\textbf{1.27e-15}} & 99.45& {\cellcolor{gray!20}\textbf{1.70e-10}} & {\cellcolor{gray!20}\textbf{6.60}} & 0.0208 & 85.02 & 0.0150 \\
    MOT-SR (4o-mini) & 99.99 & 1.42e-13 & {\cellcolor{gray!20}\textbf{99.57}} & 4.25e-10 & 6.32 & {\cellcolor{gray!20}\textbf{0.0178}} & {\cellcolor{gray!20}\textbf{86.33}} & {\cellcolor{gray!20}\textbf{0.0144}} \\
    \hline
    \end{tabular}
    }
    \caption{Overall performance of MOT-SR and baseline methods on four benchmarks.}
    \label{tab:main}
    \vspace{-2em}
\end{table*}

\section{Experiments}

\subsection{Datasets}

To assess \textsc{MOT-SR}'s performance, we adopt two sets of challenging datasets. The first includes four standard benchmarks from \textsc{LLM-SR}, spanning nonlinear oscillatory systems (\textbf{Oscillation 1 \& 2}), where Oscillation 1 focuses on periodic signal composition and Oscillation 2 introduces cross-variable interactions and non-periodic disturbances to increase modeling difficulty, an \textbf{E. coli Growth} task modeling multivariate biological dynamics with nonlinear couplings~\citep{annurev:/content/journals/10.1146/annurev.mi.03.100149.002103,Rosso1995}, and a \textbf{Stress-Strain} task from materials science featuring piecewise nonlinear deformation behavior~\citep{Aakash2019}. 

The second is \textbf{LSR-Synth–Chemistry} from \textsc{LLM-SRBENCH}~\citep{shojaee2025llmsrbenchnewbenchmarkscientific}, which consists of 36 tasks derived from a chemical kinetics base equation with progressively increasing symbolic complexity. It is specifically designed to evaluate a model’s ability to generalize across nested, unseen, and semantically rich expressions. In this work, we focus on LSR-Synth–Chemistry rather than the complete \textsc{LLM-SRBENCH} suite because the other three datasets are constructed variants of the four benchmark problems already included in our evaluation. By concentrating on LSR-Synth–Chemistry, which is both novel and complementary, we ensure a comprehensive yet non-redundant assessment of model performance while also taking into account the practical constraints of available hardware resources. Full dataset descriptions are provided in Appendix~\ref{appendix:datasets} for completeness.

\subsection{Baselines}

We compare \textsc{MOT-SR} against a range of representative baselines from both classical and LLM-based SR methods. For the four standard tasks in the \textsc{LLM-SR} benchmark, we include \textsc{GPlearn}, a classical genetic programming-based SR method; \textsc{PySR}~\citep{grayeli2024symbolicregressionlearnedconcept}, which combines evolutionary search with symbolic compression; \textsc{uDSR}~\citep{NEURIPS2022_dbca58f3}, which replaces DSR's RNN policy with a pretrained Transformer and neural-guided decoding; \textsc{RAG-SR}~\citep{zhang2025ragsr}, which augments equation generation with structure retrieval; and \textsc{LLM-SR}~\citep{shojaee2025llmsrscientificequationdiscovery}.
On the more challenging LSR-Synth–Chemistry, we compare MOT-SR with leading LLM-enhanced methods, including \textsc{SGA}~\citep{pmlr-v235-ma24m}, which combines LLM-based hypothesis generation with physics-informed parameter optimization via bilevel search, and \textsc{LaSR}~\citep{grayeli2024symbolicregressionlearnedconcept}, which extracts abstract symbolic concepts from prior equations to guide hybrid LLM-evolutionary equation generation.

\subsection{Evaluation Metrics}

We evaluate different methods using three metrics: (1) Accuracy to tolerance $\tau$, denoted as $\mathrm{Acc_{all}}(\tau)$ and $\mathrm{Acc_{avg}}(\tau)$, and (2) Normalized Mean Squared Error (NMSE). $\mathrm{Acc_{all}}(\tau)$ measures task-level correctness by requiring all test points to satisfy the relative error bound $\tau$, i.e., $\mathrm{Acc_{all}}(\tau) = \mathbf{1}\left( \textstyle \max_{1 \le i \le N_{\text{test}}} \left| \frac{\hat{y}_i - y_i}{y_i} \right| \le \tau \right)$. $\mathrm{Acc_{avg}}(\tau)$ computes the proportion of test points that meet the same criterion, defined as $\mathrm{Acc_{avg}}(\tau) = \frac{1}{N_{\text{test}}} \sum_{i=1}^{N_{\text{test}}} \mathbf{1} \left( \textstyle \left| \frac{\hat{y}_i - y_i}{y_i} \right| \le \tau \right)$.

We further evaluate the quality of the final Pareto front using Hypervolume (HV)~\citep{797969} and Inverted Generational Distance (IGD)~\citep{6787994}, two standard indicators in multi-objective optimization (see Appendix~\ref{appendix:paleituo} for details). HV measures the volume dominated by the obtained solution set with respect to a fixed reference point, capturing both convergence and diversity. It reflects the overall coverage of the objective space and favors solution sets that are both well-converged and diverse. IGD computes the average distance from the ground truth equations to its nearest counterpart in the generated front, emphasizing approximation accuracy with respect to the true Pareto-optimal equations. For LLM-SR, we construct a pseudo Pareto front by extracting the nondominated set at each generation and aggregating them over 100 generations, yielding a hindsight-aggregated (best-of-run) front for a stringent comparison.

\begin{table}[t]
\centering
\setlength{\tabcolsep}{6pt}
\begin{tabular*}{0.8\linewidth}{@{\extracolsep{\fill}}lcc@{}}
\hline
\textbf{Model} & {Acc\textsubscript{all-0.1}(\%)$\uparrow$} & \textbf{NMSE$\downarrow$} \\
\hline
SGA       & 8.33   & 0.0458   \\
LaSR      & 27.77  & 2.77e-04 \\
LLM-SR    & 66.66  & 8.01e-06 \\
\hline
MOT-SR & {\cellcolor{gray!20}\textbf{86.11}} & {\cellcolor{gray!20}\textbf{3.85e-07}} \\
\hline
\end{tabular*}
\caption{Comparison on LSR-Synth--Chemistry.}
\label{tab:crk}
\vspace{-2em}
\end{table}

\subsection{MOT-SR Configuration}

For fair comparison, we adopt the same LLMs across all methods: \textsc{LLaMA-3.1-8B} and \textsc{GPT-4o-mini}, covering both lightweight and high-performance scenarios. In each iteration, \textsc{MOT-SR} generates four candidate equations for evaluation. The total number of iterations is set to 2000 for standard benchmarks and 1000 for LSR-Synth–Chemistry, following the \textsc{LLM-SRBench} protocol. Traditional baselines are allowed more iterations to ensure convergence. Additional details, including prompt design and sampling configurations, are provided in Appendix~\ref{appendix:prompt} and~\ref{appendix:MOT-SR}.

\section{Findings}

\subsection{MOT-SR Achieves the Best Overall Performance}

As shown in Table~\ref{tab:main}, \textsc{MOT-SR} consistently outperforms both classical and LLM-based SR baselines, achieving significantly lower NMSE and higher accuracy across all benchmarks. 
For example, on Oscillator 1 with LLaMA, \textsc{MOT-SR} reaches an NMSE of $1.27 \times 10^{-15}$, far surpassing
LLM-SR's $2.55 \times 10^{-5}$. With both LLaMA and GPT-4o-mini, \textsc{MOT-SR} attains over 90\% accuracy on several datasets.

On the more challenging \textsc{LSR-Synth–Chemistry} (Table~\ref{tab:crk}), \textsc{MOT-SR} achieves the best overall performance with an average NMSE of $3.85 \times 10^{-7}$ and 86.1\% accuracy, significantly outperforming other LLM-based models.
These results validate the effectiveness of \textsc{MOT-SR}'s tool-augmented multi-objective framework in enabling the discovery of more accurate equations.

\begin{figure}[t]
  \centering

  \begin{subfigure}[b]{0.4\textwidth}
    \centering
\includegraphics[width=\linewidth]{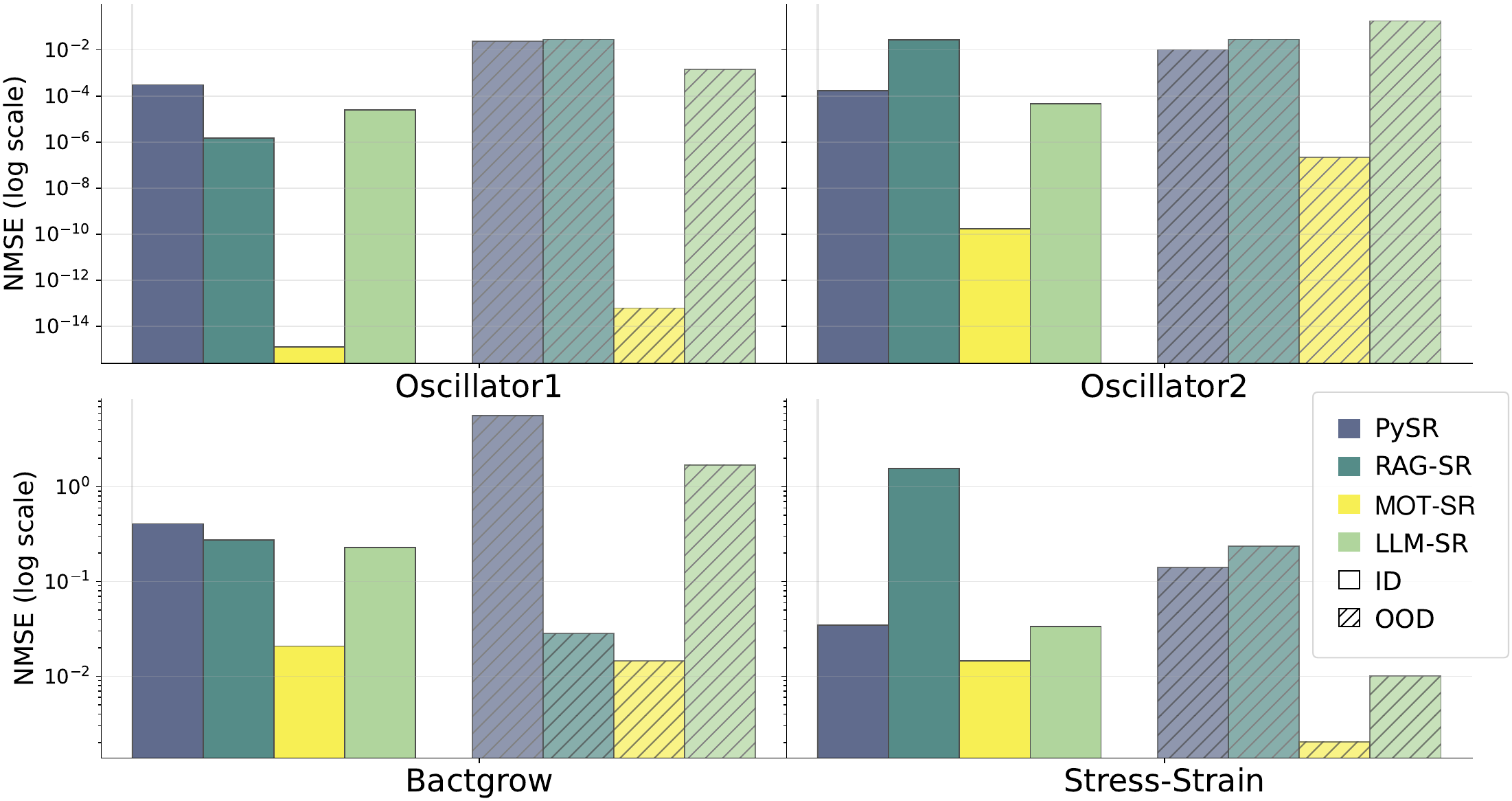}
    \caption{Generalization across domains under ID and OOD.}
    \label{fig:generalization1}
  \end{subfigure}
  \vspace{0.4em}

  \begin{subfigure}[b]{0.4\textwidth}
    \centering
\includegraphics[width=\linewidth]{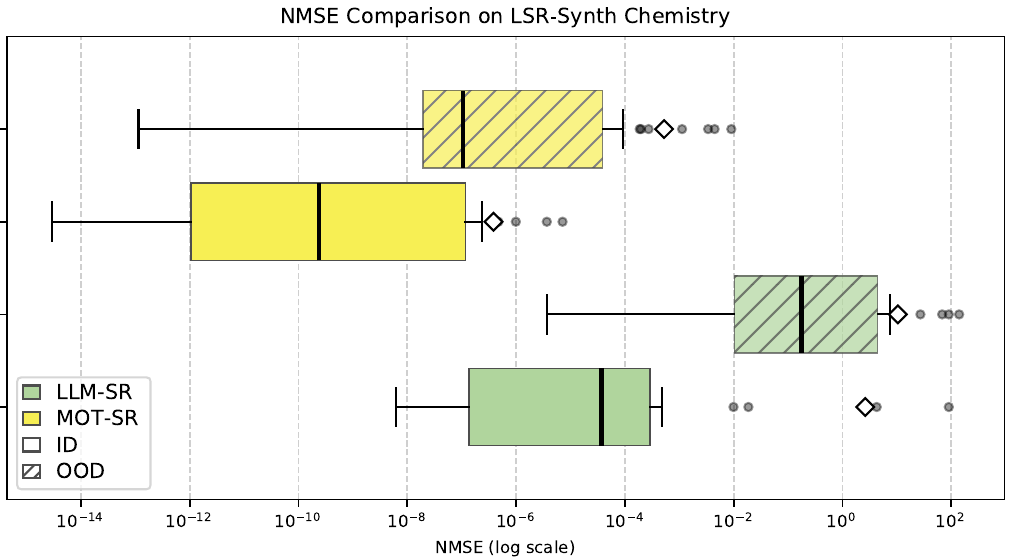}
    \caption{Boxplot of generalization on 36 chemistry tasks.}
    \label{fig:generalization2}
  \end{subfigure}

\caption{Comparison of generalization performance.}

  \label{fig:generalization_combined}
  \vspace{-2em}
\end{figure}

\subsection{MOT-SR Shows Stronger Generalization}
We evaluate generalization on the held-out test set under in-domain and out-of-domain conditions. In this section (Figure~\ref{fig:generalization_combined}), ``ID/OOD'' denotes the test partitions $D_{\mathrm{test}}^{\mathrm{ID}}$ and $D_{\mathrm{test}}^{\mathrm{OOD}}$, obtained by applying percentile-based bounds from $D_{\mathrm{train}}$ (Sec.~3.2) to the test inputs. These partitions are used only for final reporting and are not involved in parameter fitting or any discovery-loop objectives.

As shown in Figure~\ref{fig:generalization1}, MOT-SR consistently achieves the lowest NMSE across all four standard tasks under both $D_{\mathrm{test}}^{\mathrm{ID}}$ and $D_{\mathrm{test}}^{\mathrm{OOD}}$, significantly outperforming LLM-SR and other baselines. For instance, on the test OOD split of Oscillator~1 ($D_{\mathrm{test}}^{\mathrm{OOD}}$), MOT-SR reaches an NMSE of $6.20\times 10^{-14}$, nearly eleven orders of magnitude lower than LLM-SR ($1.4\times 10^{-3}$). Figure~\ref{fig:generalization2} further shows its superior median NMSEs across 36 chemistry tasks in LSR-Synth, under both test ID and test OOD conditions.

MOT-SR's generalization advantage stems from two key mechanisms. First, our multi-objective optimization explicitly incorporates performance on a training-derived OOD region, guiding the search toward expressions that maintain low error under distributional shifts. 
Second, our meta-strategy module integrates variable-level analysis and structural diagnostics: the former identifies stable input dependencies via analytical tools, while the latter detects effective substructures and potential failure modes across the population. Together, these components help MOT-SR uncover symbolic relationships that generalize beyond distribution-specific patterns.

\subsection{MOT-SR Improves Discovery Efficiency}

\begin{figure}[!t]
  \centering
\includegraphics[width=0.47\textwidth]{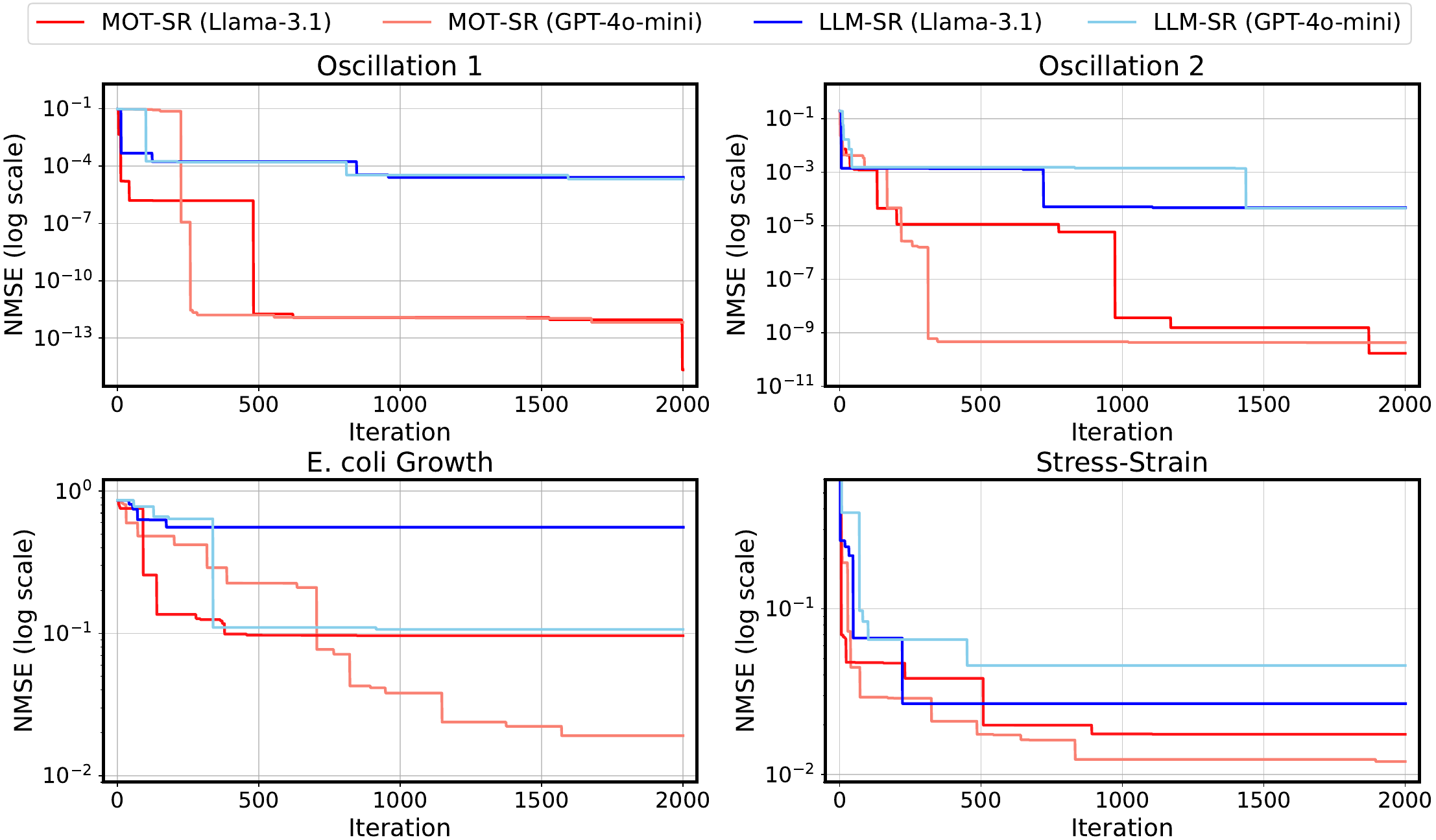}
  \caption{
  {Training convergence comparison.}
 }
  \label{fig:convergence}
\end{figure}

We compare the convergence behavior of MOT-SR and LLM-SR across four benchmark tasks. As shown in Figure~\ref{fig:convergence}, MOT-SR not only reduces error more rapidly but also converges to lower final NMSE values. In most cases, it outperforms LLM-SR’s best results (at 2000 iterations) within the first 1000 iterations.
This efficiency gain stems from \textsc{MOT-SR}’s meta-strategy design. By extracting key variable dependencies through analytical tools and summarizing structural patterns from the Pareto front, \textsc{MOT-SR} narrows the search space and avoids redundant exploration, steering the model toward high-quality equations with fewer iterations. These results highlight its efficiency in navigating the symbolic search space while maintaining both accuracy and generalization.
We further provide the computational cost analysis in Appendix~\ref{appendix:time-test}.

\subsection{MOT-SR Exhibits Superior Multi-objective Optimization Quality}

\begin{figure}[!t]
  \centering

  \begin{subfigure}[t]{0.22\textwidth}
    \centering
    \includegraphics[width=\linewidth]{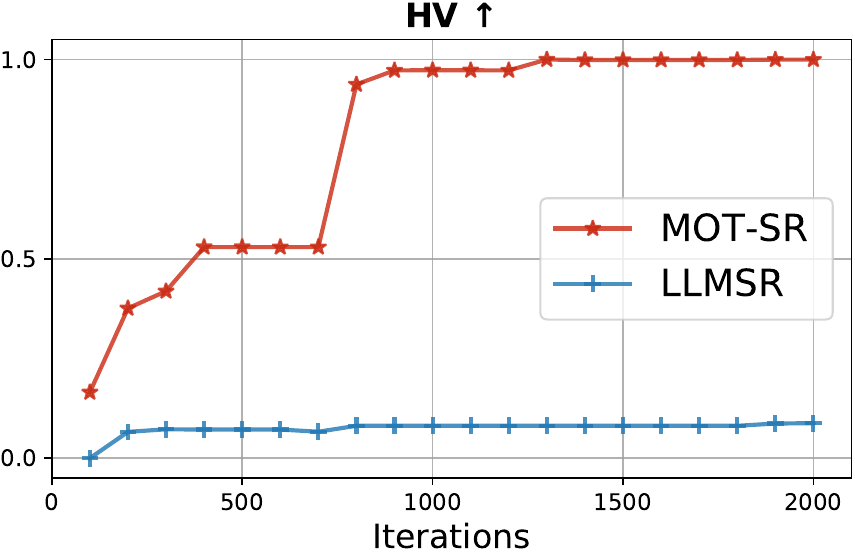}
    \caption{HV}
    \label{fig:hv}
  \end{subfigure}
\hspace{0.01\textwidth} 
  \begin{subfigure}[t]{0.22\textwidth}
    \centering
    \includegraphics[width=\linewidth]{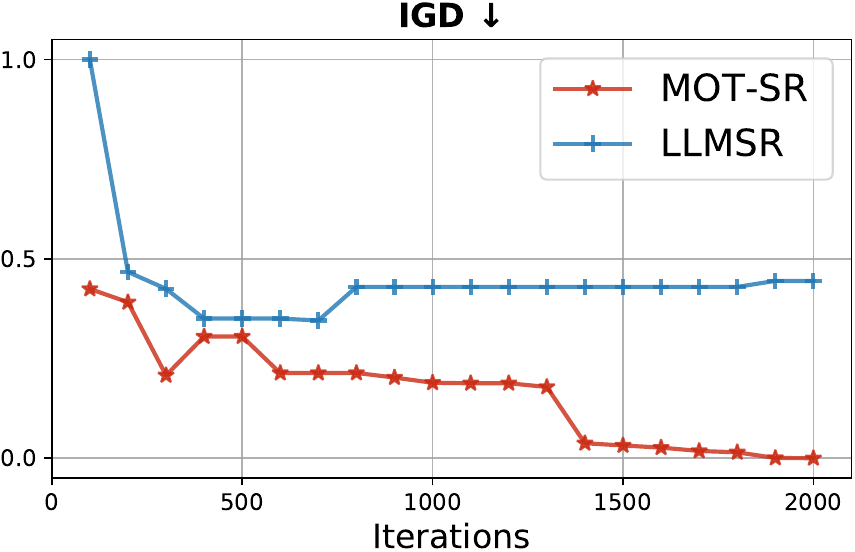}
    \caption{IGD}
    \label{fig:igd}
  \end{subfigure}

  \caption{Comparison of HV and IGD of different models.}
  \label{fig:pareto_quality}
\end{figure}

We compare MOT-SR and LLM-SR on Oscillator 1 using HV and IGD.
As shown in Figure~\ref{fig:hv}, MOT-SR achieves faster growth in HV compared to LLM-SR, indicating earlier and more effective discovery of diverse, high-quality non-dominated solutions. Similarly, Figure~\ref{fig:igd} shows that MOT-SR consistently reduces IGD at a faster rate, with better convergence toward the true Pareto front. LLM-SR, in contrast, plateaus earlier with higher IGD, reflecting a tendency to remain in suboptimal regions.
These results demonstrate that MOT-SR significantly outperforms baselines in terms of convergence speed, solution diversity, and optimization quality.

\subsection{Ablation Study}

To assess the contributions of key components in \textsc{MOT-SR}, we conduct ablation experiments on the Oscillator 1 benchmark with the LLaMA backbone. Our analysis focuses on two mechanisms: the multi-objective optimization mechanism and the meta strategy generator.

\subsubsection{Impact of Multi-objective Optimization}

We assess the impact of multi-objective optimization by replacing it with a single-objective variant (\textit{w/o MultiObj}) that optimizes only fitting error, excluding complexity and generalization.
In contrast, \textit{w/o MultiObj} recovers only 1 of 9 symbolic terms, compared to 4 of 6 for MOT-SR:

\noindent \textbf{Ground Truth}
{\small
\begin{center}
\begin{math}
\dot{v} = 0.8 \sin(x) - 0.5 x \cdot v - 0.5 v^3 - 0.2 x^3 - x \cos(x).
\end{math}
\end{center}
}

\noindent \textbf{MOT-SR w/o MultiObj}
{\small
\begin{center}
\begin{math}
\dot{v} = a_0 \tanh(a_1 x) - a_2 \tanh(a_3 v) - a_4 x - a_5 v 
- a_6\,\colorbox{pink}{$xv$} - a_7 (x - v) - a_8 (x + v) 
- a_9 v - (1 - a_9) x.
\end{math}
\end{center}
}

\noindent \textbf{MOT-SR}
{\small
\begin{center}
\begin{math}
\dot{v} = - a_0 x + a_1 v - a_2\,\colorbox{pink}{$x^3$} + a_3\,\colorbox{pink}{$v^3$} 
- a_4\,\colorbox{pink}{$x v$} + a_5\,\colorbox{pink}{$\sin(a_6\,x)$}.
\end{math}
\end{center}
}
This indicates multi-objective optimization is essential for discovering compact and interpretable equations. In contrast, single-objective optimization often leads to redundancy and overfitting.

\subsubsection{Effectiveness of the Meta Strategy Generator}

To further investigate the contribution of the meta strategy generator, we conduct a series of ablation studies by progressively removing its core components: the data analysis submodule (\textit{w/o Data}), the structure analysis submodule (\textit{w/o Struct}), and the entire module (\textit{w/o Strategy}).

\begin{figure}[!t]
  \centering
\includegraphics[width=0.4\textwidth]{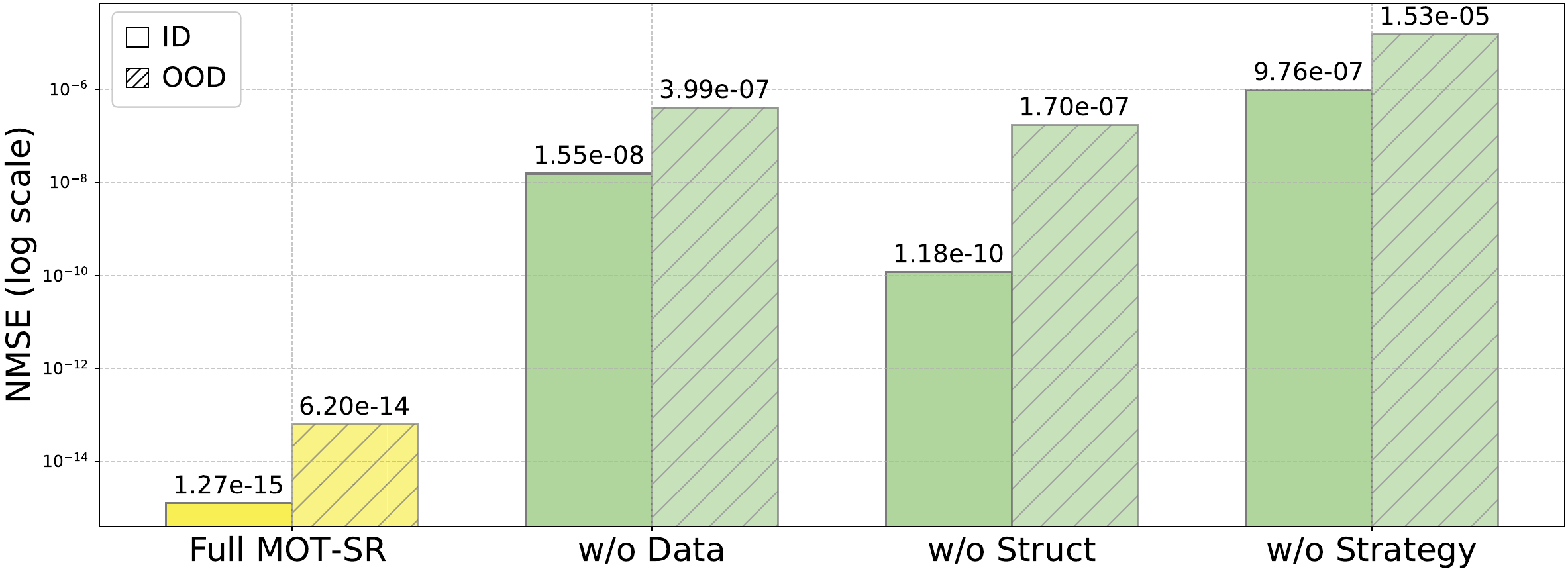}
  \caption{
  {Ablation results on the Oscillation 1 problem.}
 }
  \label{fig:Ablation}
  
\end{figure}
As shown in Figure~\ref{fig:Ablation}, removing the data analysis module impairs strategy refinement based on variable relationships, leading to a notable performance drop. Excluding the structure module results in a noticeable decline in generalization and accuracy, as the model can no longer extract structural patterns from prior equations to guide generation. Removing the entire meta strategy generator results in prompt-only generation without feedback, producing the lowest accuracy and stability among all variants.

These findings demonstrate that the data-driven and structure-guided feedback mechanisms offer complementary strengths in guiding \textsc{MOT-SR}’s equation generation. Evaluations of the ablation settings on other benchmark datasets, are provided in Appendix~\ref{app:more-xiaorong1}, and further ablation results on the role of different tools in the toolset are reported in Appendix~\ref{app:more-xiaorong2}.

\subsection{Case Study: EMRI Orbital Evolution Correction}

\paragraph{Problem and scientific objective.}
Space-based gravitational-wave detection is a major frontier in modern astronomy, opening the millihertz band to compact systems inaccessible to ground-based detectors~\citep{ruan2020taiji}. Extreme mass-ratio inspirals (EMRIs), in which a compact object inspirals around a much more massive black hole, are key sources in this band and probe the spacetime and physical processes near the central black hole~\citep{amaro2023astrophysics}. Their many orbital cycles require accurate evolution models, since small local errors can cause substantial long-term trajectory and phase deviations. We present MOT-SR as the first application of SR to EMRI orbital-evolution modeling and use this task to examine whether symbolic equation discovery can address a practical modeling problem in gravitational-wave astrophysics. Pn5AAK is widely used for its computational efficiency but may lose accuracy over long evolutions, while self-force calculations are more accurate but costly and currently limited in scope. We study eccentric equatorial orbits in Schwarzschild spacetime using \texttt{Fast Self-forced Inspirals} (FSI)~\citep{2018CQGra..35n4003V} as the reference and \texttt{Pn5AAK}~\citep{katz2021fast}, denoted PN5 below, as the approximation. This comparison quantifies accumulated orbital deviations and identifies where PN5 remains reliable.

We focus on correcting the evolution of the dimensionless semi-latus rectum \(p\). 
Given the eccentricity \(e\) and mass ratio \(\eta\), the learning target is the 
discrepancy between the local evolution rates predicted by FSI and PN5, defined as 
\(\Delta \dot{p}(p,e,\eta)=\dot{p}_{\mathrm{FSI}}(p,e,\eta)-\dot{p}_{\mathrm{PN5}}(p,e,\eta)\).

The discovered residual is added to the PN5 evolution rate and integrated over time to obtain the corrected trajectory $p_{\mathrm{corr}}(t)$. This setting evaluates whether an equation fitted to evolution-rate data remains reliable when repeatedly applied within a dynamical solver. Full details of the residual construction, unit conversion, equation search, and numerical integration are provided in Appendix~\ref{app:emri}.

\paragraph{Dataset construction.}
We generate Schwarzschild eccentric-EMRI trajectories with central black hole masses between \(10^4\) and \(10^7 M_\odot\), secondary masses between \(5\) and \(100 M_\odot\), initial eccentricities \(0 \leq e_0 \leq 0.2\), and initial semi-latus recta satisfying \(6+2e_0+0.05 \leq p_0 \leq 12\). Each case is simulated with both FSI and PN5, and the resulting evolution rates are aligned at the same orbital states before the residual labels are computed. Among the 100 generated cases, Cases 89--100 contain only one or two valid states and are removed before evaluation. The remaining 88 cases are divided at the trajectory level: 58 complete trajectories are used for equation discovery and parameter fitting, and 30 complete trajectories are held out for testing. No trajectory contributes samples to both subsets. The equation structure, fitted constants, and candidate selection are fixed before evaluation on the 30 held-out cases.

\paragraph{Baselines and evaluation.}
We compare against NN, an adapted neural residual baseline motivated by the neural UDE formulation introduced by Keith et al.~\cite{PhysRevResearch.3.043101}. NN uses a feed-forward ResidualMLP with hidden dimensions \([128,128,64]\) and SiLU activations to predict \(\Delta\dot p\) from \((p,e,\eta)\). The predicted residual is added to the PN5 evolution rate before integration. We also report LLM-SR as an alternative LLM-based symbolic correction.

Following the standard a posteriori evaluation protocol for scientific equation discovery, 
we assess the complete integrated trajectory rather than point-wise residual fitting. 
For each case, we apply the NMSE defined in Eq.~\eqref{eq:nmse} to the predicted trajectory 
\(p_{\mathrm{corr}}(t)\) and the FSI reference \(p_{\mathrm{FSI}}(t)\).

\paragraph{Results.}

\begin{table}[t]
    \centering

    \small
    \setlength{\tabcolsep}{4pt}
    \renewcommand{\arraystretch}{1.08}

    \begin{tabular}{@{}lccc@{}}
        \toprule
        Evaluation group
        & NN
        & MOT-SR
        & LLM-SR \\
        \midrule

        \multicolumn{4}{@{}l}{\textit{(a) Overall evaluation}} \\
        Discovery cases (58)
        & 1.62
        & \textbf{3.48e-03}
        & 2.65e-01 \\
        Held-out cases (30)
        & 1.20
        & \textbf{1.17e-03}
        & 3.13e-02 \\
        All valid cases (88)
        & 1.48
        & \textbf{2.69e-03}
        & 1.86e-01 \\

        \addlinespace[3pt]
        \multicolumn{4}{@{}l}{\textit{(b) Held-out cases by trajectory length}} \\
        Medium (500--2,000)
        & 4.47e-01
        & \textbf{1.32e-04}
        & 1.09e-02 \\
        Long (2,000--10,000)
        & 1.47
        & \textbf{8.18e-05}
        & 4.19e-03 \\
        Extra-long (\(>10{,}000\))
        & 2.81
        & \textbf{1.40e-03}
        & 8.56e-02 \\
        \bottomrule
    \end{tabular}

    \caption{Mean case-level NMSE of the integrated \(p(t)\) trajectories.
    Panel (a) reports results on the discovery, held-out, and complete sets.
    Panel (b) groups the held-out cases by trajectory length.}
    \label{tab:emri_results}
    \vspace{-1em}
\end{table}

Table~\ref{tab:emri_results} reports the mean case-level NMSE. Lower values indicate closer agreement with the FSI reference trajectory.
On the 30 held-out trajectories, MOT-SR achieves a mean NMSE of \(1.17\times10^{-3}\), approximately three orders of magnitude lower than NN and \(26.8\) times lower than LLM-SR. MOT-SR also consistently achieves the lowest mean NMSE among the three methods on the 58 discovery cases and across all 88 valid cases, demonstrating its accuracy under complete-trajectory integration. More importantly, the evaluation split is constructed at the case level: each held-out trajectory represents a distinct EMRI configuration that is entirely excluded from equation discovery. The shared symbolic correction discovered from 58 cases remains effective when integrated over 30 unseen configurations. This result provides evidence that MOT-SR generalizes across EMRI cases, instead of relying on samples drawn from trajectories encountered during discovery.

To examine error accumulation over longer integration horizons, Table~\ref{tab:emri_results} groups the held-out trajectories with more than 500 valid integration points by trajectory length. The length-stratified results reveal clear differences in long-term integration behavior. The mean NMSE of NN increases monotonically from \(4.47\times10^{-1}\) on medium trajectories to \(1.47\) on long trajectories and \(2.81\) on extra-long trajectories. This trend indicates that small errors in the neural residual accumulate when the fitted correction is repeatedly applied during integration. In comparison, MOT-SR maintains a mean NMSE between \(8.18\times10^{-5}\) and \(1.40\times10^{-3}\) and achieves the lowest error in every trajectory-length group. 

This behavior is consistent with the training design of MOT-SR. MOT-SR divides \(\mathcal{D}_{\mathrm{train}}\) into ID and OOD subsets and evaluates candidate equations for both ID accuracy and OOD generalization during discovery. This internal generalization criterion discourages equations whose accuracy is restricted to a narrow subset of the discovery configurations, yielding a correction that remains reliable under repeated long-term integration. Compared with LLM-SR, MOT-SR reduces the mean NMSE by factors of \(82.6\), \(51.2\), and \(61.1\) on the medium, long, and extra-long trajectories, respectively. The consistent advantage over LLM-SR shows that producing a symbolic expression alone does not fully realize the potential of SR for this task; the ID/OOD-aware discovery process of MOT-SR further exploits the ability of symbolic equations to capture correction structures shared across EMRI configurations. Collectively, these results show that MOT-SR preserves low integration error across different trajectory lengths and unseen EMRI configurations.

\begin{figure}[!t]
    \centering

    \begin{subfigure}[t]{0.49\columnwidth}
        \centering
        \includegraphics[
            width=\linewidth
        ]{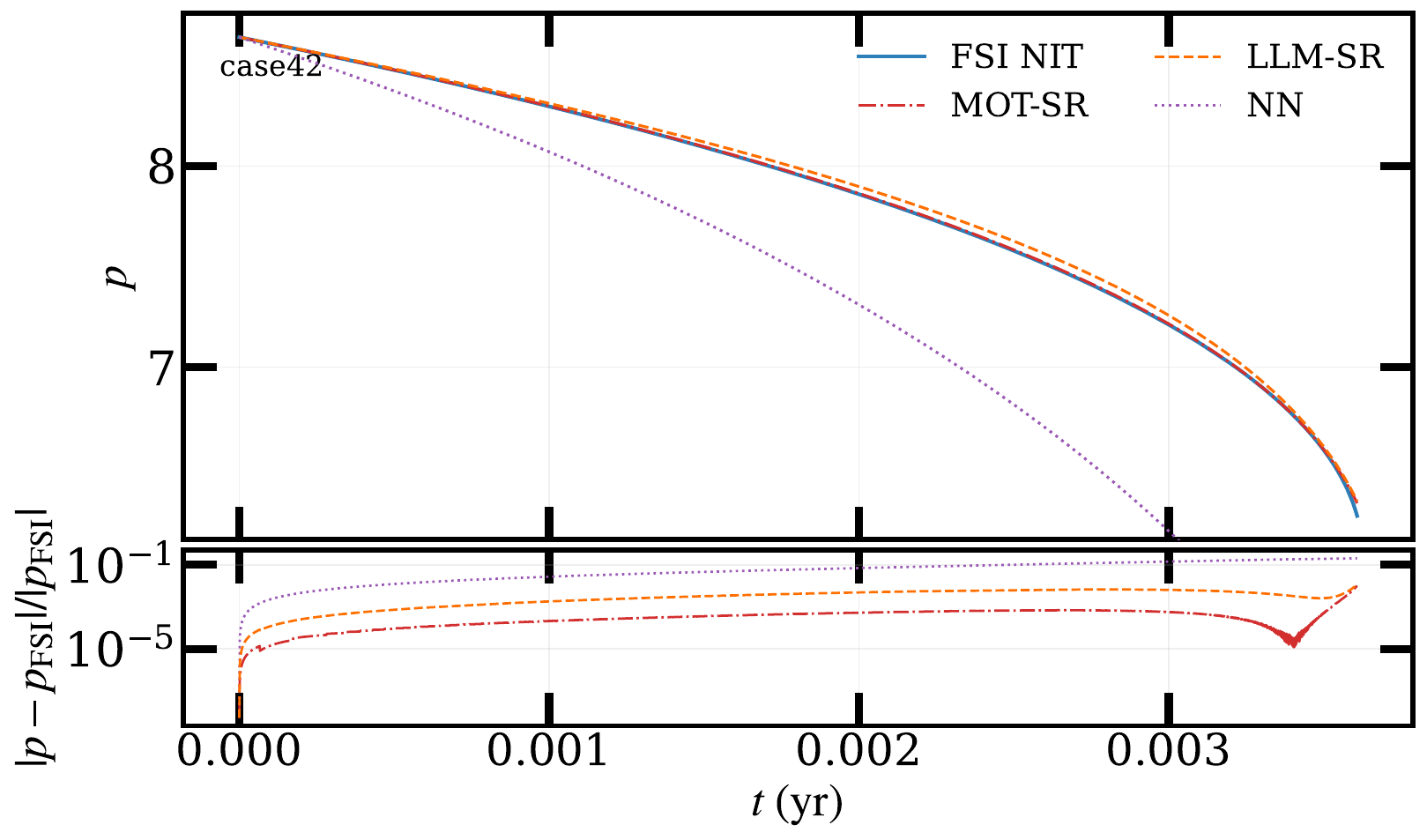}
        \caption{Case 42}
        \label{fig:emri_case42}
    \end{subfigure}
    \hfill
    \begin{subfigure}[t]{0.49\columnwidth}
        \centering
        \includegraphics[
            width=\linewidth
        ]{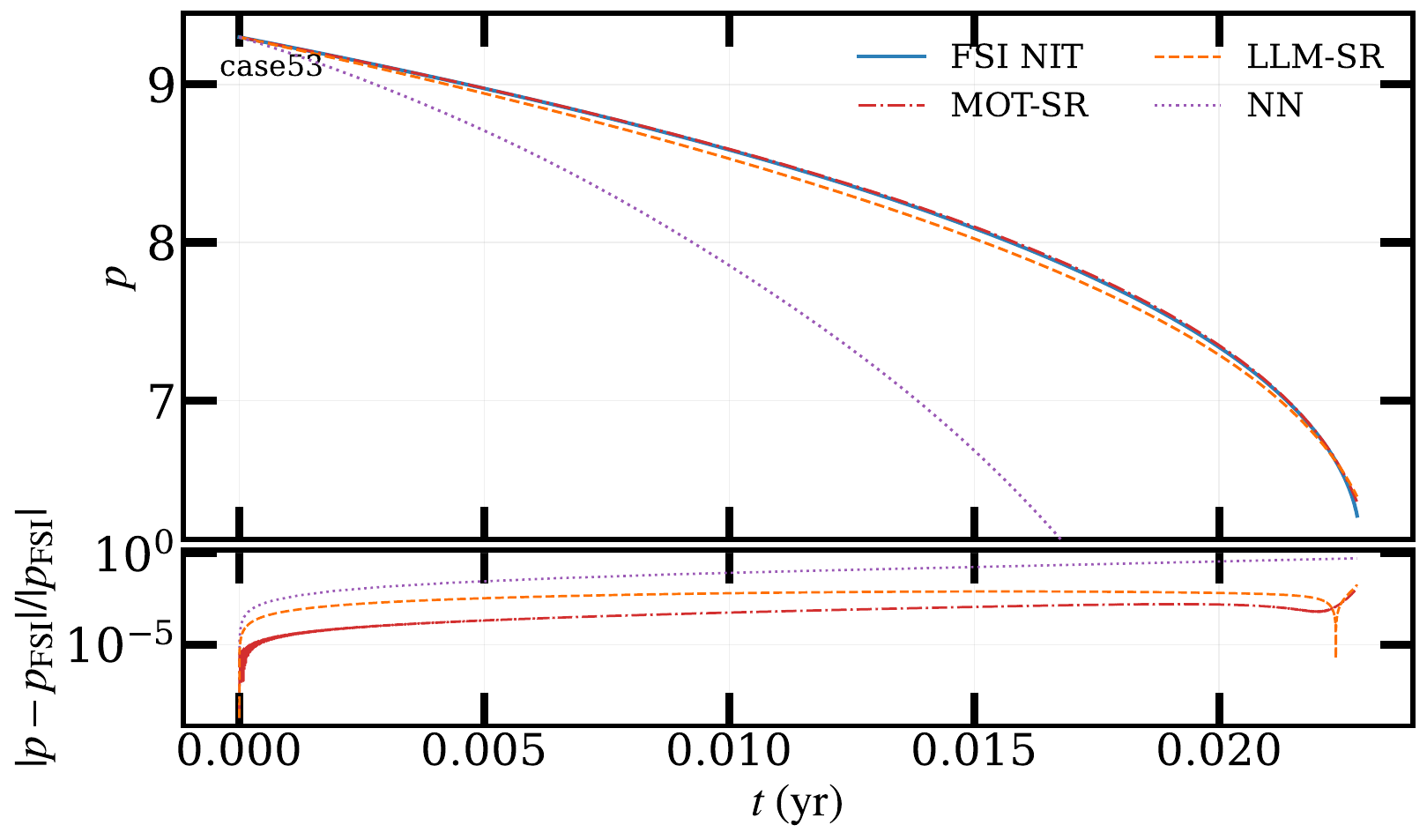}
        \caption{Case 53}
        \label{fig:emri_case53}
    \end{subfigure}

    \vspace{-1mm}
    \caption{A posteriori integration for two held-out EMRI configurations, comparing FSI with NN, LLM-SR, and MOT-SR. Quantitative results are based on all 30 held-out cases.}
    \label{fig:emri_cases}
   
\end{figure}

Figure~\ref{fig:emri_cases} provides two illustrative examples. In both cases, the trajectory produced by MOT-SR remains closer to the FSI reference over the plotted integration horizon, while the neural residual leads to a visibly larger deviation. Together with the aggregate results, these examples provide evidence of trajectory-level transfer across unseen EMRI configurations within the considered Schwarzschild equatorial eccentric regime.

\section{Conclusion}

We introduced \textsc{MOT-SR}, a unified symbolic regression framework combining tool-augmented variable analysis, multi-objective evaluation, and cooperative LLMs for scientific equation discovery. Across standard benchmarks, \textsc{MOT-SR} consistently improves predictive accuracy, generalization, equation compactness, and search efficiency over classical and LLM-based baselines. Beyond benchmark tasks, \textsc{MOT-SR} discovers a compact symbolic correction for EMRI orbital evolution that substantially reduces long-horizon trajectory-level integration error and remains effective across held-out configurations. These results demonstrate the potential of tool-augmented, multi-objective equation discovery for building interpretable and transferable models of complex scientific dynamics.

\section*{Limitations and Ethical Considerations}

\textsc{MOT-SR} currently relies on a fixed set of evaluation objectives and manually specified analytical tools. Future work will introduce domain-specific objectives, automate tool discovery and refinement, and extend \textsc{MOT-SR} to high-dimensional scientific data and more complex dynamical systems. This work involves no human participants or personal data. The benchmark and astrophysical data are used solely for scientific equation discovery and model evaluation.

\newpage

\bibliographystyle{ACM-Reference-Format}
\bibliography{example_paper}

@article{article,
author = {Makke, Nour and Chawla, Sanjay},
year = {2024},
month = {01},
pages = {},
title = {Interpretable scientific discovery with symbolic regression: a review},
volume = {57},
journal = {Artificial Intelligence Review},
doi = {10.1007/s10462-023-10622-0}
}

@article{10.1093/pnasnexus/pgae467,
    author = {Makke, Nour and Chawla, Sanjay},
    title = {Data-driven discovery of Tsallis-like distribution using symbolic regression in high-energy physics},
    journal = {PNAS Nexus},
    volume = {3},
    number = {11},
    pages = {pgae467},
    year = {2024},
    month = {10},
    issn = {2752-6542},
    doi = {10.1093/pnasnexus/pgae467},
    url = {https://doi.org/10.1093/pnasnexus/pgae467},
    eprint = {https://academic.oup.com/pnasnexus/article-pdf/3/11/pgae467/60816181/pgae467.pdf},
}

@InProceedings{10.1007/978-3-031-29573-7_3,
author="Reuter, Julia
and Elmestikawy, Hani
and Evrard, Fabien
and Mostaghim, Sanaz
and van Wachem, Berend",
editor="Pappa, Gisele
and Giacobini, Mario
and Vasicek, Zdenek",
title="Graph Networks as Inductive Bias for Genetic Programming: Symbolic Models for Particle-Laden Flows",
booktitle="Genetic Programming",
year="2023",
publisher="Springer Nature Switzerland",
address="Cham",
pages="36--51",
isbn="978-3-031-29573-7"
}

@inproceedings{ICLR2025_a76b693f,
 author = {Chen, Jindou and Tian, Jidong and Wu, Liang and ChenXinWei and Yang, Xiaokang and Jin, Yaohui and Xu, Yanyan},
 booktitle = {International Conference on Representation Learning},
 editor = {Y. Yue and A. Garg and N. Peng and F. Sha and R. Yu},
 pages = {67058--67080},
 title = {KinFormer: Generalizable Dynamical Symbolic Regression for Catalytic Organic Reaction Kinetics},
 url = {https://proceedings.iclr.cc/paper_files/paper/2025/file/a76b693f36916a5ed84d6e5b39a0dc03-Paper-Conference.pdf},
 volume = {2025},
 year = {2025}
}

@article{DENG2023109010,
title = {EV charging load forecasting model mining algorithm based on hybrid intelligence},
journal = {Computers and Electrical Engineering},
volume = {112},
pages = {109010},
year = {2023},
issn = {0045-7906},
doi = {https://doi.org/10.1016/j.compeleceng.2023.109010},
url = {https://www.sciencedirect.com/science/article/pii/S0045790623004342},
author = {Song Deng and Junjie Wang and Li Tao and Su Zhang and Hongwei Sun}
}

@article{Wahlquist2024,
  author = {Wahlquist, Ylva and Sundell, Jesper and Soltesz, Kristian},
  title = {Learning pharmacometric covariate model structures with symbolic regression networks},
  journal = {Journal of Pharmacokinetics and Pharmacodynamics},
  year = {2024},
  volume = {51},
  number = {2},
  pages = {155-167},
  doi = {10.1007/s10928-023-09887-3},
  url = {https://doi.org/10.1007/s10928-023-09887-3},
  issn = {1573-8744}
}

@misc{shi2024alphaforgeframeworkdynamicallycombine,
      title={AlphaForge: A Framework to Mine and Dynamically Combine Formulaic Alpha Factors}, 
      author={Hao Shi and Weili Song and Xinting Zhang and Jiahe Shi and Cuicui Luo and Xiang Ao and Hamid Arian and Luis Seco},
      year={2024},
      eprint={2406.18394},
      archivePrefix={arXiv},
      primaryClass={q-fin.CP},
      url={https://arxiv.org/abs/2406.18394}, 
}

@misc{virgolin2022symbolicregressionnphard,
      title={Symbolic Regression is NP-hard}, 
      author={Marco Virgolin and Solon P. Pissis},
      year={2022},
      eprint={2207.01018},
      archivePrefix={arXiv},
      primaryClass={cs.NE},
      url={https://arxiv.org/abs/2207.01018}, 
}

@article{
doi:10.1126/science.1165893,
author = {Michael Schmidt  and Hod Lipson },
title = {Distilling Free-Form Natural Laws from Experimental Data},
journal = {Science},
volume = {324},
number = {5923},
pages = {81-85},
year = {2009},
doi = {10.1126/science.1165893},
URL = {https://www.science.org/doi/abs/10.1126/science.1165893},
eprint = {https://www.science.org/doi/pdf/10.1126/science.1165893}}

@misc{cranmer2023interpretablemachinelearningscience,
      title={Interpretable Machine Learning for Science with PySR and SymbolicRegression.jl}, 
      author={Miles Cranmer},
      year={2023},
      eprint={2305.01582},
      archivePrefix={arXiv},
      primaryClass={astro-ph.IM},
      url={https://arxiv.org/abs/2305.01582}, 
}

@misc{petersen2021deepsymbolicregressionrecovering,
      title={Deep symbolic regression: Recovering mathematical expressions from data via risk-seeking policy gradients}, 
      author={Brenden K. Petersen and Mikel Landajuela and T. Nathan Mundhenk and Claudio P. Santiago and Soo K. Kim and Joanne T. Kim},
      year={2021},
      eprint={1912.04871},
      archivePrefix={arXiv},
      primaryClass={cs.LG},
      url={https://arxiv.org/abs/1912.04871}, 
}

@misc{biggio2021neuralsymbolicregressionscales,
      title={Neural Symbolic Regression that Scales}, 
      author={Luca Biggio and Tommaso Bendinelli and Alexander Neitz and Aurelien Lucchi and Giambattista Parascandolo},
      year={2021},
      eprint={2106.06427},
      archivePrefix={arXiv},
      primaryClass={cs.LG},
      url={https://arxiv.org/abs/2106.06427}, 
}

@misc{kamienny2022endtoendsymbolicregressiontransformers,
      title={End-to-end symbolic regression with transformers}, 
      author={Pierre-Alexandre Kamienny and Stéphane d'Ascoli and Guillaume Lample and François Charton},
      year={2022},
      eprint={2204.10532},
      archivePrefix={arXiv},
      primaryClass={cs.LG},
      url={https://arxiv.org/abs/2204.10532}, 
}

@misc{shojaee2025llmsrscientificequationdiscovery,
      title={LLM-SR: Scientific Equation Discovery via Programming with Large Language Models}, 
      author={Parshin Shojaee and Kazem Meidani and Shashank Gupta and Amir Barati Farimani and Chandan K Reddy},
      year={2025},
      eprint={2404.18400},
      archivePrefix={arXiv},
      primaryClass={cs.LG},
      url={https://arxiv.org/abs/2404.18400}, 
}

@misc{grayeli2024symbolicregressionlearnedconcept,
      title={Symbolic Regression with a Learned Concept Library}, 
      author={Arya Grayeli and Atharva Sehgal and Omar Costilla-Reyes and Miles Cranmer and Swarat Chaudhuri},
      year={2024},
      eprint={2409.09359},
      archivePrefix={arXiv},
      primaryClass={cs.LG},
      url={https://arxiv.org/abs/2409.09359}, 
}

@misc{openai_gpt4o_mini2024,
  author = {OpenAI},
  title = {GPT‑4o mini: advancing cost‑efficient intelligence},
  year = {2024},
  month = jul,
  howpublished = {\url{https://openai.com/index/gpt-4o-mini-advancing-cost-efficient-intelligence/}},
  note = {Accessed: 2025‑07‑29}
}

@misc{kassianik2025llama31foundationaisecurityllmbase8btechnicalreport,
      title={Llama-3.1-FoundationAI-SecurityLLM-Base-8B Technical Report}, 
      author={Paul Kassianik and Baturay Saglam and Alexander Chen and Blaine Nelson and Anu Vellore and Massimo Aufiero and Fraser Burch and Dhruv Kedia and Avi Zohary and Sajana Weerawardhena and Aman Priyanshu and Adam Swanda and Amy Chang and Hyrum Anderson and Kojin Oshiba and Omar Santos and Yaron Singer and Amin Karbasi},
      year={2025},
      eprint={2504.21039},
      archivePrefix={arXiv},
      primaryClass={cs.CR},
      url={https://arxiv.org/abs/2504.21039}, 
}

@book{fletcher1987_practical_methods,
  author       = {Fletcher, Roger},
  title        = {Practical Methods of Optimization},
  edition      = {2nd},
  publisher    = {John Wiley \& Sons},
  address      = {Chichester, New York},
  year         = {1987},
  isbn         = {0471915475},
}

@article{a0dc553c-0830-3fe2-ab2a-eece0d66a7db,
 ISSN = {03701662},
 URL = {http://www.jstor.org/stable/115794},
 author = {Karl Pearson},
 journal = {Proceedings of the Royal Society of London},
 pages = {240--242},
 publisher = {The Royal Society},
 title = {Note on Regression and Inheritance in the Case of Two Parents},
 urldate = {2025-07-27},
 volume = {58},
 year = {1895}
}

@book{Montgomery2013,
  author    = {Douglas C. Montgomery and Elizabeth A. Peck and G. Geoffrey Vining},
  title     = {Introduction to Linear Regression Analysis},
  edition   = {5},
  publisher = {Wiley},
  year      = {2013},
  address   = {Hoboken, NJ},
  isbn      = {978‑1118627365}
}

@book{jolliffe2002_principal_component_analysis,
  author    = {Jolliffe, I.~T.},
  title     = {Principal Component Analysis},
  edition   = {2nd},
  series    = {Springer Series in Statistics},
  publisher = {Springer‑Verlag},
  address   = {New York, NY, USA},
  year      = {2002},
  isbn      = {0-387-95442-2},
  doi       = {10.1007/b98835},
}

@article{ca468a70-0be4-389a-b0b9-5dd1ff52b33f,
 ISSN = {00029556},
 URL = {http://www.jstor.org/stable/1412159},
 author = {C. Spearman},
 journal = {The American Journal of Psychology},
 number = {1},
 pages = {72--101},
 publisher = {University of Illinois Press},
 title = {The Proof and Measurement of Association between Two Things},
 urldate = {2025-07-27},
 volume = {15},
 year = {1904}
}

@ARTICLE{6773024,
  author={Shannon, C. E.},
  journal={The Bell System Technical Journal}, 
  title={A mathematical theory of communication}, 
  year={1948},
  volume={27},
  number={3},
  pages={379-423},
  doi={10.1002/j.1538-7305.1948.tb01338.x}}

@article{pedregosa2011_scikitlearn,
  author  = {Pedregosa, Fabian and Varoquaux, Gaël and Gramfort, Alexandre and Michel, Vincent and Thirion, Bertrand and Grisel, Olivier and Blondel, Mathieu and Prettenhofer, Peter and Weiss, Ron and Dubourg, Vincent and Vanderplas, Jake and Passos, Alexandre and Cournapeau, David and Brucher, Matthieu and Perrot, Matthieu and Duchesnay, Édouard},
  title   = {Scikit‑learn: Machine Learning in Python},
  journal = {Journal of Machine Learning Research},
  volume  = {12},
  pages   = {2825--2830},
  year    = {2011}
}

@article{Cooley:1965zz,
    author = "Cooley, James W. and Tukey, John W.",
    title = "{An Algorithm for the Machine Calculation of Complex Fourier Series}",
    doi = "10.1090/S0025-5718-1965-0178586-1",
    journal = "Math. Comput.",
    volume = "19",
    pages = "297--301",
    year = "1965"
}

@book{daubechies1992_ten_lectures,
  author    = {Daubechies, Ingrid},
  title     = {Ten Lectures on Wavelets},
  series    = {CBMS‐NSF Regional Conference Series in Applied Mathematics},
  publisher = {Society for Industrial and Applied Mathematics (SIAM)},
  address   = {Philadelphia, PA, USA},
  year      = {1992},
  isbn      = {978-0-89871-274-2}
}

@article{702ab909-8cb1-3c30-a5f1-ab4517d6cf1c,
 ISSN = {00129682, 14680262},
 URL = {http://www.jstor.org/stable/1912791},
 author = {C. W. J. Granger},
 journal = {Econometrica},
 number = {3},
 pages = {424--438},
 publisher = {[Wiley, Econometric Society]},
 title = {Investigating Causal Relations by Econometric Models and Cross-spectral Methods},
 urldate = {2025-07-27},
 volume = {37},
 year = {1969}
}

@article{
doi:10.1126/science.1227079,
author = {George Sugihara  and Robert May  and Hao Ye  and Chih-hao Hsieh  and Ethan Deyle  and Michael Fogarty  and Stephan Munch },
title = {Detecting Causality in Complex Ecosystems},
journal = {Science},
volume = {338},
number = {6106},
pages = {496-500},
year = {2012},
doi = {10.1126/science.1227079},
URL = {https://www.science.org/doi/abs/10.1126/science.1227079},
eprint = {https://www.science.org/doi/pdf/10.1126/science.1227079},
}

@article{WOLF1985285,
title = {Determining Lyapunov exponents from a time series},
journal = {Physica D: Nonlinear Phenomena},
volume = {16},
number = {3},
pages = {285-317},
year = {1985},
issn = {0167-2789},
doi = {https://doi.org/10.1016/0167-2789(85)90011-9},
url = {https://www.sciencedirect.com/science/article/pii/0167278985900119},
author = {Alan Wolf and Jack B. Swift and Harry L. Swinney and John A. Vastano},
}

@article{GRASSBERGER1983189,
title = {Measuring the strangeness of strange attractors},
journal = {Physica D: Nonlinear Phenomena},
volume = {9},
number = {1},
pages = {189-208},
year = {1983},
issn = {0167-2789},
doi = {https://doi.org/10.1016/0167-2789(83)90298-1},
url = {https://www.sciencedirect.com/science/article/pii/0167278983902981},
author = {Peter Grassberger and Itamar Procaccia}
}

@ARTICLE{1163055,
  author={Sakoe, H. and Chiba, S.},
  journal={IEEE Transactions on Acoustics, Speech, and Signal Processing}, 
  title={Dynamic programming algorithm optimization for spoken word recognition}, 
  year={1978},
  volume={26},
  number={1},
  pages={43-49},
  doi={10.1109/TASSP.1978.1163055}}

@article{16e7f618-c06b-3d10-8705-1086b218d827,
 ISSN = {01621459, 1537274X},
 URL = {http://www.jstor.org/stable/2280095},
 author = {Frank J. Massey},
 journal = {Journal of the American Statistical Association},
 number = {253},
 pages = {68--78},
 publisher = {[American Statistical Association, Taylor & Francis, Ltd.]},
 title = {The Kolmogorov-Smirnov Test for Goodness of Fit},
 urldate = {2025-07-27},
 volume = {46},
 year = {1951}
}

@inproceedings{10.1145/1083142.1083143,
author = {Neamtiu, Iulian and Foster, Jeffrey S. and Hicks, Michael},
title = {Understanding source code evolution using abstract syntax tree matching},
year = {2005},
isbn = {1595931236},
publisher = {Association for Computing Machinery},
address = {New York, NY, USA},
url = {https://doi.org/10.1145/1083142.1083143},
doi = {10.1145/1083142.1083143},
booktitle = {Proceedings of the 2005 International Workshop on Mining Software Repositories},
pages = {1–5},
numpages = {5},
location = {St. Louis, Missouri},
series = {MSR '05}
}

@misc{shojaee2025llmsrbenchnewbenchmarkscientific,
      title={LLM-SRBench: A New Benchmark for Scientific Equation Discovery with Large Language Models}, 
      author={Parshin Shojaee and Ngoc-Hieu Nguyen and Kazem Meidani and Amir Barati Farimani and Khoa D Doan and Chandan K Reddy},
      year={2025},
      eprint={2504.10415},
      archivePrefix={arXiv},
      primaryClass={cs.CL},
      url={https://arxiv.org/abs/2504.10415}, 
}

@article{annurev:/content/journals/10.1146/annurev.mi.03.100149.002103,
   author = "Monod, Jacques",
   title = "THE GROWTH OF BACTERIAL CULTURES", 
   journal= "Annual Review of Microbiology",
   year = "1949",
   volume = "3",
   number = "Volume 3, 1949",
   pages = "371-394",
   doi = "https://doi.org/10.1146/annurev.mi.03.100149.002103",
   url = "https://www.annualreviews.org/content/journals/10.1146/annurev.mi.03.100149.002103",
   publisher = "Annual Reviews",
   issn = "1545-3251",
   type = "Journal Article",
  }

@article{Rosso1995,
  author = {Rosso, L and Lobry, J. R. and Bajard, S and Flandrois, J. P.},
  title = {Convenient Model To Describe the Combined Effects of Temperature and {pH} on Microbial Growth},
  journal = {Applied and Environmental Microbiology},
  year = {1995},
  volume = {61},
  number = {2},
  pages = {610-6},
  month = {Feb},
  doi = {10.1128/aem.61.2.610-616.1995},
  url = {https://doi.org/10.1128/aem.61.2.610-616.1995},
  issn = {0099-2240},
  language = {eng},
  publisher = {American Society for Microbiology},
  country = {United States}
}

@article{Aakash2019,
  author = {Aakash, B. S. and Connors, JohnPatrick and Shields, Michael D},
  title = {Stress-strain data for aluminum 6061-T651 from 9 lots at 6 temperatures under uniaxial and plane strain tension},
  journal = {Data in Brief},
  year = {2019},
  volume = {25},
  pages = {104085},
  month = {Aug},
  doi = {10.1016/j.dib.2019.104085},
  url = {https://doi.org/10.1016/j.dib.2019.104085},
  issn = {2352-3409},
  language = {eng},
  publisher = {Elsevier},
  country = {Netherlands}
}

@inproceedings{NEURIPS2022_dbca58f3,
 author = {Landajuela, Mikel and Lee, Chak Shing and Yang, Jiachen and Glatt, Ruben and Santiago, Claudio P and Aravena, Ignacio and Mundhenk, Terrell and Mulcahy, Garrett and Petersen, Brenden K},
 booktitle = {Advances in Neural Information Processing Systems},
 editor = {S. Koyejo and S. Mohamed and A. Agarwal and D. Belgrave and K. Cho and A. Oh},
 pages = {33985--33998},
 publisher = {Curran Associates, Inc.},
 title = {A Unified Framework for Deep Symbolic Regression},
 url = {https://proceedings.neurips.cc/paper_files/paper/2022/file/dbca58f35bddc6e4003b2dd80e42f838-Paper-Conference.pdf},
 volume = {35},
 year = {2022}
}

@InProceedings{pmlr-v235-ma24m,
  title = 	 {{LLM} and Simulation as Bilevel Optimizers: A New Paradigm to Advance Physical Scientific Discovery},
  author =       {Ma, Pingchuan and Wang, Tsun-Hsuan and Guo, Minghao and Sun, Zhiqing and Tenenbaum, Joshua B. and Rus, Daniela and Gan, Chuang and Matusik, Wojciech},
  booktitle = 	 {Proceedings of the 41st International Conference on Machine Learning},
  pages = 	 {33940--33962},
  year = 	 {2024},
  editor = 	 {Salakhutdinov, Ruslan and Kolter, Zico and Heller, Katherine and Weller, Adrian and Oliver, Nuria and Scarlett, Jonathan and Berkenkamp, Felix},
  volume = 	 {235},
  series = 	 {Proceedings of Machine Learning Research},
  month = 	 {21--27 Jul},
  publisher =    {PMLR},
  url = 	 {https://proceedings.mlr.press/v235/ma24m.html}
}

@ARTICLE{797969,
  author={Zitzler, E. and Thiele, L.},
  journal={IEEE Transactions on Evolutionary Computation}, 
  title={Multiobjective evolutionary algorithms: a comparative case study and the strength Pareto approach}, 
  year={1999},
  volume={3},
  number={4},
  pages={257-271},
  doi={10.1109/4235.797969}}

@ARTICLE{6787994,
  author={Zitzler, Eckart and Deb, Kalyanmoy and Thiele, Lothar},
  journal={Evolutionary Computation}, 
  title={Comparison of Multiobjective Evolutionary Algorithms: Empirical Results}, 
  year={2000},
  volume={8},
  number={2},
  pages={173-195},
  doi={10.1162/106365600568202}}

@conference{Biggioetal21,
  title = {Neural Symbolic Regression that Scales},
  booktitle = {Proceedings of 38th International Conference on Machine Learning (ICML 2021)},
  volume = {139},
  pages = {936--945},
  series = {Proceedings of Machine Learning Research},
  editors = {Meila, Marina and Zhang, Tong},
  publisher = {PMLR},
  month = jul,
  year = {2021},
  note = {*equal contribution},
  slug = {biggioetal21},
  author = {Biggio*, L. and Bendinelli*, T. and Neitz, A. and Lucchi, A. and Parascandolo, G.},
  url = {https://proceedings.mlr.press/v139/biggio21a.html},
  month_numeric = {7}
}

@INPROCEEDINGS{130444,
  author={Koza, J.R.},
  booktitle={[1990] Proceedings of the 2nd International IEEE Conference on Tools for Artificial Intelligence}, 
  title={Genetically breeding populations of computer programs to solve problems in artificial intelligence}, 
  year={1990},
  volume={},
  number={},
  pages={819-827},
  doi={10.1109/TAI.1990.130444}}

@misc{mundhenk2021symbolicregressionneuralguidedgenetic,
      title={Symbolic Regression via Neural-Guided Genetic Programming Population Seeding}, 
      author={T. Nathan Mundhenk and Mikel Landajuela and Ruben Glatt and Claudio P. Santiago and Daniel M. Faissol and Brenden K. Petersen},
      year={2021},
      eprint={2111.00053},
      archivePrefix={arXiv},
      primaryClass={cs.NE},
      url={https://arxiv.org/abs/2111.00053}, 
}

@misc{landajuela2021improvingexplorationpolicygradient,
      title={Improving exploration in policy gradient search: Application to symbolic optimization}, 
      author={Mikel Landajuela and Brenden K. Petersen and Soo K. Kim and Claudio P. Santiago and Ruben Glatt and T. Nathan Mundhenk and Jacob F. Pettit and Daniel M. Faissol},
      year={2021},
      eprint={2107.09158},
      archivePrefix={arXiv},
      primaryClass={cs.LG},
      url={https://arxiv.org/abs/2107.09158}, 
}

@inproceedings{crochepierre:hal-03695471,
  TITLE = {{Interactive Reinforcement Learning for Symbolic Regression from Multi-Format Human-Preference Feedbacks}},
  AUTHOR = {Crochepierre, Laure and Boudjeloud-Assala, Lydia and Barbesant, Vincent},
  URL = {https://hal.science/hal-03695471},
  BOOKTITLE = {{IJCAI 2022- 31st International Joint Conference on Artificial Intelligence}},
  ADDRESS = {Vienne, Austria},
  YEAR = {2022},
  MONTH = Jul,
  HAL_ID = {hal-03695471},
  HAL_VERSION = {v1},
}

@misc{du2023discoverdeepidentificationsymbolically,
      title={DISCOVER: Deep identification of symbolically concise open-form PDEs via enhanced reinforcement-learning}, 
      author={Mengge Du and Yuntian Chen and Dongxiao Zhang},
      year={2023},
      eprint={2210.02181},
      archivePrefix={arXiv},
      primaryClass={cs.LG},
      url={https://arxiv.org/abs/2210.02181}, 
}

@misc{valipour2021symbolicgptgenerativetransformermodel,
      title={SymbolicGPT: A Generative Transformer Model for Symbolic Regression}, 
      author={Mojtaba Valipour and Bowen You and Maysum Panju and Ali Ghodsi},
      year={2021},
      eprint={2106.14131},
      archivePrefix={arXiv},
      primaryClass={cs.LG},
      url={https://arxiv.org/abs/2106.14131}, 
}

@ARTICLE{10462113,
  author={Vastl, Martin and Kulhánek, Jonáš and Kubalík, Jiří and Derner, Erik and Babuška, Robert},
  journal={IEEE Access}, 
  title={SymFormer: End-to-End Symbolic Regression Using Transformer-Based Architecture}, 
  year={2024},
  volume={12},
  number={},
  pages={37840-37849},
  doi={10.1109/ACCESS.2024.3374649}}

@inproceedings{Li2023TransformerbasedMF,
  title={Transformer-based model for symbolic regression via joint supervised learning},
  author={Wenqiang Li and Weijun Li and Linjun Sun and Min Wu and Lina Yu and Jingyi Liu and Yanjie Li and Song Tian},
  booktitle={International Conference on Learning Representations},
  year={2023},
  url={https://api.semanticscholar.org/CorpusID:259298765}
}

@inproceedings{Merler_2024,
   title={In-Context Symbolic Regression: Leveraging Large Language Models for Function Discovery},
   url={http://dx.doi.org/10.18653/v1/2024.acl-srw.49},
   DOI={10.18653/v1/2024.acl-srw.49},
   booktitle={Proceedings of the 62nd Annual Meeting of the Association for Computational Linguistics (Volume 4: Student Research Workshop)},
   publisher={Association for Computational Linguistics},
   author={Merler, Matteo and Haitsiukevich, Katsiaryna and Dainese, Nicola and Marttinen, Pekka},
   year={2024},
   pages={589–606} }

@misc{yao2025multiobjectiveevolutionheuristicusing,
      title={Multi-objective Evolution of Heuristic Using Large Language Model}, 
      author={Shunyu Yao and Fei Liu and Xi Lin and Zhichao Lu and Zhenkun Wang and Qingfu Zhang},
      year={2025},
      eprint={2409.16867},
      archivePrefix={arXiv},
      primaryClass={cs.AI},
      url={https://arxiv.org/abs/2409.16867}, 
}

@inproceedings{
zhang2025ragsr,
title={{RAG}-{SR}: Retrieval-Augmented Generation for Neural Symbolic Regression},
author={Hengzhe Zhang and Qi Chen and Bing XUE and Wolfgang Banzhaf and Mengjie Zhang},
booktitle={The Thirteenth International Conference on Learning Representations},
year={2025},
url={https://openreview.net/forum?id=NdHka08uWn}
}

@article{https://doi.org/10.1111/j.1475-3995.2011.00840.x,
author = {Lust, Thibaut and Teghem, Jacques},
title = {The multiobjective multidimensional knapsack problem: a survey and a new approach},
journal = {International Transactions in Operational Research},
volume = {19},
number = {4},
pages = {495-520},
doi = {https://doi.org/10.1111/j.1475-3995.2011.00840.x},
url = {https://onlinelibrary.wiley.com/doi/abs/10.1111/j.1475-3995.2011.00840.x},
eprint = {https://onlinelibrary.wiley.com/doi/pdf/10.1111/j.1475-3995.2011.00840.x},
year = {2012}
}

@misc{lin2022paretosetlearningneural,
      title={Pareto Set Learning for Neural Multi-objective Combinatorial Optimization}, 
      author={Xi Lin and Zhiyuan Yang and Qingfu Zhang},
      year={2022},
      eprint={2203.15386},
      archivePrefix={arXiv},
      primaryClass={cs.LG},
      url={https://arxiv.org/abs/2203.15386}, 
}

@article{andersson2000survey,
  title={A survey of multiobjective optimization in engineering design},
  author={Andersson, Johan},
  journal={Department of Mechanical Engineering, Linktjping University. Sweden},
  pages={38},
  year={2000}
}

@article{cerda2022applications,
  title={Applications of multi-objective optimization to industrial processes: a literature review},
  author={Cerda-Flores, Sandra C and Rojas-Punzo, Arturo A and N{\'a}poles-Rivera, Fabricio},
  journal={Processes},
  volume={10},
  number={1},
  pages={133},
  year={2022},
  publisher={MDPI}
}

@article{fromer2023computer,
  title={Computer-aided multi-objective optimization in small molecule discovery},
  author={Fromer, Jenna C and Coley, Connor W},
  journal={Patterns},
  volume={4},
  number={2},
  year={2023},
  publisher={Elsevier}
}

@article{kubalik2021multi,
  title={Multi-objective symbolic regression for physics-aware dynamic modeling},
  author={Kubal{\'\i}k, Ji{\v{r}}{\'\i} and Derner, Erik and Babu{\v{s}}ka, Robert},
  journal={Expert Systems with Applications},
  volume={182},
  pages={115210},
  year={2021},
  publisher={Elsevier}
}

@inproceedings{kommenda2015complexity,
  title={Complexity measures for multi-objective symbolic regression},
  author={Kommenda, Michael and Beham, Andreas and Affenzeller, Michael and Kronberger, Gabriel},
  booktitle={International Conference on Computer Aided Systems Theory},
  pages={409--416},
  year={2015},
  organization={Springer}
}

@article{ruan2020taiji,
  title={Taiji program: Gravitational-wave sources},
  author={Ruan, Wen-Hong and Guo, Zong-Kuan and Cai, Rong-Gen and Zhang, Yuan-Zhong},
  journal={International Journal of Modern Physics A},
  volume={35},
  number={17},
  pages={2050075},
  year={2020},
  publisher={World Scientific}
}

@article{amaro2023astrophysics,
  title={Astrophysics with the laser interferometer space antenna},
  author={Amaro-Seoane, Pau and Andrews, Jeff and Arca Sedda, Manuel and Askar, Abbas and Baghi, Quentin and Balasov, Razvan and Bartos, Imre and Bavera, Simone S and Bellovary, Jillian and Berry, Christopher PL and others},
  journal={Living Reviews in Relativity},
  volume={26},
  number={1},
  pages={2},
  year={2023},
  publisher={Springer}
}

@article{katz2021fast,
  title={Fast extreme-mass-ratio-inspiral waveforms: New tools for millihertz gravitational-wave data analysis},
  author={Katz, Michael L and Chua, Alvin JK and Speri, Lorenzo and Warburton, Niels and Hughes, Scott A},
  journal={Physical Review D},
  volume={104},
  number={6},
  pages={064047},
  year={2021},
  publisher={APS}
}

@ARTICLE{2018CQGra..35n4003V,
       author = {{van de Meent}, Maarten and {Warburton}, Niels},
        title = "{Fast self-forced inspirals}",
      journal = {Classical and Quantum Gravity},
         year = 2018,
        month = jul,
       volume = {35},
       number = {14},
          eid = {144003},
        pages = {144003},
          doi = {10.1088/1361-6382/aac8ce},
archivePrefix = {arXiv},
       eprint = {1802.05281},
 primaryClass = {gr-qc},
       adsurl = {https://ui.adsabs.harvard.edu/abs/2018CQGra..35n4003V}
}

@article{PhysRevResearch.3.043101,
  title     = {Learning orbital dynamics of binary black hole systems from gravitational wave measurements},
  author    = {Keith, Brendan and Khadse, Akshay and Field, Scott E.},
  journal   = {Phys. Rev. Research},
  volume    = {3},
  issue     = {4},
  pages     = {043101},
  year      = {2021},
  month     = {Nov},
  publisher = {American Physical Society},
  doi       = {10.1103/PhysRevResearch.3.043101},
  url       = {https://link.aps.org/doi/10.1103/PhysRevResearch.3.043101}
}
\newpage
\appendix
\onecolumn

\section*{\textbf{Appendix}}

\section{\textbf{Related Work}}

\subsection{Symbolic Regression}

Traditional SR methods mainly rely on evolutionary algorithms, reinforcement learning~\citep{petersen2021deepsymbolicregressionrecovering}, and Transformers~\citep{Biggioetal21}. For instance, genetic programming~\citep{130444} formulates equation discovery as an evolutionary search over tree-based representations, refining structures via mutation and crossover. Reinforcement learning-based symbolic regression was first introduced by Petersen et al.~\citep{petersen2021deepsymbolicregressionrecovering} and has since developed into various policy-optimization frameworks~\citep{mundhenk2021symbolicregressionneuralguidedgenetic,landajuela2021improvingexplorationpolicygradient,crochepierre:hal-03695471,du2023discoverdeepidentificationsymbolically}. More recently, Transformer-based models~\citep{valipour2021symbolicgptgenerativetransformermodel,10462113,kamienny2022endtoendsymbolicregressiontransformers,Li2023TransformerbasedMF,zhang2025ragsr} have been applied to SR, leveraging large-scale pretraining to improve equation generation.
However, these models typically lack mechanisms to incorporate physical priors.

With advances in natural language processing, LLM-based SR methods such as LLM-SR~\citep{shojaee2025llmsrscientificequationdiscovery}, LaSR~\citep{grayeli2024symbolicregressionlearnedconcept}, and ICSR~\citep{Merler_2024} have emerged. LLM-SR utilizes scientific priors encoded in LLMs to generate plausible equation forms, followed by data-driven parameter fitting. LaSR introduces abstract concept generation to guide hypothesis construction, while ICSR formats training data as in-context prompts to induce function generation. Yet these approaches still rely heavily on pre-trained knowledge and lack explicit modeling of variable relationships or structured reasoning over data, which limits both their search efficiency and generalization ability.

\subsection{Multi-objective Optimization}
Multi-objective optimization is a fundamental research direction in the optimization community and has been widely applied in areas such as combinatorial optimization~\citep{https://doi.org/10.1111/j.1475-3995.2011.00840.x,lin2022paretosetlearningneural,yao2025multiobjectiveevolutionheuristicusing} and industrial design~\citep{andersson2000survey,cerda2022applications,fromer2023computer}. 
In the context of SR, Kommenda et al. systematically compared various complexity measures for multi-objective SR and proposed a new metric that preserves semantic information while improving search efficiency~\citep{kommenda2015complexity}. 
Kubalík et al. introduced a physics-aware multi-objective approach~\citep{kubalik2021multi} that jointly optimizes model accuracy and physical consistency, enhancing interpretability and reliability. 
MOT-SR is the first framework to integrate LLMs into multi-objective SR, enabling data-driven equation discovery with improved accuracy, simplicity, and generalization.

\section{\textbf{Algorithmic Pseudocode for MOT-SR}}

\begin{algorithm}[H]
\caption{MOT-SR}
\label{alg:main}
\begin{algorithmic}[1]
\State \textbf{Input:} Training set $\boldsymbol{D}$; Maximum iterations $\boldsymbol{T}$; $\boldsymbol{n}$ samples per iteration; Initial population $\boldsymbol{\mathcal{P}_0}$ (optional); Meta Strategy Generator $\boldsymbol{\pi_\text{stg}}$; Equation Generator $\boldsymbol{\pi_\text{eq}}$; \textbf{Toolbox}: Data analysis toolset
\State \textbf{Output:} Approximate Pareto front $\boldsymbol{\mathcal{P}^*}$
\State Initialize $\boldsymbol{\mathcal{P}_0}$
\For{$\boldsymbol{t} = 1, \ldots, T$}
    \State $\boldsymbol{Res} \gets \text{EvaluateEquationResiduals}(\boldsymbol{\mathcal{P}_{t-1}^*}, \boldsymbol{D})$
    \State $\boldsymbol{\mathcal{R}_{\text{varrel}}} \gets \boldsymbol{\pi_\text{stg}}.\text{DataAnalysis}(\textbf{Toolbox}, \boldsymbol{Res},\boldsymbol{\mathcal{P}_{t-1}^*})$
    \State $\boldsymbol{\mathcal{R}_{\text{struct}}} \gets \boldsymbol{\pi_\text{stg}}.\text{GenerateStructuralPrompts}(\boldsymbol{\mathcal{P}_{t-1}^*})$
    \State $\boldsymbol{\mathcal{S}_t} \gets (\boldsymbol{\mathcal{R}_{\text{varrel}}}, \boldsymbol{\mathcal{R}_{\text{struct}}})$
    \State $\boldsymbol{\mathcal{P}}_{\text{parent}} \gets \text{ParentSelection}(\boldsymbol{\mathcal{P}_{t-1}^*})$
    \State $\boldsymbol{\mathcal{P}_t} \gets \boldsymbol{\mathcal{P}_{t-1}^*}$
    \For{$\boldsymbol{i} = 1, \ldots, n$}
        \State $\boldsymbol{f} \sim \boldsymbol{\pi}_{\text{eq}}(\boldsymbol{\mathcal{S}_t} , \boldsymbol{\mathcal{P}}_{\text{parent}})$
        \State $\boldsymbol{score} \gets \text{MultiObjectiveEvaluation}(\boldsymbol{f}, \boldsymbol{D})$
        \State $\boldsymbol{\mathcal{P}_t} \gets \boldsymbol{\mathcal{P}_t} \cup \{\boldsymbol{f}, \boldsymbol{score}\}$
    \EndFor
    \State $\boldsymbol{\mathcal{P}_t^*} \gets \text{PopulationManagement}(\boldsymbol{\mathcal{P}_t})$
\EndFor
\State $\boldsymbol{\mathcal{P}^*} \gets \boldsymbol{\mathcal{P}_T}$
\end{algorithmic}
\end{algorithm}

\section{\textbf{Data Partition and Thresholding Protocol for ID/OOD Evaluation}}
\label{appendix:id-ood}

\subsection{Motivation and Protocol Definition}
In scientific discovery, a valid symbolic model must not only fit the observed data (interpolation) but also capture the underlying physical laws to predict system behaviors in unobserved regions (extrapolation). To rigorously evaluate this capability, we adopt a \textbf{percentile-based masking strategy} to partition the global data distribution $\mathcal{D}_{\mathrm{total}}$ into In-Domain (ID) and Out-of-Domain (OOD) subsets.

Unlike random splitting, which results in independent and identically distributed (i.i.d.) subsets, our protocol enforces a strictly defined \textbf{distributional shift} between training (ID) and testing (OOD), thereby assessing the model's robustness against extrapolation risks.

\subsection{Partition Methodology}
\label{app:data_partition_method}
For each task, let $\mathbf{x}\in\mathbb{R}^d$ denote the input variables and $y$ the target. The protocol is defined based on \textit{per-dimension} marginal percentiles of the training inputs in $D_{\mathrm{train}}$.

\paragraph{Percentile candidates.}
We iterate over percentile values
\begin{equation}
p \in \{10,11,\dots,49\}.
\end{equation}

\paragraph{Per-dimension central intervals.}
For each input dimension $j\in\{1,\dots,d\}$, we compute lower and upper bounds from the training inputs:
\begin{equation}
\ell_j(p) = \mathrm{percentile}(X_j,p),\qquad
u_j(p) = \mathrm{percentile}(X_j,100-p),
\end{equation}
where $X_j$ denotes the set of training values of the $j$-th coordinate.

\paragraph{Training-derived ID region as a Cartesian product.}
We define the training-derived ID subset as the Cartesian product of these per-dimension central intervals:
\begin{equation}
D_{\mathrm{train}}^{\mathrm{ID}}(p)
=
\Big\{(\mathbf{x},y)\in D_{\mathrm{train}}:\ \ell_j(p)\le x_j\le u_j(p),\ \forall j\Big\}.
\label{eq:train_id_cartesian}
\end{equation}
Geometrically, $D_{\mathrm{train}}^{\mathrm{ID}}(p)$ forms a central hyper-rectangle in the input space (Figure~2).

\paragraph{Training-derived OOD region as the complement within $D_{\mathrm{train}}$.}
The training-derived OOD subset is defined as the spatial complement of the ID region \textit{within the training set}:
\begin{equation}
D_{\mathrm{train}}^{\mathrm{OOD}}(p)
=
D_{\mathrm{train}}\setminus D_{\mathrm{train}}^{\mathrm{ID}}(p).
\label{eq:train_ood_complement}
\end{equation}
This subset corresponds to peripheral regions along at least one dimension, inducing an explicit distributional shift relative to $D_{\mathrm{train}}^{\mathrm{ID}}(p)$.

\paragraph{Selecting the split percentile.}
Among candidate percentiles, we select $p^\star$ to balance the sample counts in the two regions:
\begin{equation}
p^\star
=
\arg\max_{p\in\{10,\dots,49\}}
\min\Big(\big|D_{\mathrm{train}}^{\mathrm{ID}}(p)\big|,\ \big|D_{\mathrm{train}}^{\mathrm{OOD}}(p)\big|\Big),
\label{eq:pstar_balance}
\end{equation}
which encourages a non-trivial OOD subset while retaining sufficient ID samples for stable fitting.

Finally, we set
\begin{equation}
D_{\mathrm{train}}^{\mathrm{ID}} := D_{\mathrm{train}}^{\mathrm{ID}}(p^\star),\qquad
D_{\mathrm{train}}^{\mathrm{OOD}} := D_{\mathrm{train}}^{\mathrm{OOD}}(p^\star).
\end{equation}

\subsection{Usage in MOT-SR and No-Leakage Guarantee}
\label{app:data_partition_usage}
MOT-SR uses the above training-derived split as follows:
\begin{itemize}
  \item \textbf{Parameter fitting (discovery loop).} Equation constants are fitted \textit{only} on $D_{\mathrm{train}}^{\mathrm{ID}}$.
  \item \textbf{Generalization feedback (discovery loop).} Extrapolation feedback is computed on $D_{\mathrm{train}}^{\mathrm{OOD}}$ to guide multi-objective selection, but this subset is \textit{never} used for parameter fitting.
  \item \textbf{Final evaluation (locked test).} Benchmark test sets (e.g., $D_{\mathrm{test}}^{\mathrm{ID}}$ and $D_{\mathrm{test}}^{\mathrm{OOD}}$) are held out and used \textit{only} for final reporting.
\end{itemize}
Therefore, the ``OOD'' objective used inside the discovery loop refers strictly to the training-derived region $D_{\mathrm{train}}^{\mathrm{OOD}}$ (Eq.~\ref{eq:train_ood_complement}), and does not involve any held-out test data.

\subsection{Dynamic NMSE Thresholding for Pre-Filtering}
\label{app:thresholding}
To avoid flooding the Pareto selection with clearly underfitting candidates, we apply a loose \textit{upper-bound} NMSE threshold in each generation. For every candidate $i$, we compute its worst-case training error across the two training-derived regions:
\begin{equation}
e_i = \max\big(\mathrm{NMSE}_{\mathrm{ID}}^{(i)},\ \mathrm{NMSE}_{\mathrm{OOD}}^{(i)}\big),
\end{equation}
and define the current best worst-case error in the population as
\begin{equation}
e^\star = \min_i e_i.
\end{equation}
We then construct a dynamic upper bound
\begin{equation}
T = 10^{0.5\,\log_{10}(e^\star)} = \sqrt{e^\star},
\end{equation}
so that $T$ is on the order of $\sqrt{e^\star}$ in log-scale. Any candidate satisfying
\begin{equation}
\max\big(\mathrm{NMSE}_{\mathrm{ID}}^{(i)},\ \mathrm{NMSE}_{\mathrm{OOD}}^{(i)}\big) > T
\end{equation}
is treated as clearly underfitting; in implementation we assign it a very large objective value so it cannot enter the Pareto front. This step does not penalize low-NMSE solutions, and only removes candidates whose errors are orders of magnitude worse than the current best.

\section{\textbf{MOT-SR Configuration and Language Model Details}}

\label{appendix:MOT-SR}

\subsection{MOT-SR Configuration}

We implement \textsc{MOT-SR} with both open-source and commercial LLM backbones: \textbf{LLaMA-3.1-8B-Instruct} and \textbf{GPT-4o-mini}. The \textbf{LLaMA-3.1} model is quantized and deployed locally on NVIDIA H100 80GB GPUs for efficient inference. And \textbf{GPT-4o-mini} is accessed via the OpenAI API, providing high-quality reasoning without requiring local resources. Following the \textsc{LLM-SRBench} protocol, the maximum number of optimization iterations is set to 2000 for the four main benchmark datasets, and 1000 for the LSR-Synth–Chemistry dataset. In each iteration, the Meta Strategy Generator selects 3 tools from the tool set to analyze variable-level relationships and generates a natural language strategy prompt based on both data characteristics and symbolic structural patterns. Both the \textbf{Meta Strategy Generator} and \textbf{Equation Generator} in \textsc{MOT-SR} utilize LLM decoding with top-$k = 30$, top-$p = 0.3$, and temperature 0.6. In each iteration, the Equation Generator produces 4 candidate expressions. During Pareto frontier construction, we apply an NMSE-based filtering mechanism to eliminate trivial expressions: for each candidate, we compute the base-10 logarithm of its minimal NMSE (across ID and OOD), and discard it if it exceeds \(10^{\lceil 0.5 \cdot \min\log_{10}\text{NMSE} \rceil}\), where the minimum is taken over the current candidate pool.

\subsection{Large Language Models}

\textsc{MOT-SR} employs two backbone LLMs: \texttt{LLaMA-3.1-8B-Instruct} and \texttt{GPT-4o-mini}, covering both open-source and API-accessible commercial models. The LLaMA-3.1 model is locally quantized to 4-bit precision and deployed on NVIDIA H100 80GB GPUs, requiring approximately 8GB of VRAM during inference, thus allowing execution on consumer-grade hardware with appropriate quantization. The GPT-4o-mini model is accessed via OpenAI’s API, offering high-quality reasoning with minimal deployment overhead.

\section{Symbolic Equation Discovery for EMRI Orbital Evolution}
\label{app:emri}

\subsection{EMRI Orbital-Evolution Dataset}
\label{app:emri_dataset}

Here, we provide the preprocessing
details used to construct the symbolic regression dataset.

Each case is simulated using both FSI and PN5. The resulting
evolution rates are aligned at the same orbital states
\((p,e,\eta)\) before the residual labels are computed. If the
output points of the two models do not coincide, they are aligned
through interpolation or recomputation.

The original data retain time and evolution rates in seconds. To
express the rates with respect to the dimensionless time \(t/M\),
we define
\begin{equation}
    M_{\mathrm{sec}}=\frac{GM}{c^3}.
\end{equation}
For \(X\in\{\mathrm{FSI},\mathrm{PN5}\}\), the dimensionless
semi-latus-rectum evolution rate is
\begin{equation}
    \dot p_X
    =
    M_{\mathrm{sec}}\dot p^{\,\mathrm{sec}}_X,
\end{equation}
where
\(\dot p^{\,\mathrm{sec}}_X=dp/dt_{\mathrm{sec}}\).
The symbolic regression target is then constructed as
\begin{equation}
    \Delta\dot p
    =
    M_{\mathrm{sec}}
    \left(
        \dot p^{\,\mathrm{sec}}_{\mathrm{FSI}}
        -
        \dot p^{\,\mathrm{sec}}_{\mathrm{PN5}}
    \right).
    \label{eq:emri_residual_target}
\end{equation}
The resulting regression samples have the form
\begin{equation}
    (p,e,\eta)\longmapsto\Delta\dot p,
    \qquad
    \eta=\frac{\mu}{M}.
\end{equation}

Among the 100 generated cases, Cases 89--100 contain only one or
two valid states and are removed before evaluation. The remaining
88 cases are split at the trajectory level: 58 complete
trajectories are used for equation discovery and parameter
fitting, while 30 complete trajectories are held out for testing.
No trajectory contributes samples to both subsets. The equation
structure, fitted constants, and candidate selection are fixed
before evaluation on the held-out cases.

Only \(p\), \(e\), and \(\eta\) are used as inputs to the symbolic
regression model. The masses \(M\) and \(\mu\) determine the mass
ratio \(\eta\), while \(M\) additionally defines the time-conversion
factor \(M_{\mathrm{sec}}\). Extrinsic parameters, including
redshift, sky location, source orientation, distance, and initial
phases, are excluded because they do not determine the
orbit-averaged evolution rate considered here.

\subsection{Formulating EMRI Orbital Evolution as Symbolic Regression}
\label{app:emri_sr_casting}

\paragraph{Predicting a local correction instead of a trajectory.}
Directly regressing the full trajectory \(p(t)\) would entangle the
local modeling error with numerical integration and the selected
initial conditions. We instead search for a local and
configuration-independent symbolic residual
\begin{equation}
    f_{\boldsymbol{\theta}}:
    (p,e,\eta)
    \longmapsto
    \Delta\dot p,
\end{equation}
where \(f_{\boldsymbol{\theta}}\) is an explicit symbolic expression
and \(\boldsymbol{\theta}\) denotes its fitted constants. The
symbolic hypothesis is trained to approximate the rate discrepancy
in Equation~\eqref{eq:emri_residual_target}:
\begin{equation}
    f_{\boldsymbol{\theta}}(p,e,\eta)
    \simeq
    \dot p_{\mathrm{FSI}}(p,e,\eta)
    -
    \dot p_{\mathrm{PN5}}(p,e,\eta).
\end{equation}
This formulation asks the symbolic model to identify the systematic
part of the approximation error shared across EMRI configurations,
rather than memorizing a separate time-dependent correction for
each trajectory.

\paragraph{Corrected orbital-evolution rate.}
Given a candidate symbolic expression, the corrected PN5 evolution
rate is reconstructed as
\begin{equation}
    \dot p_{\mathrm{corr}}(p,e,\eta)
    =
    \dot p_{\mathrm{PN5}}(p,e,\eta)
    +
    f_{\boldsymbol{\theta}}(p,e,\eta).
    \label{eq:emri_corrected_rate}
\end{equation}
When the orbital evolution is integrated in physical time, the
corresponding rate is
\begin{equation}
    \dot p^{\,\mathrm{sec}}_{\mathrm{corr}}
    =
    \frac{
    \dot p_{\mathrm{PN5}}
    +
    f_{\boldsymbol{\theta}}(p,e,\eta)
    }{
    M_{\mathrm{sec}}
    }.
\end{equation}
The symbolic correction is therefore repeatedly evaluated at the
current orbital state during numerical integration, rather than
being applied once as a post-processing adjustment.

\paragraph{Symbolic fitting and multi-objective evaluation.}
For a set of residual samples \(\mathcal D\), the point-wise
prediction error of a candidate expression is measured by
\begin{equation}
    \operatorname{NMSE}_{\Delta\dot p}
    \left(
    f_{\boldsymbol{\theta}},\mathcal D
    \right)
    =
    \frac{
    \sum_{i}
    \left[
    f_{\boldsymbol{\theta}}(p_i,e_i,\eta_i)
    -
    \Delta\dot p_i
    \right]^2
    }{
    \sum_{i}
    \left[
    \Delta\dot p_i
    -
    \overline{\Delta\dot p}_{\mathcal D}
    \right]^2
    }.
\end{equation}
Within the 58-trajectory discovery set, MOT-SR applies the same
training-derived ID/OOD partition used in the general framework.
The free constants are optimized using the ID subset, while the OOD
subset supplies extrapolation feedback. Candidate expressions are
then jointly evaluated according to
\begin{equation}
    \left(
    \operatorname{NMSE}_{\mathrm{ID}},
    \operatorname{NMSE}_{\mathrm{OOD}},
    \operatorname{ASTLen}
    \right),
\end{equation}
so that the search favors expressions that balance local fitting
accuracy, cross-region generalization, and structural compactness.

\paragraph{A posteriori trajectory evaluation.}
A low point-wise residual error does not necessarily guarantee an
accurate long-term orbit, because small systematic errors can
accumulate when the learned expression is repeatedly evaluated
inside the dynamical solver. We therefore perform an
\emph{a posteriori} evaluation by inserting
Equation~\eqref{eq:emri_corrected_rate} into the orbital-evolution
procedure and integrating from the initial conditions of each
complete trajectory. This produces the corrected trajectory
\(p_{\mathrm{corr}}(t)\).

For each EMRI case, trajectory-level accuracy is measured against
the FSI reference using
\begin{equation}
    \operatorname{NMSE}_{p(t)}
    =
    \frac{
    \sum_{k}
    \left[
    p_{\mathrm{corr}}(t_k)
    -
    p_{\mathrm{FSI}}(t_k)
    \right]^2
    }{
    \sum_{k}
    \left[
    p_{\mathrm{FSI}}(t_k)
    -
    \overline{p}_{\mathrm{FSI}}
    \right]^2
    }.
    \label{eq:emri_trajectory_nmse}
\end{equation}
The equation structure, fitted constants, and final candidate
selection are fixed before integration on the 30 held-out
configurations. Hence, Equation~\eqref{eq:emri_trajectory_nmse}
evaluates whether a single symbolic correction discovered from the
58 discovery cases remains reliable under repeated integration
across unseen EMRI systems and different trajectory lengths.

\section{\textbf{Datasets}}
\label{appendix:datasets}
\subsection{Nonlinear Oscillator}

Oscillatory systems with nonlinear damping are foundational in physics and engineering for modeling the motion of objects subjected to restoring and dissipative forces. These systems are governed by second-order differential equations of the form \(\ddot{x} + f(t, x, \dot{x}) = 0\), where the nonlinear function \(f\) encapsulates dynamic interactions among position, velocity, and possibly time.

To evaluate a model's ability to recover such complex dynamics, two synthetic oscillator tasks are used. The first system is defined by:
\[
\dot{v} = F \sin(\omega x) - \alpha v^3 - \beta x^3 - \gamma x v - x \cos(x)
\quad \scriptsize
(F = 0.8,\ \alpha = 0.5,\ \beta = 0.2,\ \gamma = 0.5,\ \omega = 1.0)
\]
The second system follows:
\[
\dot{v} = F \sin(\omega t) - \alpha v^3 - \beta x v - \delta x \exp(\gamma x)
\quad \scriptsize
(F = 0.3,\ \alpha = 0.5,\ \beta = 1.0,\ \delta = 5.0,\ \gamma = 0.5,\ \omega = 1.0)
\]
with initial conditions \(x = 0.5\), \(v = \dot{x} = 0.5\), and simulation time \(t \in [0, 50]\). These equations exhibit rich nonlinear structures and variable couplings, making them ideal benchmarks for testing symbolic reasoning and generalization beyond simple oscillatory behavior.

\subsection{Bacterial Growth}

Accurately capturing the dynamics of E. coli proliferation under varying environmental conditions is of critical importance in areas such as biotechnology and microbiological risk assessment. This benchmark reflects realistic yet challenging biological modeling, where growth depends on multiple interacting factors.

The dataset models population change via:
\[
\frac{dB}{dt} = f_B(B) \cdot f_S(S) \cdot f_T(T) \cdot f_{\text{pH}}(\text{pH})
\]
where \(B\) is bacterial density, \(S\) is nutrient concentration, \(T\) is temperature, and pH denotes acidity. Each term accounts for a distinct physiological influence on growth.

The explicit expression used is:
\[
\frac{dB}{dt} = \mu_{\text{max}} B \left( \frac{S}{K_S + S} \right)
\left(\frac{\tanh{k(T - x_0)}}{1 + c(T - x_{\text{decay}})^4}\right)
\exp\left(-|\text{pH} - \text{pH}_{\text{opt}}|\right)
\sin\left( \frac{(\text{pH} - \text{pH}_{\text{min}})\pi}{\text{pH}_{\text{max}} - \text{pH}_{\text{min}}} \right)^2
\]

This formulation introduces complex nonlinearities and multimodal interactions across environmental axes, challenging models to integrate biological structure while avoiding rote memorization.

\subsection{Material Stress Behavior}

To evaluate symbolic models under realistic experimental conditions, this benchmark focuses on the stress-strain response of Aluminum 6061-T651 under thermal influence. The dataset records tensile strength measurements at six different temperatures, ranging from room temperature to 300\textdegree C, simulating diverse material states.

In contrast to synthetic equations, this task lacks an explicit ground-truth formula, requiring data-driven inference of hidden physical laws. It serves as a test of a model’s capacity to identify empirical regularities in noisy, high-variance regimes.

A widely used approximation of this behavior is given by:
\[
\sigma = \left( A + B \varepsilon^n \right) \left( 1 - \left( \frac{T - T_r}{T_m - T_r} \right)^m \right)
\]
where \(\sigma\) denotes stress, \(\varepsilon\) is strain, \(T\) is the temperature, \(T_r\) is a reference point, and \(T_m\) is the melting point. The coefficients \(A\), \(B\), \(n\), and \(m\) are empirically determined for the alloy.

This benchmark emphasizes the need for robust symbolic reasoning in the absence of prior symbolic templates, bridging theory-driven modeling with experimental data interpretation.

\subsection{LSR-Synth–Chemistry}

The \textbf{LSR-Synth–Chemistry} dataset, introduced as part of the \textsc{LLM-SRBench} benchmark by Shojaee et al., is designed to evaluate symbolic regression models on chemically motivated yet synthetically constructed reaction kinetics. It comprises 36 differential equation discovery tasks, each modeling the time evolution of a reactant concentration \( A(t) \) using novel, data-driven expressions.

Each equation features a distinct combination of symbolic components—ranging from classic kinetic motifs (e.g., first- and second-order decay terms like \( -kA(t) \), \( -kA(t)^2 \)) to synthetic nonlinearities. These include exponential decays such as \(\exp(-k_s t)\), logarithmic forms like \(\log(A(t) + 1)\), square roots, and oscillatory terms such as \(\sin(\omega A(t))\) and \(\cos(\cdot)\). In addition, rational expressions like \(\frac{A(t)^2}{1 + \beta A(t)^4}\) introduce challenges in terms of singularity avoidance and numerical stability.

The dataset emphasizes symbolic diversity and parametric variability, making it a rigorous testbed for evaluating model generalization, interpretability, and robustness across structurally distinct formulations. Each task was carefully validated for analytical solvability, numerical stability, and scientific plausibility via expert review.

\section{\textbf{Additional Ablation Studies}}
\label{app:more-xiaorong}

\subsection{Ablation Study on Remaining Benchmarks}
\label{app:more-xiaorong1}

\begin{figure}[H]
  \centering
  \includegraphics[width=0.8\textwidth]{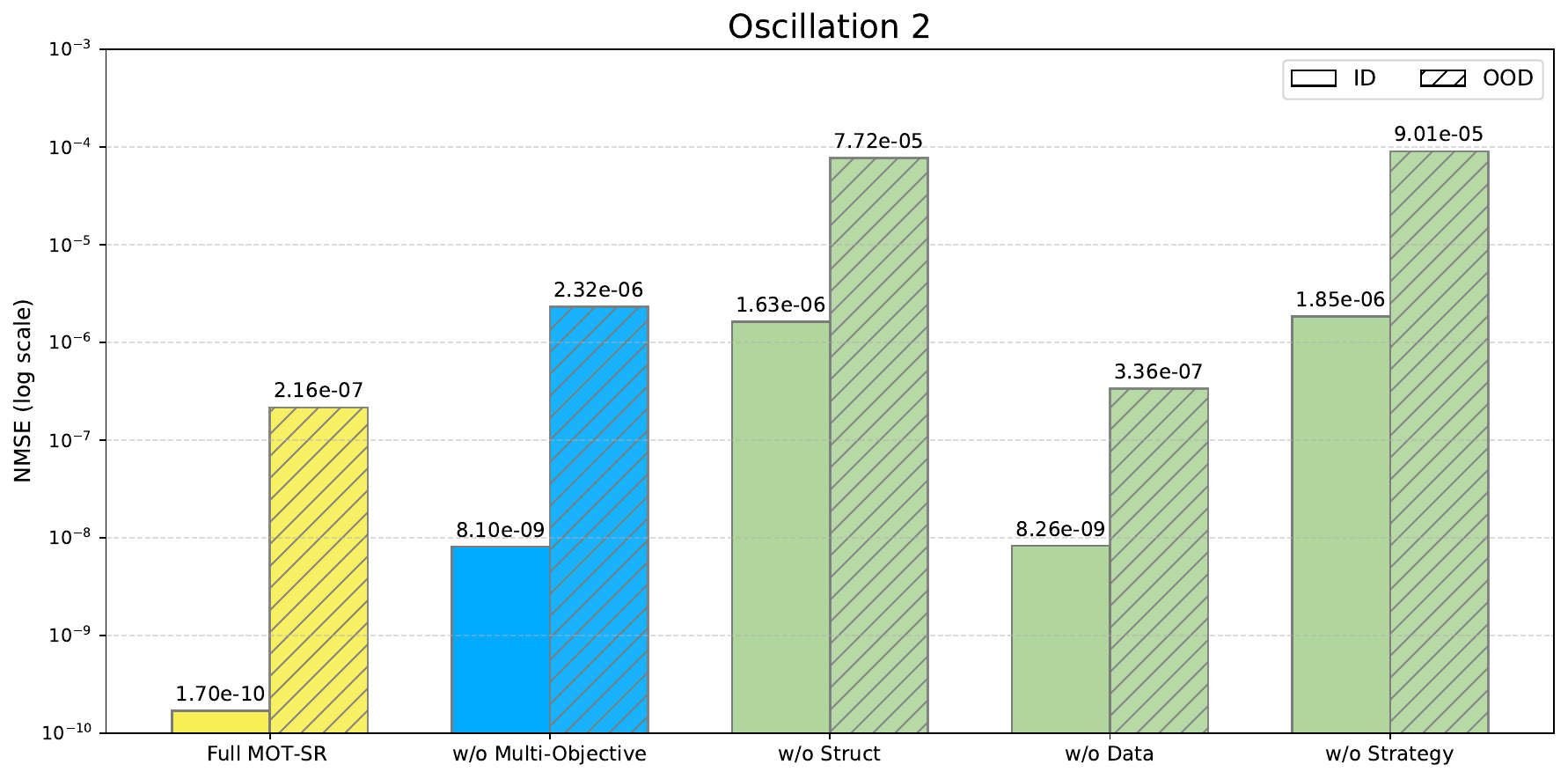}
  \caption{Ablation study on Oscillation 2.}
  \label{fig:ablation_oscillation}
\end{figure}

\begin{figure}[H]
  \centering
  \includegraphics[width=0.8\textwidth]{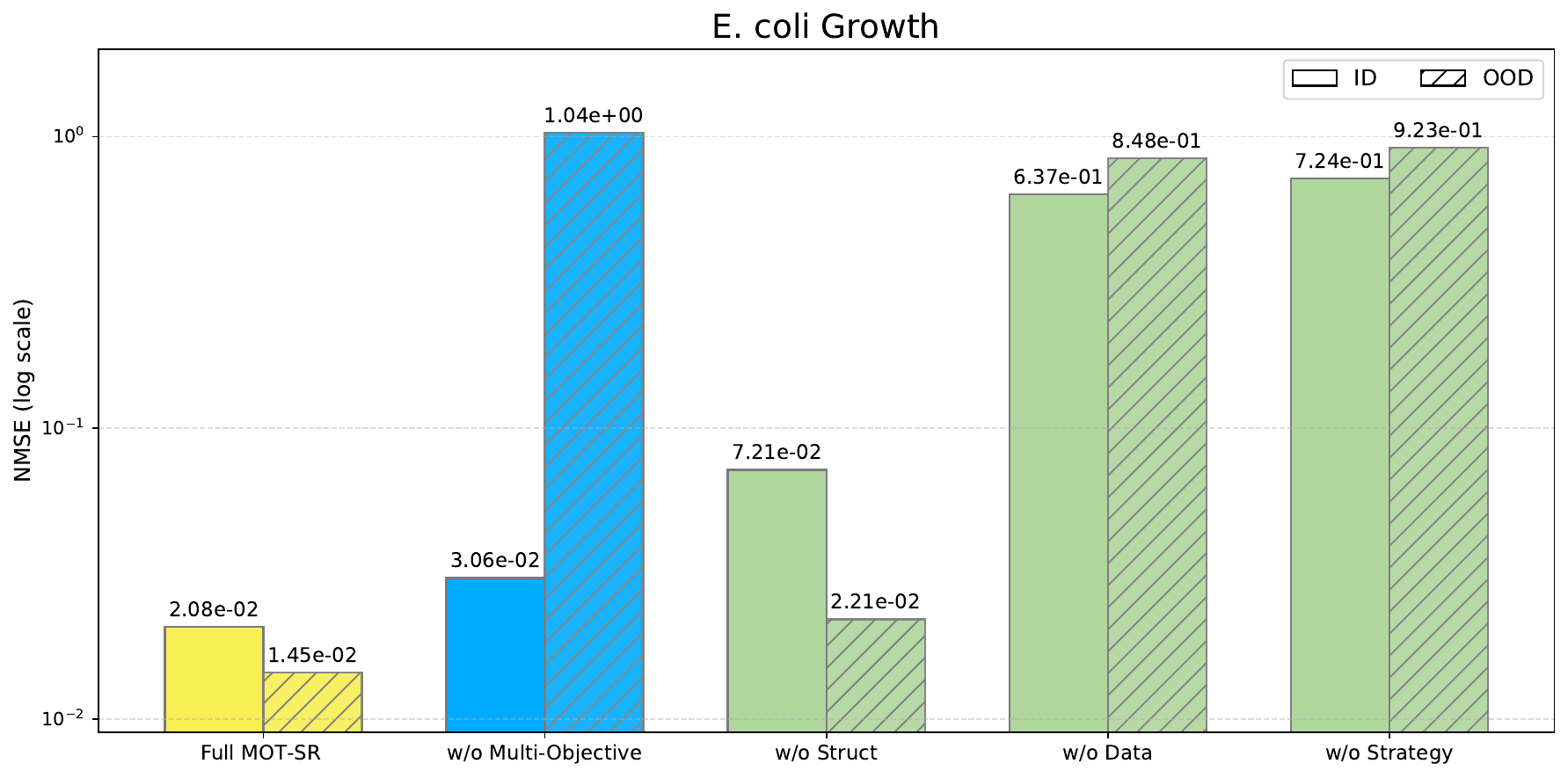}
  \caption{Ablation study on E. coli growth.}
  \label{fig:ablation_ecoli}
\end{figure}

\begin{figure}[H]
  \centering
  \includegraphics[width=0.8\textwidth]{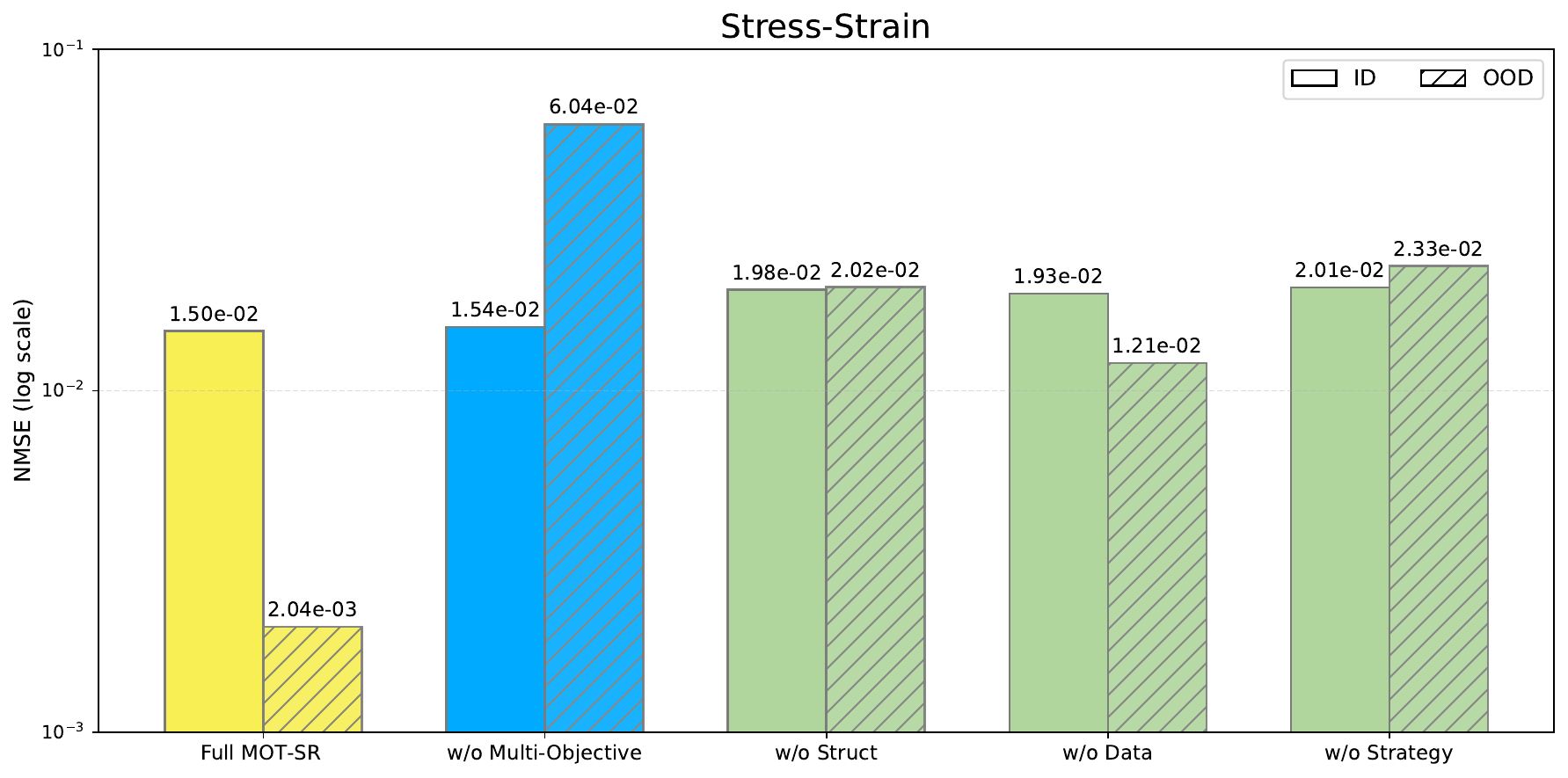}
  \caption{Ablation study on Stress-strain.}
  \label{fig:ablation_stress}
\end{figure}

We further validated the key components via ablations on the remaining three benchmarks, with the results shown in Figure~\ref{fig:ablation_oscillation} to Figure~\ref{fig:ablation_stress}. The experimental results are highly consistent with the findings presented in the main text. First, for the single-objective optimization variant (w/o MultiObj), a significant degradation in out-of-domain (OOD) generalization ability was observed across all benchmarks when compared to the full model. Although the in-domain (ID) fitting error of this variant is comparable to that of the full MOT-SR on some tasks, its performance on OOD data systematically confirms the tendency of single-objective optimization to overfit the ID data. This tendency consequently hinders the discovery of universally applicable scientific laws.

Second, the ablation experiments on the internal components of the Meta Strategy Generator show that removing either the data analysis module (w/o Data) or the structure analysis module (w/o Struct) leads to a distinct decline in model performance. More critically, the degree of performance degradation caused by removing both modules simultaneously (w/o Strategy) exceeds the impact of removing either single module. This phenomenon clearly reveals the functional complementarity and synergistic effect between the data-driven prior knowledge and the structure-driven feedback mechanism. Both are indispensable for achieving MOT-SR's final performance.

\subsection{Ablation Study on Data Analysis Tools}
\label{app:more-xiaorong2}
To evaluate the contributions of the data analysis tools in MOT-SR, we conducted a series of ablation experiments. Our framework integrates six categories of analysis tools, and we set up six control groups (w/o tool-1 to 6), each corresponding to the removal of one category of tools.

\begin{figure}[H]
  \centering
  \includegraphics[width=0.8\textwidth]{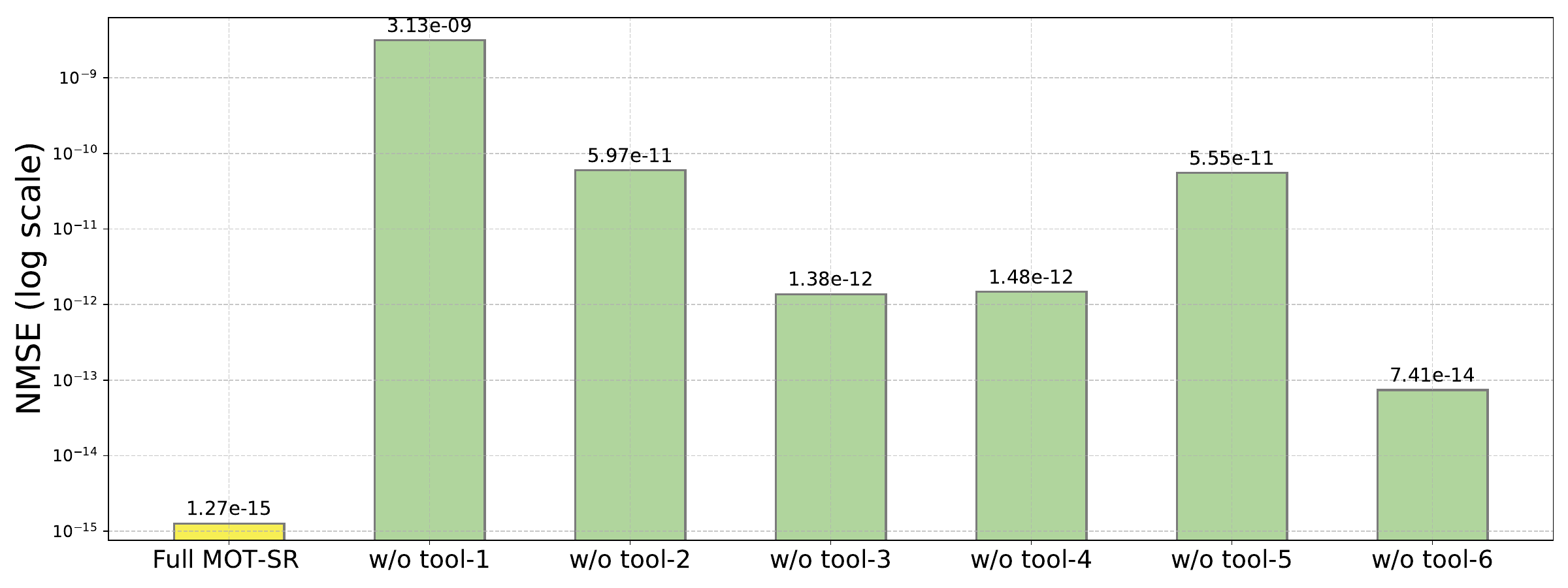}
  \caption{
  {Ablation study of the Data Analysis Tools on Oscillator 1. }
 }
  \label{fig:more_ablation_tools}
\end{figure}

As shown in the Figure~\ref{fig:more_ablation_tools}, removing any single category of analysis tools in this test leads to a decline in the model's final performance. It is noteworthy that although there is only one tool in the sixth category, its removal still resulted in a final error nearly an order of magnitude higher than that of the full MOT-SR. This result strongly demonstrates that external analysis tools can effectively enhance a large language model's insight into the underlying structure and variable relationships within data, enabling it to capture deeper trends behind the data points and thereby generate more accurate symbolic equations.

\section{\textbf{HV and IGD Definition}}
\label{appendix:paleituo}
\paragraph{F.1 HV}

The hypervolume (HV) measures the volume in the objective space that is dominated by the approximate Pareto front and bounded by a reference point. It reflects the convergence and diversity of the solutions. The HV is formally defined as follows:

\[
\text{HV}(P, r^*) = \text{VOL} \left( \bigcup_{\mathbf{v} \in P} [v_1, r_1^*] \times [v_2, r_2^*] \times \cdots \times [v_m, r_m^*] \right),
\]

where \(P\) denotes the approximate Pareto front obtained by the symbolic regression algorithm, \(\mathbf{v} = (v_1, \ldots, v_m)^\top\) represents an objective vector, \(\text{VOL}(\cdot)\) indicates the Lebesgue measure, and \(r^* = (r_1^*, \ldots, r_m^*)^\top\) is the reference point.

To eliminate the impact of varying scales and units among objectives in HV calculation, we normalize each objective to the interval \([0, 1]\). For an objective value \(f_i(x)\), the normalized value is:

\[
f_i'(x) = \frac{f_i(x) - z_i^{\text{ideal}}}{z_i^{\text{nadir}} - z_i^{\text{ideal}}},
\]

where

\[
z_i^{\text{ideal}} = \min_{\mathbf{v} \in P} v_i, \quad z_i^{\text{nadir}} = \max_{\mathbf{v} \in P} v_i
\]

After normalization, each objective is scaled to the range \([0, 1]\). Based on this, the reference point is set as \(r^* = (1.05, \ldots, 1.05)^\top\).

\paragraph{F.2 IGD}

The Inverted Generational Distance (IGD) measures both the convergence and diversity of the predicted front by computing the distance from points on the true Pareto front to their nearest counterparts in the predicted set. The formula is as follows:

\[
\text{IGD}(P, P^*) = \frac{1}{|P^*|} \sum_{\mathbf{p}^* \in P^*} \min_{\mathbf{p} \in P} d(\mathbf{p}, \mathbf{p}^*),
\]

where \(P\) is the predicted Pareto front, \(P^*\) is a set of reference points sampled from the true Pareto front, and \(d(\mathbf{p}, \mathbf{p}^*)\) represents the Euclidean distance between vectors \(\mathbf{p}\) and \(\mathbf{p}^*\) in objective space.

The computation of the Inverted Generational Distance (IGD) metric conventionally depends on the availability of a known true Pareto front, which is not always accessible in practical scenarios. In the context of symbolic regression evaluation, however, the IGD metric benefits from an inherent advantage, as the true Pareto front can be explicitly defined by the ground truth equation \(f^\text{true}\). This allows for an exact assessment of the quality of generated equations relative to the ground truth, without reliance on an approximate Pareto front. Accordingly, the IGD formulation in this study is simplified as follows:
\[
\text{IGD}(P, \{f^\text{true}\}) = \min_{f \in P} d(f,f^\text{true})
\]

Afterwards, all objectives are normalized to the \([0, 1]\) range to ensure consistency with the true objective values.

\section{\textbf{Data Analysis Tools in MOT-SR}}

\label{appendix:Tool Set}

\subsection{Linear Correlation Tools}

Linear correlation tools provide foundational insights into how variables are related through linear or approximately linear patterns. These tools are instrumental for identifying primary dependencies, detecting dominant axes of variation, and establishing structural priors for symbolic equation modeling. MOT-SR incorporates four linear-correlation-based modules:

\begin{itemize}[leftmargin=*, labelsep=0.5em]
  \item \textbf{Pearson Correlation Coefficient:} \\
  It quantifies the degree of linear association between two variables \(x\) and \(y\) using the classical Pearson correlation coefficient \(r\). The input consists of two real-valued vectors, the output includes the correlation score and the associated $p$-value indicating statistical significance. A high absolute value of \(r\) (close to 1) suggests a strong linear trend, while values near 0 indicate little to no linear dependency. This measure is especially suitable for Gaussian-like distributions and helps identify candidate variables for symbolic terms with additive or multiplicative linear effects.

  \item \textbf{Simple Linear Regression:} \\
  This module fits a univariate linear model \(y = a + bx\) using least squares estimation and returns the regression coefficients, the $R^2$ value (explained variance), and associated statistics such as $p$-values and standard errors. The inputs are one-dimensional arrays \(x\) and \(y\), and the output reflects how well a straight line fits the data. The $R^2$ value is particularly informative, indicating how much of the variance in \(y\) can be explained by \(x\). It provides a predictive perspective on the relationship beyond correlation, supporting model selection based on explanatory power.

  \item \textbf{Residual Variance:} \\
  This tool computes the variance of residuals \(y - \hat{y}\) after applying linear regression. The lower the residual variance, the better the linear model captures the relationship between \(x\) and \(y\). It takes the same inputs as Simple Linear Regression and internally reuses the linear regression output. This metric emphasizes prediction error dispersion, offering a direct measure of the noise or unmodeled nonlinear structure in the data. It is particularly valuable for pruning noisy or unstable variables from candidate model terms.

  \item \textbf{PCA Explained Variance:} \\
  Principal Component Analysis (PCA) is applied to the joint space of \(x\) and \(y\), and the proportion of variance explained by the first principal component is reported. The input is a two-dimensional matrix formed by stacking \(x\) and \(y\), the output is a scalar value indicating the strength of the shared linear structure. This approach is robust to scaling and rotation and can capture the dominant direction of variation when the relationship is more general than univariate regression. It aids in identifying latent linear couplings and guides the selection of structurally informative variables.

\end{itemize}

\subsection{Nonlinear Dependency Tools}

Nonlinear dependency tools are employed to capture complex, non-monotonic interactions between variables, which are essential for accurately modeling nonlinear systems. Unlike linear tools, these methods can reveal relationships that are not adequately described by simple linear transformations. MOT-SR incorporates three nonlinear-dependency-based modules:

\begin{itemize}[leftmargin=*, labelsep=0.5em]
  \item \textbf{Spearman Rank Correlation:} This tool measures the strength and direction of monotonic relationships between two variables using their rank values. It computes Spearman’s $\rho$, a non-parametric counterpart to Pearson’s $r$, by evaluating how well the relationship between $x$ and $y$ can be described by a monotonic function. The input consists of two numerical vectors, the output includes the correlation coefficient and its $p$-value. This method is particularly effective when data exhibit nonlinear but monotonic trends, such as saturating growth or sigmoid-like behavior.

  \item \textbf{Mutual Information:} This tool quantifies the total amount of information shared between two variables, regardless of the specific functional form of their relationship. It estimates mutual information by discretizing the input variables $x$ and $y$ into bins and calculating the joint entropy. The output is a scalar metric representing dependency strength. Unlike correlation coefficients, mutual information can capture both linear and highly nonlinear dependencies, making it suitable for detecting complex statistical associations in symbolic modeling.

  \item \textbf{Mutual Information Regression Score:} This variant employs the \texttt{scikit-learn} implementation of mutual information to assess the relevance of $x$ in predicting $y$ using a regression-based formulation. The method internally estimates how much knowing $x$ reduces uncertainty about $y$. It accepts one-dimensional arrays as input and outputs a scalar score. This score is robust to arbitrary nonlinearities and discontinuities, making it valuable for feature selection in nonlinear symbolic regression tasks.
\end{itemize}

\subsection{Time-Frequency Analysis Tools}

Time-frequency analysis tools are essential for detecting periodicities, oscillations, and transient dynamics that are characteristic of nonlinear and multi-scale systems. These tools enable symbolic regression models to incorporate periodic or time-varying terms where appropriate. MOT-SR employs two such modules:

\begin{itemize}[leftmargin=*, labelsep=0.5em]
  \item \textbf{Fast Fourier Transform Frequency Difference:} This module applies the discrete Fourier transform to the input signals $x$ and $y$ to extract their respective power spectra. It identifies the dominant frequency component for each variable and returns the absolute difference between them. The input consists of two one-dimensional arrays representing time-series data, and the output includes the dominant frequencies of both variables as well as their difference. This tool captures global periodic patterns and is useful for detecting whether the signals share synchronized or harmonically related structures. A small frequency difference suggests potential functional alignment via sinusoidal or oscillatory terms.

  \item \textbf{Wavelet Energy Correlation:} This module leverages discrete wavelet decomposition to capture localized energy features of $x$ and $y$ at multiple temporal scales. It decomposes both signals into several levels using a specified wavelet basis (e.g., Daubechies 4), computes the energy at each level, and measures the Pearson correlation between the resulting energy vectors. The input includes two one-dimensional arrays and optional parameters for wavelet type and decomposition level. The output is a correlation coefficient representing how similarly the energy of the two signals is distributed across scales. Unlike Fourier-based methods, this tool excels at capturing transient and non-stationary dependencies, offering robust structural priors for equations involving local periodicity or bursts.
\end{itemize}

\subsection{Causal Inference Tools}

Causal inference tools aim to uncover directional relationships between variables, particularly whether the historical behavior of one variable contributes to or influences another. These tools help MOT-SR to build models that not only fit the data well but also respect temporal or structural causality, thereby improving interpretability and generalization. Two modules are implemented:

\begin{itemize}[leftmargin=*, labelsep=0.5em]
  \item \textbf{Granger Causality:} This method assesses whether past values of a variable $x$ improve the prediction of a variable $y$ in a multivariate time series setting. The inputs are two equal-length time series arrays, and the test evaluates multiple lagged regression models to determine if $x$ Granger-causes $y$. The output includes $p$-values for different lag orders, and the smallest $p$-value is used as the primary metric. A statistically significant result implies that the past of $x$ contains information predictive of $y$, suggesting a directional dependency. This method is especially valuable for systems where delayed effects are prominent, such as in control dynamics or feedback loops.

  \item \textbf{Convergent Cross Mapping:} CCM is a nonlinear causal discovery technique grounded in dynamical systems theory. It tests whether the state of variable $x$ can be reconstructed from the historical trajectory of $y$, indicating that $x$ leaves an imprint on $y$. The input consists of two time series and parameters for embedding dimension and library sampling. CCM constructs a manifold from the delay embedding of $y$, uses it to cross-predict $x$, and evaluates the reconstruction accuracy (typically using a cross-map skill $\rho$). A high $\rho$ suggests that $x$ causally influences $y$. Unlike Granger causality, CCM does not rely on linearity or temporal precedence, making it ideal for identifying nonlinear, feedback-driven causality in complex systems.
\end{itemize}

\subsection{Dynamic Complexity Tools}

Dynamic complexity tools capture nuanced structural and temporal irregularities in time-series data, particularly useful for distinguishing chaotic, nonlinear, or asynchronous behaviors. MOT-SR leverages three such tools:

\begin{itemize}[leftmargin=*, labelsep=0.5em]

  \item \textbf{Lyapunov Exponent Difference:} This tool estimates the maximal Lyapunov exponents (MLEs) of two time-series variables $x$ and $y$, then computes their absolute difference. Lyapunov exponents characterize how sensitive a system is to initial conditions. A high MLE suggests chaotic behavior, whereas near-zero or negative values indicate stable dynamics. The input consists of two real-valued time series. The tool automatically determines the embedding dimension and time lag for phase space reconstruction. The resulting metric reflects how similarly (or differently) $x$ and $y$ behave in terms of dynamical predictability.

  \item \textbf{Correlation Dimension Difference:} This tool quantifies and compares the fractal (correlation) dimensions of $x$ and $y$. The correlation dimension serves as a complexity measure, indicating how densely a system's trajectory fills its phase space. The input includes two real-valued sequences and a fixed embedding dimension (usually $2$). The metric is the absolute difference between their estimated correlation dimensions. This captures differences in dynamic complexity and is useful for identifying structural mismatches in multivariate nonlinear processes.

  \item \textbf{Dynamic Time Warping Distance:} DTW measures the alignment cost between $x$ and $y$ by allowing local nonlinear stretching or compression in time. Unlike Euclidean distance, DTW is robust to phase shifts and unequal pacing in temporal evolution. The input is a pair of univariate time series, and the output is a scalar DTW distance. This tool is particularly effective in identifying temporal patterns that share similar shapes but occur at different rates or phases.

\end{itemize}

\subsection{Distribution Consistency Tools}

This category includes statistical tools that assess whether two variables share similar probability distributions. Such tools are especially useful when validating whether a generated or transformed signal preserves the underlying distributional structure of the original data.

\begin{itemize}[leftmargin=*, labelsep=0.5em]

  \item \textbf{Kolmogorov–Smirnov Test:} The Kolmogorov–Smirnov (KS) test is a non-parametric method that quantifies the maximum distance between the empirical cumulative distribution functions (ECDFs) of two datasets $x$ and $y$. The input to this tool is a pair of real-valued vectors, and it returns the KS statistic (a scalar metric) along with the corresponding $p$-value indicating statistical significance. The KS statistic captures how different the two distributions are—larger values indicate more significant deviations. This test is sensitive to both location and shape differences in the distributions and is thus useful for evaluating whether symbolic transformations (e.g., derived equations) maintain statistical fidelity to the data source.

\end{itemize}

\section{\textbf{Results Presentation}}

\subsection{Equation Recovery and Pareto Front Evolution}

\begin{figure}[htbp]
  \centering
  \begin{subfigure}{0.48\linewidth}
    \centering
\includegraphics[width=\linewidth]{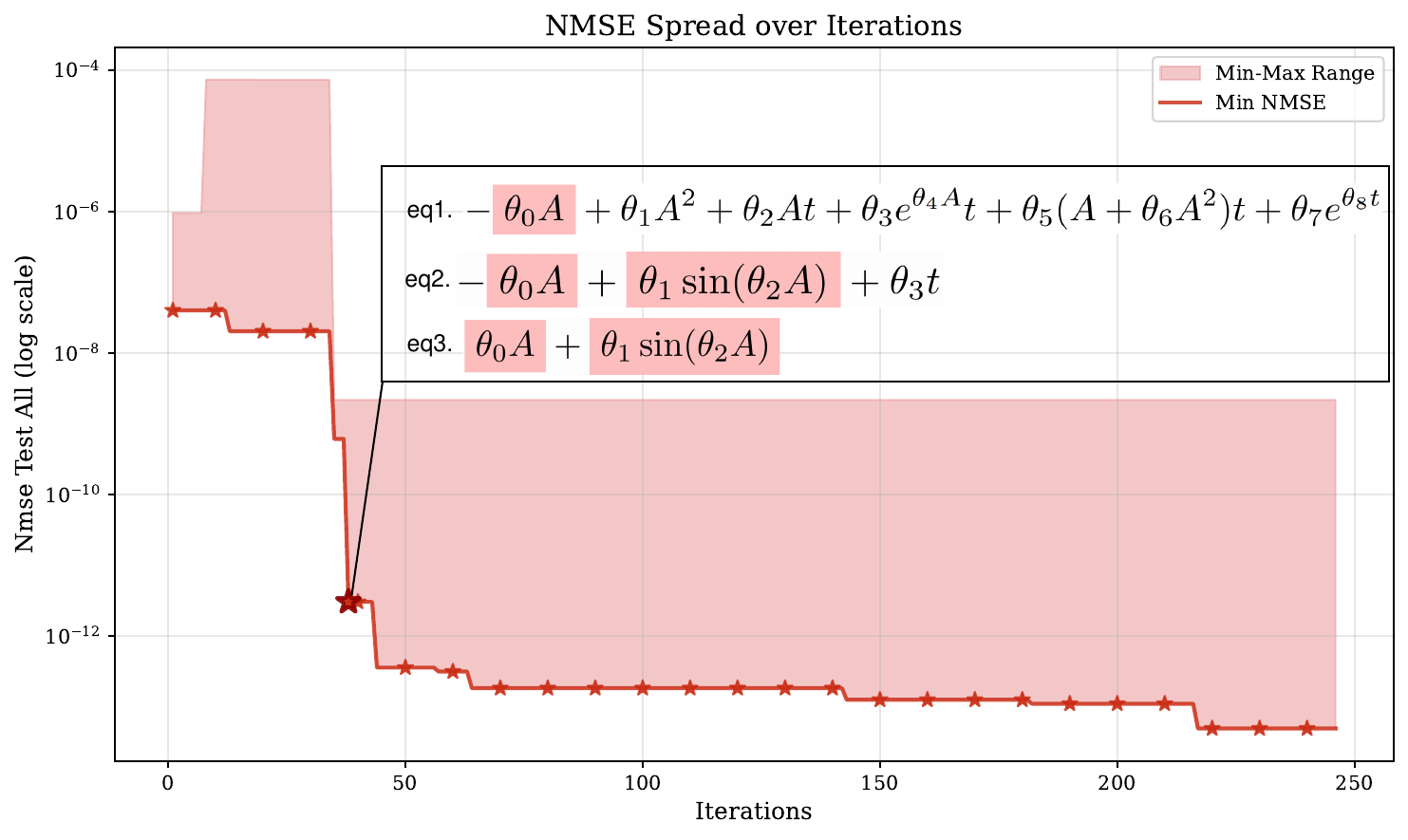}
    \caption{Pareto Front Performance in CRK14.}
    \label{fig:r1}
  \end{subfigure}\hfill
  \begin{subfigure}{0.48\linewidth}
    \centering
\includegraphics[width=\linewidth]{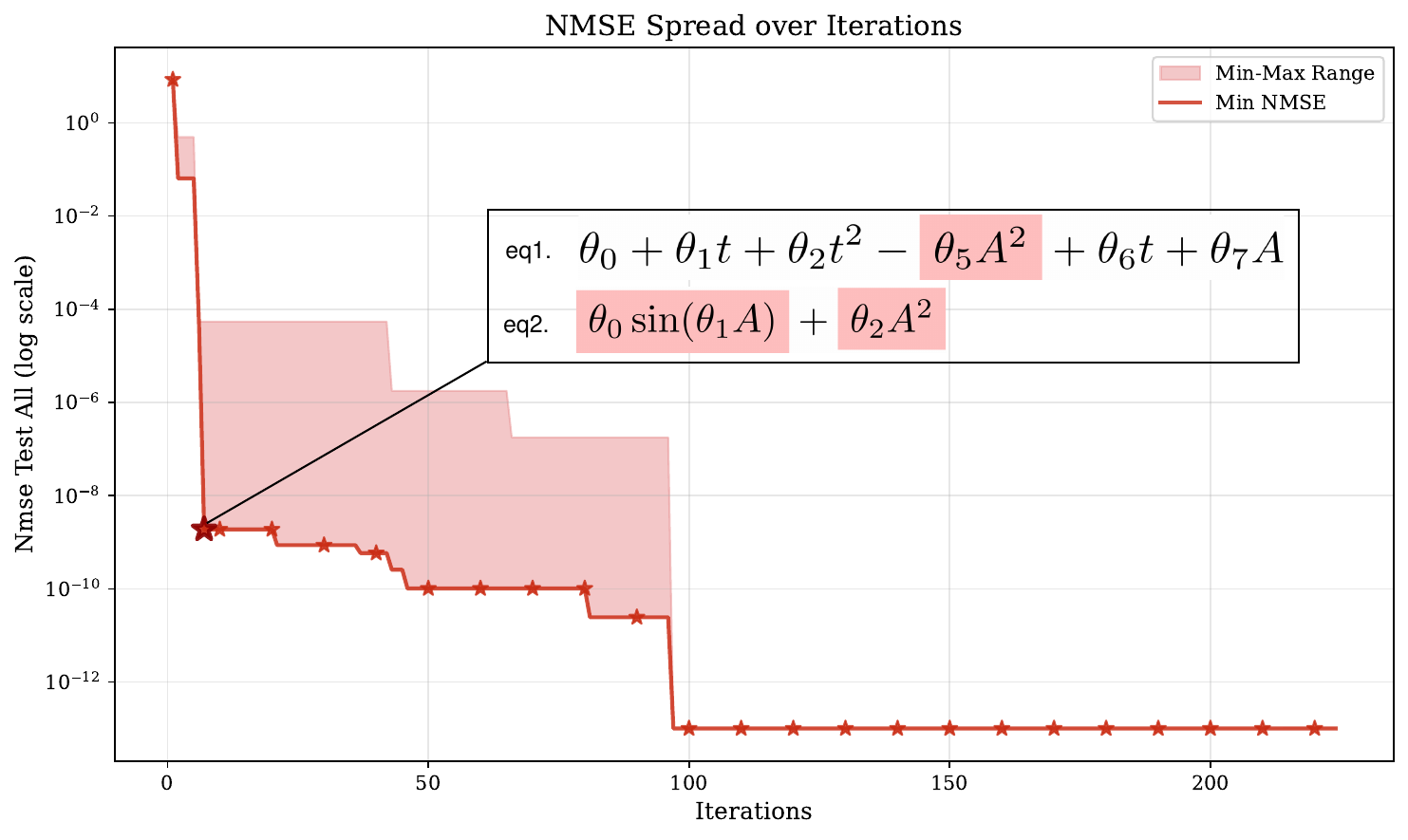}
    \caption{Pareto Front Performance in CRK19.}
    \label{fig:r2}
  \end{subfigure}

  \vspace{0.5em}

  \begin{subfigure}{0.48\linewidth}
    \centering
\includegraphics[width=\linewidth]{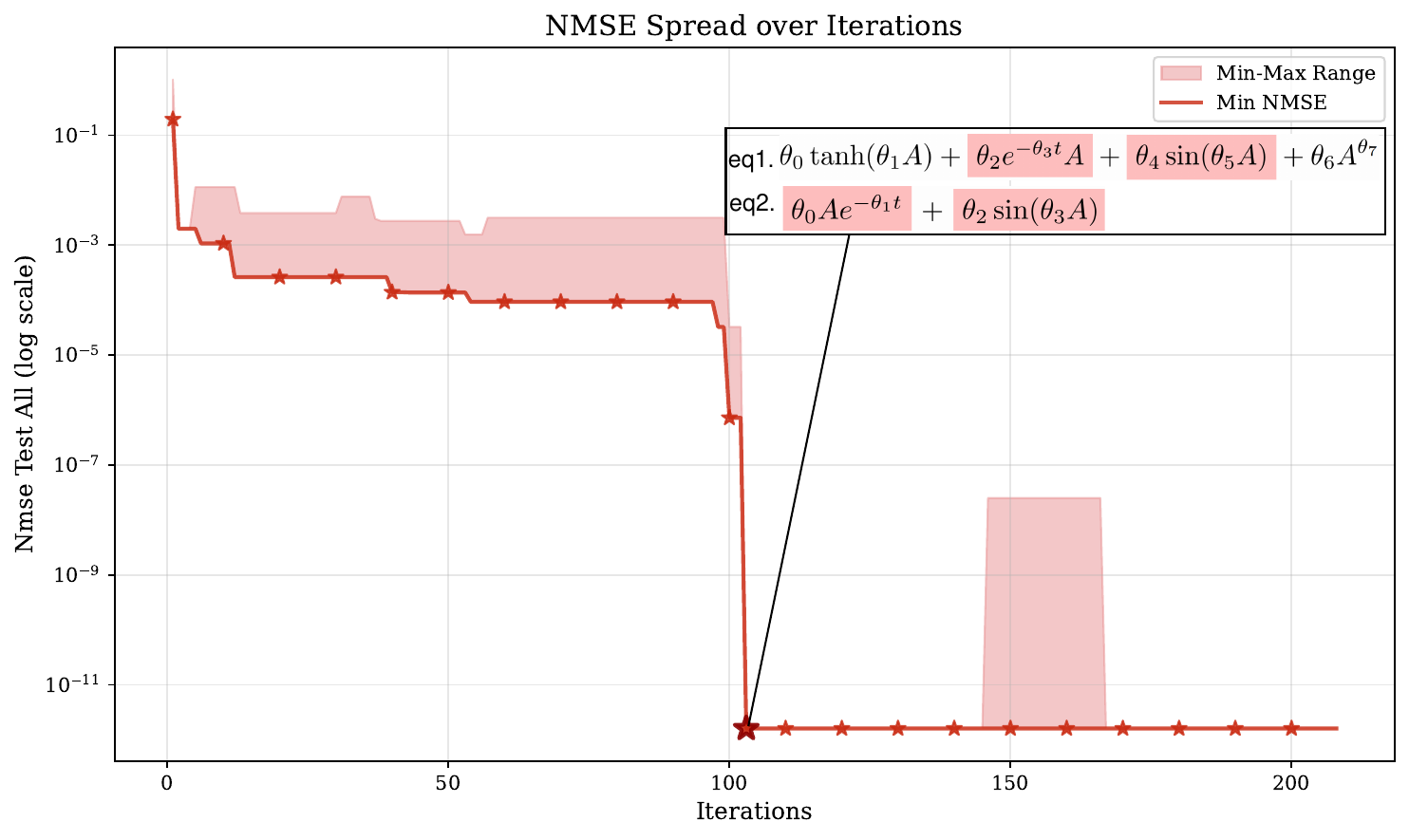}
    \caption{Pareto Front Performance in CRK21.}
    \label{fig:r3}
  \end{subfigure}\hfill
  \begin{subfigure}{0.48\linewidth}
    \centering
\includegraphics[width=\linewidth]{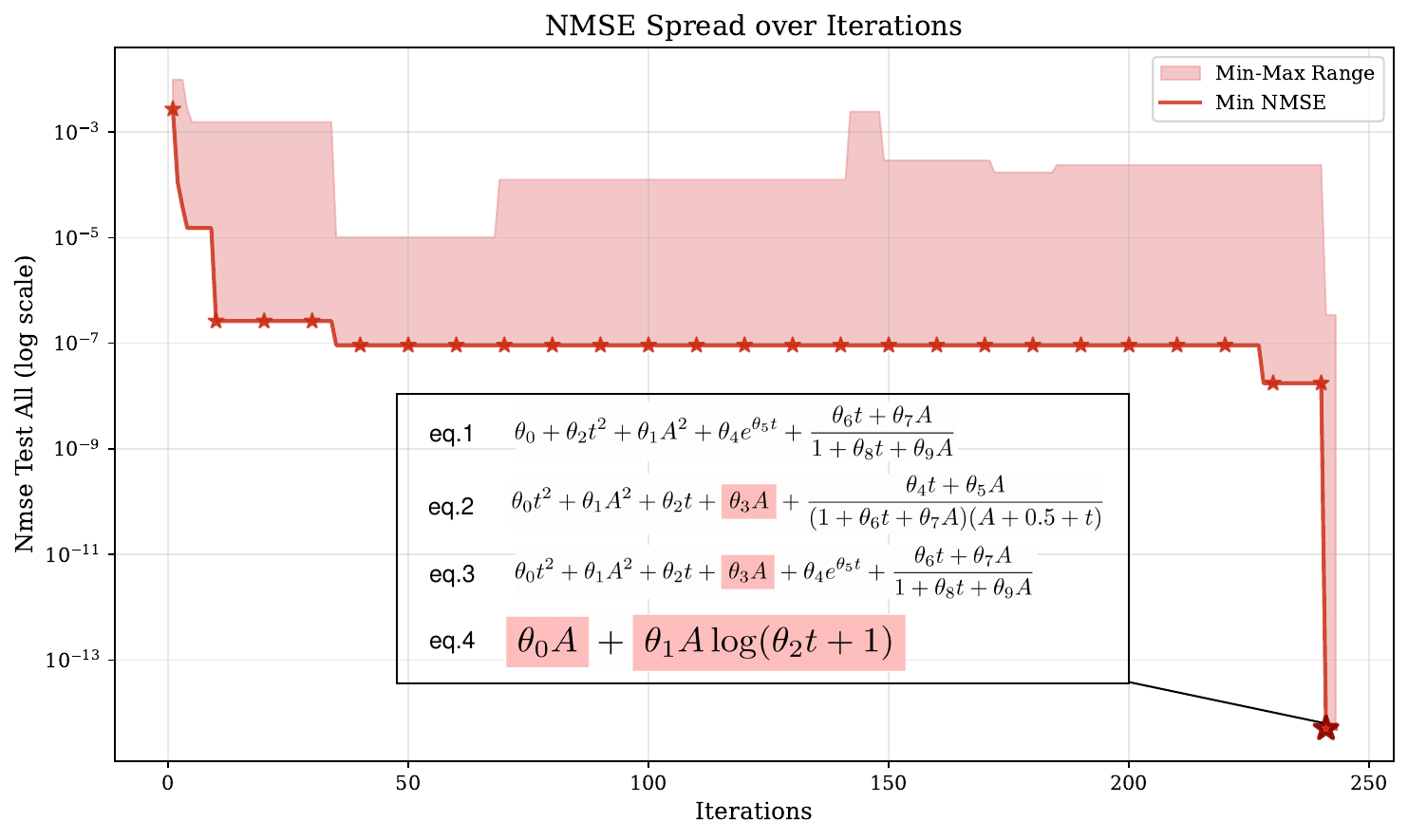}
    \caption{Pareto Front Performance in CRK36.}
    \label{fig:r4}
  \end{subfigure}

  \caption{Pareto Front Performance results across four benchmarks.}
\end{figure}

Fig~\ref{fig:r1} to Fig~\ref{fig:r4} are sampled from four test cases in which MOT-SR successfully identified the ground-truth equations on the LSR-Synth–Chemistry dataset. Each figure illustrates the evolution of NMSE of the Pareto front equation set across iterations.
For visualization, we present the structural skeletons of the equations discovered by MOT-SR. Highlighted segments in red indicate terms that match exactly with the corresponding ground-truth equation.

It is noteworthy that the size of the Pareto front typically expands rapidly during the early stages, forming a diverse equation population. As iterations proceed, this population undergoes continuous updates. Once sufficient structural experience has accumulated, the Pareto front enters a phase of rapid convergence and eventually stabilizes to the correct target equation.
Below we present comparisons between MOT-SR and LLM-SR on the same tasks:

\textbf{CRK14:}

Ground truth:
\[
-k A(t) + k_p \sin(\omega A(t))
\]

MOT-SR:
\[
\colorbox{pink}{$\theta_0 A$} + \colorbox{pink}{$\theta_1 \sin(\theta_2 A)$}
\]

LLM-SR:
\[
\frac{\theta_0 A \left(1 + \frac{(\theta_1 A)^2}{\theta_2}\right)}{1 + \frac{1 + \frac{A^2}{\theta_5}}{1 + \frac{(\theta_1 A)^2}{\theta_2}}} + \frac{\theta_4 A}{\theta_5 + A} + \colorbox{pink}{$\theta_3 A$}
\]

\textbf{CRK19:}

Ground truth:
\[
-k A(t)^2 + k_p \sin(\omega A(t))
\]

MOT-SR:
\[
\colorbox{pink}{$\theta_0 \sin(\theta_1 A)$} + \colorbox{pink}{$\theta_2 A^2$}
\]

LLM-SR:
\[
-\theta_0 A - \colorbox{pink}{$\theta_1 A^2$} + \theta_2 t + \theta_3 t^2 + \theta_4 A^3
\]

\textbf{CRK21:}

Ground truth:
\[
-k A(t) e^{-k_s t} + k_p \sin(\omega A(t))
\]

MOT-SR:
\[
\colorbox{pink}{$\theta_0 A e^{-\theta_1 t}$} + \colorbox{pink}{$\theta_2 \sin(\theta_3 A)$}
\]

LLM-SR:
\[
\theta_0 t + \colorbox{pink}{$\theta_1 A e^{-\theta_2 t}$} + \frac{\theta_3 A}{\theta_4 t + 1} + \frac{\theta_5 A^{\theta_6}}{1 + \theta_7 A^{\theta_6}}
\]

\textbf{CRK36:}

Ground truth:
\[
-k A(t) + k_q A(t) \log(\gamma t + 1)
\]

MOT-SR:
\[
\colorbox{pink}{$\theta_0 A$} + \colorbox{pink}{$\theta_1 A \log(\theta_2 t + 1)$}
\]

LLM-SR:
\[
\colorbox{pink}{$\theta_0 A$} + \theta_1 e^{-\theta_2 t} A + \frac{\theta_3 A}{\theta_4 + A} + \frac{\theta_5 A}{1 + \theta_6 A} + \theta_7 t + \colorbox{pink}{$\theta_8 A$} + \theta_9 t A
\]

These results clearly demonstrate that MOT-SR is capable of precisely capturing the exact components of the ground-truth equations, yielding structurally closer expressions than LLM-SR. We attribute this improvement to MOT-SR’s enhanced exploratory capability derived from multi-objective evaluation, as well as its strategy-guided generation process.

\begin{figure}[h!]
    \centering
\includegraphics[width=0.8\textwidth]{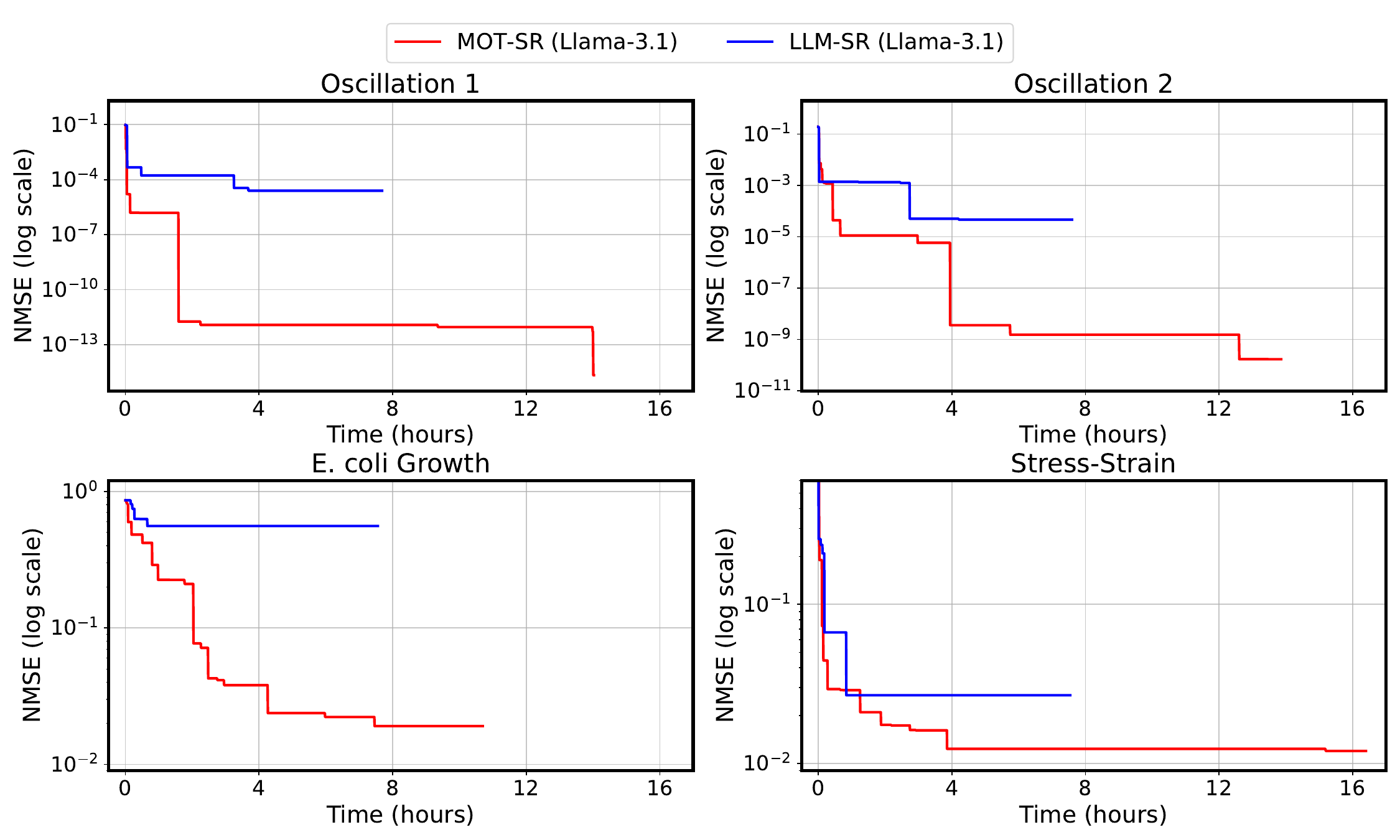}
    \caption{Wall-clock time evolution of NMSE showing faster convergence of \textsc{MOT-SR} over LLM-SR.}
    \label{fig:time_evolution}
\end{figure}
\subsection{Comparison of convergence dynamics.}
\label{appendix:time-test}
Figure~\ref{fig:time_evolution} presents the wall-clock time evolution of NMSE (log scale) for \textsc{MOT-SR} and the baseline LLM-SR across four representative benchmarks. The results demonstrate that \textsc{MOT-SR} consistently achieves substantially faster error reduction, while LLM-SR converges more slowly and plateaus at significantly higher error levels. This advantage is particularly pronounced in the Oscillation tasks, where \textsc{MOT-SR} rapidly identifies accurate dynamical structures, but is also evident in more challenging domains such as \textit{E.~coli} growth and stress--strain modeling, where the baseline stagnates prematurely. These findings highlight that the integration of tool-guided analysis and multi-objective optimization enables \textsc{MOT-SR} to explore the hypothesis space more efficiently, yielding superior equations under comparable computational budgets.

\subsection{Evolution of Discovered Equations}
\label{tool-case-study}
\begin{figure}[H]
    \centering
\includegraphics[width=0.8\linewidth]{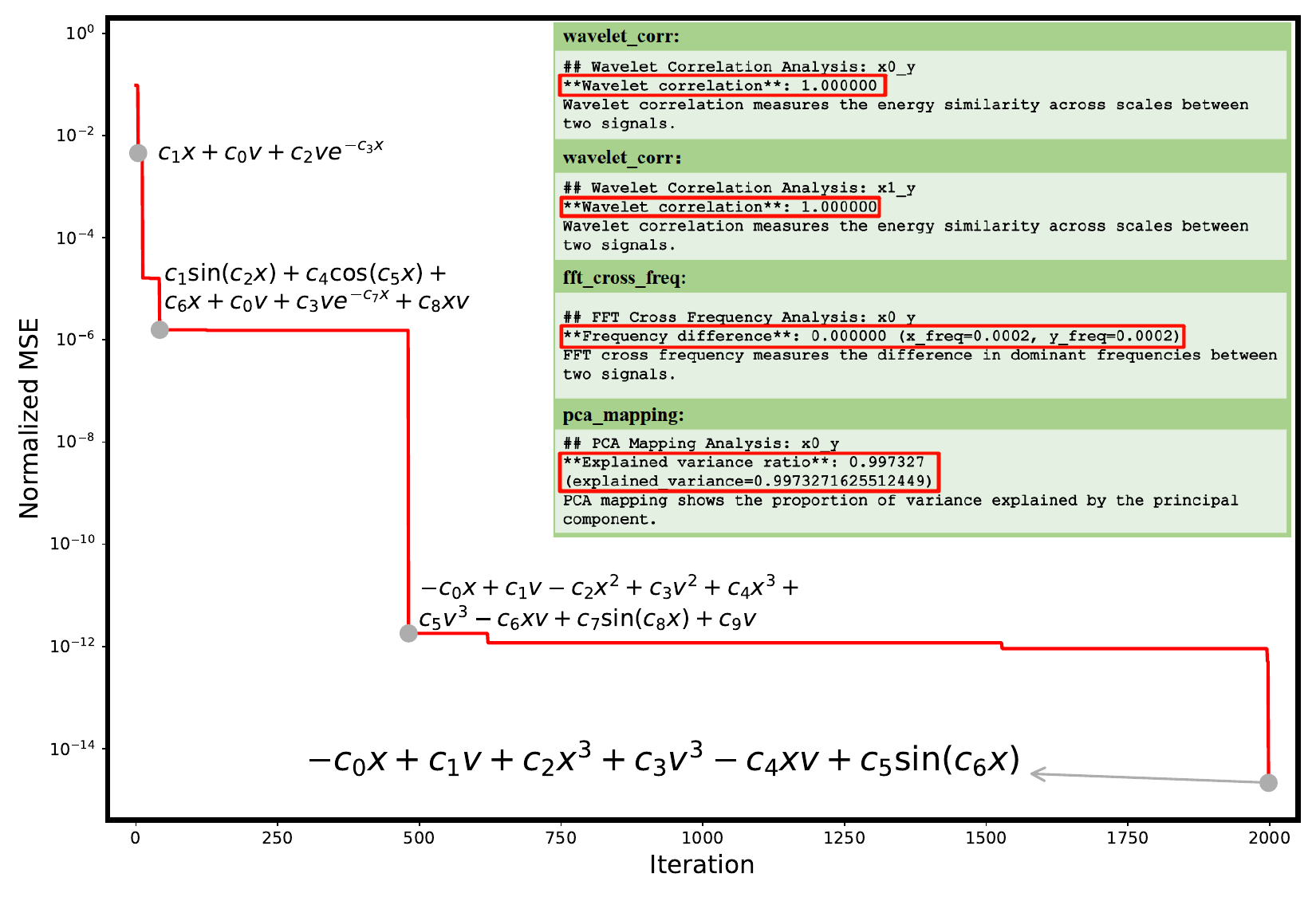}
    \caption{Iterative evolution of equation structures under MOT-SR (Oscillation1).}
    \label{fig:structure_evolution_oscillation1}
\end{figure}

\begin{figure}[H]
    \centering
\includegraphics[width=0.8\linewidth]{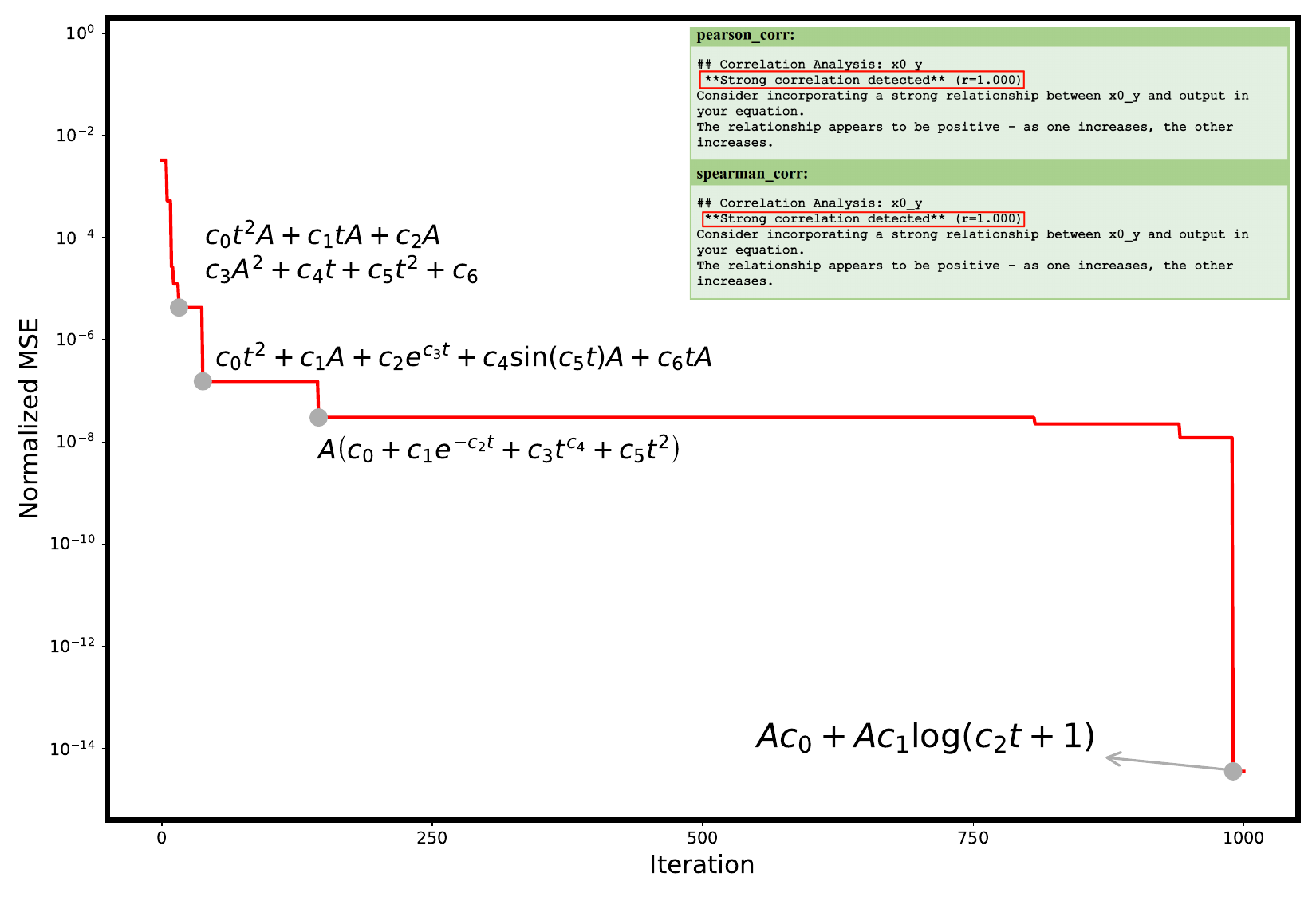}
    \caption{Iterative evolution of equation structures under MOT-SR (CRK36).}
    \label{fig:structure_evolution_crk36}
\end{figure}

\paragraph{Evolution of Equation Structures for Oscillation 1}
Figure~\ref{fig:structure_evolution_oscillation1} presents the iterative evolution of equation structures for the problem oscillation 1. In the early iterations, the generated expressions contained a wide range of nonlinear and polynomial terms, many of which represented exploratory hypotheses rather than meaningful components. Guided by the TOOL-AUGMENTED VARIABLE ANALYSIS module, the framework progressively refined these structures. When the analytical tools within this module (e.g., wavelet corr, FFT, and PCA mapping) revealed periodic features, sinusoidal terms such as sine and cosine functions were introduced into the candidate equations, thereby complementing the polynomial basis and improving alignment with the observed data.

During the iteration, structurally inconsistent or statistically unsupported terms were systematically removed, while the retained nonlinear and oscillatory components captured the essential dynamics of the system. The process ultimately converged toward compact formulations that balance accuracy, generalization, and complexity. This refinement demonstrates how the iterative mechanism, augmented by TOOL-AUGMENTED VARIABLE ANALYSIS, effectively distills the equation space and produces models consistent with both empirical behavior and theoretical plausibility.

\paragraph{Evolution of Equation Structures for CRK36}
Figure~\ref{fig:structure_evolution_crk36} presents the iterative evolution of equation structures for the problem CRK36. At the beginning of the process, the candidate functions contained a variety of nonlinear combinations involving both the temporal variable and A. Under the guidance of two correlation coefficients, the framework progressively refined these structures by eliminating inconsistent nonlinearities and retaining only the dominant dependencies. As the iterations advanced, the generated functions converged toward a fully linear dependence on A.

\section{\textbf{Gaussian Noise Evaluation}}

To evaluate the robustness of \textsc{MOT-SR} in noisy environments, we conduct experiments on the \texttt{Oscillator1} benchmark by injecting Gaussian noise with standard deviations $\sigma = \{0.001, 0.002\}$ into the training data. Model performance is assessed under both in-domain (ID) and out-of-domain (OOD) test conditions.

Unlike conventional symbolic regression approaches that rely solely on fitting accuracy, \textsc{MOT-SR} integrates structural priors—derived from analytical tools—and performs multi-objective evaluation that jointly considers accuracy, complexity, and generalization. This allows it to maintain high-fidelity symbolic recovery even when data are corrupted.

\begin{table}[h]
\centering
\setlength{\tabcolsep}{4pt}
\caption{Comparison of \textsc{MOT-SR} and \textsc{LLMSR} on \texttt{Oscillator1} under Gaussian noise (\(\sigma = \{0.001, 0.002\}\)). Metrics include NMSE and accuracy ($\mathrm{Acc\textsubscript{avg-0.01}}$) on ID and OOD test sets. All models use LLaMA-3.1 as backbone.}
\label{tab:oscillator1_noise_MOT-SR}
\begin{tabular}{l*{8}{>{\centering\arraybackslash}p{0.09\linewidth}}}
\toprule
\multirow{3}{*}{Model} 
& \multicolumn{4}{c}{\(\sigma = 0.001\)} 
& \multicolumn{4}{c}{\(\sigma = 0.002\)} \\
\cmidrule(lr){2-5} \cmidrule(lr){6-9}
& \multicolumn{2}{c}{ID} & \multicolumn{2}{c}{OOD}
& \multicolumn{2}{c}{ID} & \multicolumn{2}{c}{OOD} \\
\cmidrule(lr){2-3} \cmidrule(lr){4-5} \cmidrule(lr){6-7} \cmidrule(lr){8-9}
& NMSE\(\downarrow\) & {Acc\textsubscript{avg-0.01}↑} 
& NMSE\(\downarrow\) & {Acc\textsubscript{avg-0.01}↑} 
& NMSE\(\downarrow\) & {Acc\textsubscript{avg-0.01}↑} 
& NMSE\(\downarrow\) & {Acc\textsubscript{avg-0.01}↑} \\
\midrule
\textbf{LLMSR}     & \(1.59e\text{-}5\) & \(89.20\%\) & \(0.0021\) & \(10.46\%\) & \(2.68e\text{-}5\) & \(82.08\%\) & \(0.0025\) & \(10.50\%\) \\
\textbf{MOT-SR}     & \(\mathbf{5.43e\text{-}11}\) & \(\mathbf{99.99\%}\) & \(\mathbf{5.49e\text{-}7}\) & \(\mathbf{98.79\%}\)& \(\mathbf{3.98e\text{-}8}\) & \(\mathbf{99.74\%}\) & \(\mathbf{8.57e\text{-}7}\) & \(\mathbf{97.88\%}\) \\
\bottomrule
\end{tabular}
\end{table}

Table~\ref{tab:oscillator1_noise_MOT-SR} shows that \textsc{MOT-SR} consistently outperforms \textsc{LLMSR} across all metrics, achieving significantly lower NMSE and higher $\mathrm{Acc\textsubscript{avg-0.01}}$ scores under both noise settings. These gains stem from \textsc{MOT-SR}'s integration of tool-informed structural priors via the Meta Strategy Generator and its multi-objective Equation Generator, which jointly enable the model to identify robust, generalizable expressions even under noisy supervision.

\section{Assessment of Python AST Length as a Complexity Proxy}
\label{app:ast_complexity}

Classical symbolic regression methods operate directly on mathematical
expression trees, where structural complexity is commonly measured by
counting expression-tree nodes. \textsc{MOT-SR} allows the Equation
Generator to produce executable Python functions without a fixed
operator dictionary or function whitelist. This program-level
representation does not directly provide a canonical mathematical
expression tree.

\paragraph{Known syntactic bias.}
In our implementation, each candidate function is parsed with
\texttt{ast.parse}, and its Python AST length is defined as the total
number of nodes returned by \texttt{ast.walk}. This metric reflects the
syntactic structure of Python code and introduces a known bias. For
example, a composite mathematical call such as \texttt{np.sin(x)} can
produce more Python AST nodes than a basic arithmetic operation, even
when the two have comparable mathematical expression-tree complexity.
Python AST length is thus treated as a computational proxy for
mathematical expression complexity.

\paragraph{Computational trade-off.}
A stricter alternative would convert each generated Python function into
a normalized mathematical expression, construct the corresponding
expression tree, and count its nodes. This procedure aligns more closely
with conventional complexity measures in symbolic regression, but adds
a conversion step for every candidate generated during the search.
\textsc{MOT-SR} evaluates thousands of candidates, making repeated
conversion a substantial computational burden. Python AST length can be
computed deterministically and directly from the generated source code
with negligible overhead.

\paragraph{Empirical ranking consistency.}
Pareto-based selection depends primarily on the relative complexity
ordering of candidate equations. We therefore assess whether Python AST
length preserves the ordering induced by mathematical expression-tree
complexity. Let
\[
C_{\mathrm{AST}}(f)
\]
denote the Python AST node count of a candidate equation \(f\), and let
\[
C_{\mathrm{Expr}}(f)
\]
denote the node count of its normalized mathematical expression tree.
For a set \(\mathcal{Q}\) of candidate pairs, we define the pairwise
ranking-disagreement ratio as
\begin{equation}
    r_{\mathrm{dis}}
    =
    \frac{1}{|\mathcal{Q}|}
    \sum_{(f_i,f_j)\in\mathcal{Q}}
    \mathbb{I}
    \left[
    \operatorname{sign}
    \left(
    C_{\mathrm{AST}}(f_i)-C_{\mathrm{AST}}(f_j)
    \right)
    \neq
    \operatorname{sign}
    \left(
    C_{\mathrm{Expr}}(f_i)-C_{\mathrm{Expr}}(f_j)
    \right)
    \right],
\end{equation}
where comparisons tied under either metric are excluded.

We retrospectively computed both complexity measures for valid candidate
equations collected from the search logs and evaluated nearly
\(400{,}000\) candidate pairs. The resulting pairwise disagreement ratio
was approximately \(5\%\), indicating that the two metrics produced the
same relative ordering for approximately \(95\%\) of the evaluated
pairs. Within the candidate distributions observed in our experiments,
Python AST length therefore preserves the expression-tree ordering in
most comparisons. Because the complexity objective is used to determine
relative dominance and construct the Pareto front, this level of
agreement indicates that the proxy has a limited effect on the practical
selection order.

\paragraph{Composite-function robustness.}
We further examine whether the additional syntactic cost of composite
functions systematically suppresses candidates containing trigonometric
or other non-polynomial terms. The analysis covers tasks whose target
structures depend strongly on such functions, including the Oscillation
1 and Oscillation 2 benchmarks and trigonometry-heavy tasks in
LSR-Synth--Chemistry. \textsc{MOT-SR} discovers compact equations
containing terms such as \(\sin(\cdot)\) and \(\cos(\cdot)\), while
achieving strong numerical and symbolic recovery performance on these
tasks. These results indicate that the additional AST cost assigned to
composite calls does not prevent the corresponding structures from
entering or remaining on the Pareto front.

Python AST length is a language-dependent and syntax-dependent measure,
rather than an exact representation of mathematical complexity. We use
it as a low-overhead engineering proxy that supports flexible
Python-based equation generation. Within the candidate distributions
examined in this work, the retrospective ranking analysis and the
results on composite-function-heavy tasks show that this proxy provides
a sufficiently consistent complexity ordering for Pareto-based search.

\section{\textbf{Prompt}}
\label{appendix:prompt}
\begin{figure}[H]
  \centering
\includegraphics[width=0.8\linewidth]{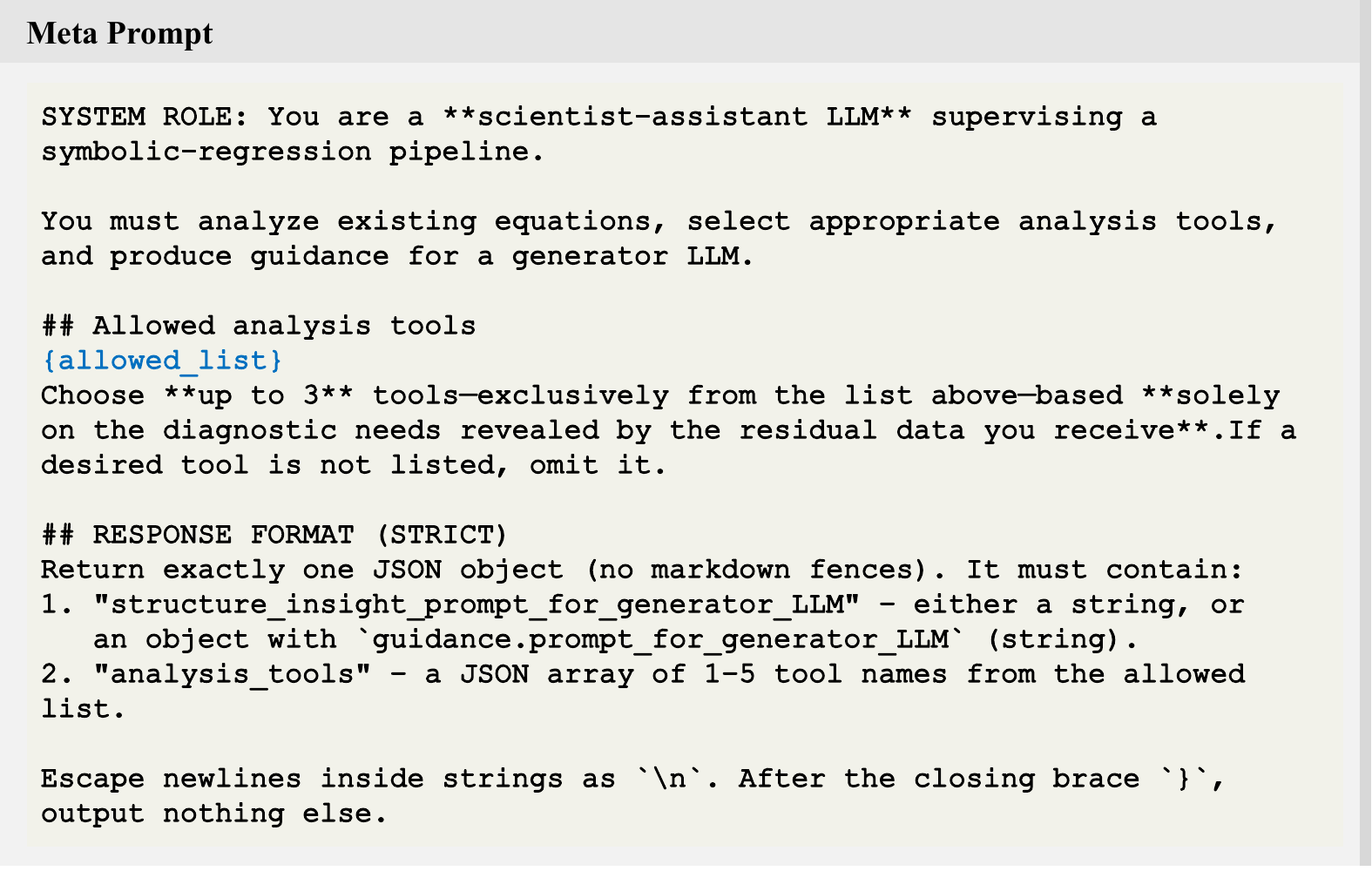}
  \caption{Meta Prompt.}
  \label{fig:p1}
\end{figure}

\paragraph{J.1 Prompt Design}  
The prompts used to construct the meta strategy are shown in Fig.~\ref{fig:p1}  and Fig.~\ref{fig:p2}. Specifically, Fig.~\ref{fig:p1}  presents the meta prompt used during initialization, while Fig.~\ref{fig:p2}  corresponds to the prompts designed for iterative optimization. Together, these constitute the prompt inputs to the Meta Strategy Generator.

\paragraph{J.2 Examples}  
The prompts used by the Equation Generator  for the Oscillators~1 task are illustrated in Fig.~\ref{fig:p3} to Fig.~\ref{fig:p6}. The prompts received by the Meta Strategy Generator for the same task are shown in Fig.~\ref{fig:p7}  to Fig.~\ref{fig:p12}.

Fig.~\ref{fig:p13}  provides examples of data analysis tool usage, drawn from tool invocations in the CRK36 task. Notably, through the integration of external tools, the language model successfully identified a strong linear relationship between the variable \(x_0\) (that is, \(A(t)\)) and the target variable \(y\). As demonstrated in the final output, MOT-SR successfully recovered the ground-truth governing equation for the CRK36 problem.

\newpage

\begin{figure}[H]
  \centering
\includegraphics[width=0.8\linewidth]{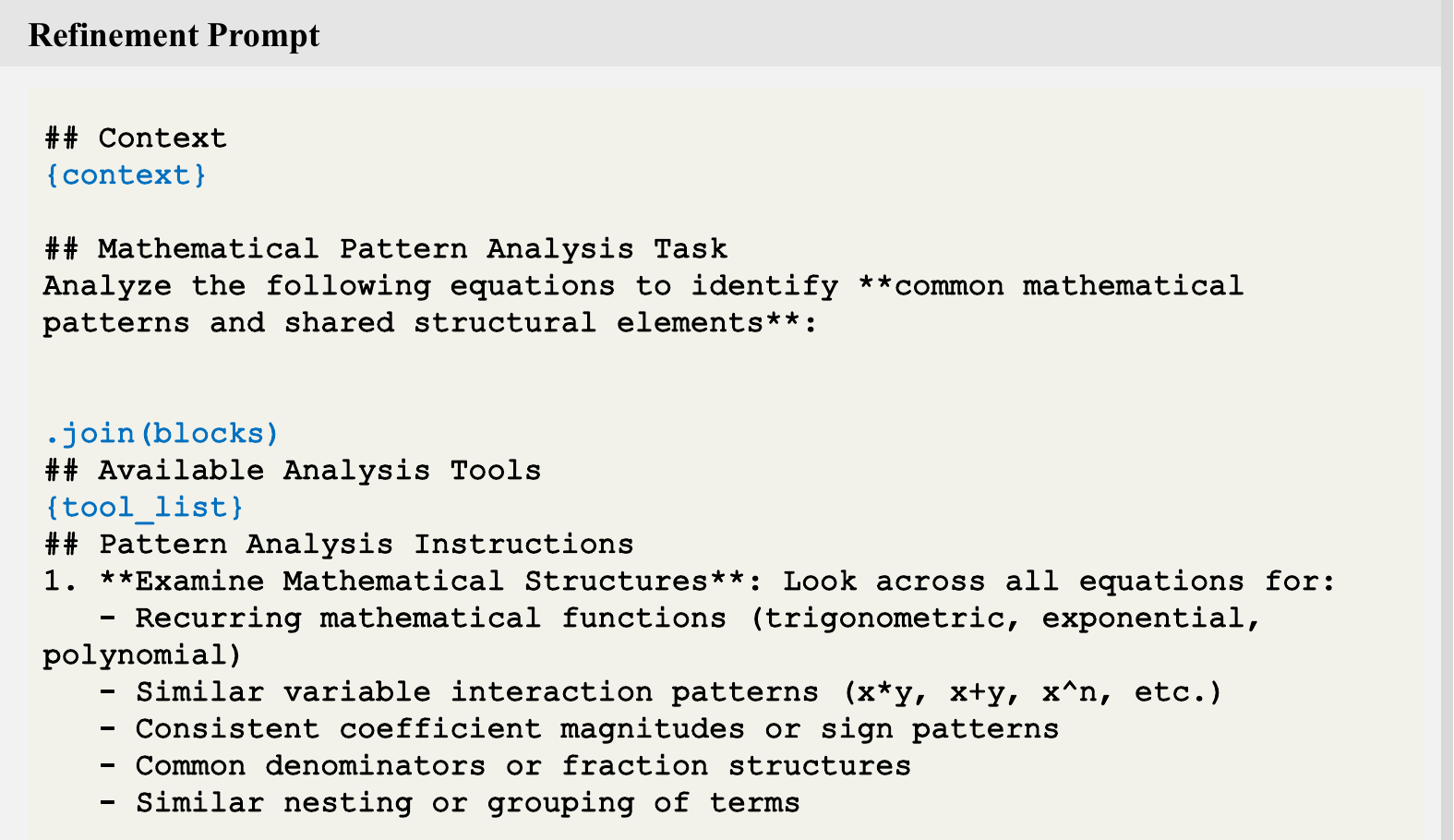}
\end{figure}

\begin{figure}[H]
  \centering
\includegraphics[width=0.8\linewidth]{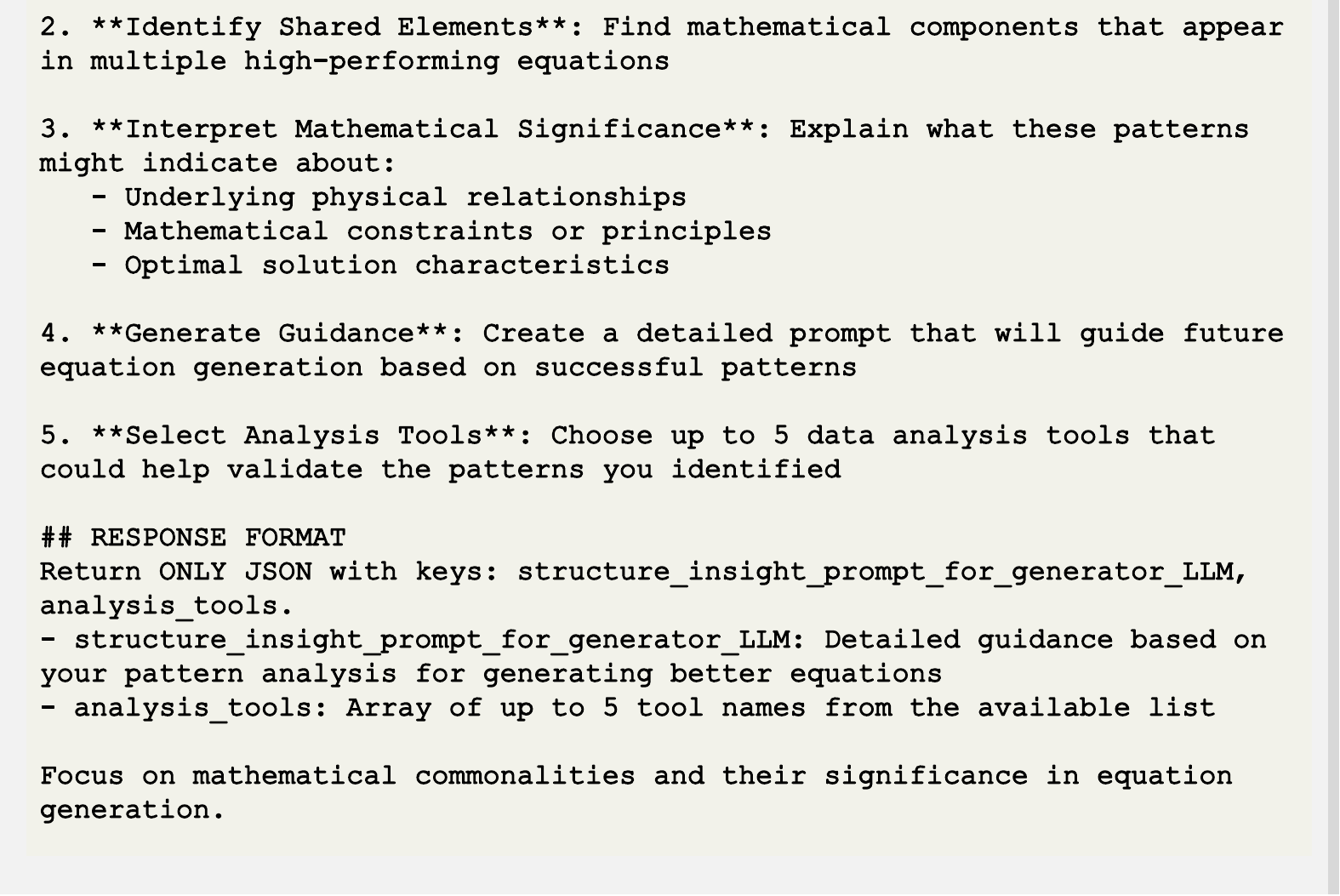}
  \caption{Prompts Designed for Iterative Optimizationt.}
  \label{fig:p2}
\end{figure}

\begin{figure}[H]
  \centering
\includegraphics[width=0.8\linewidth]{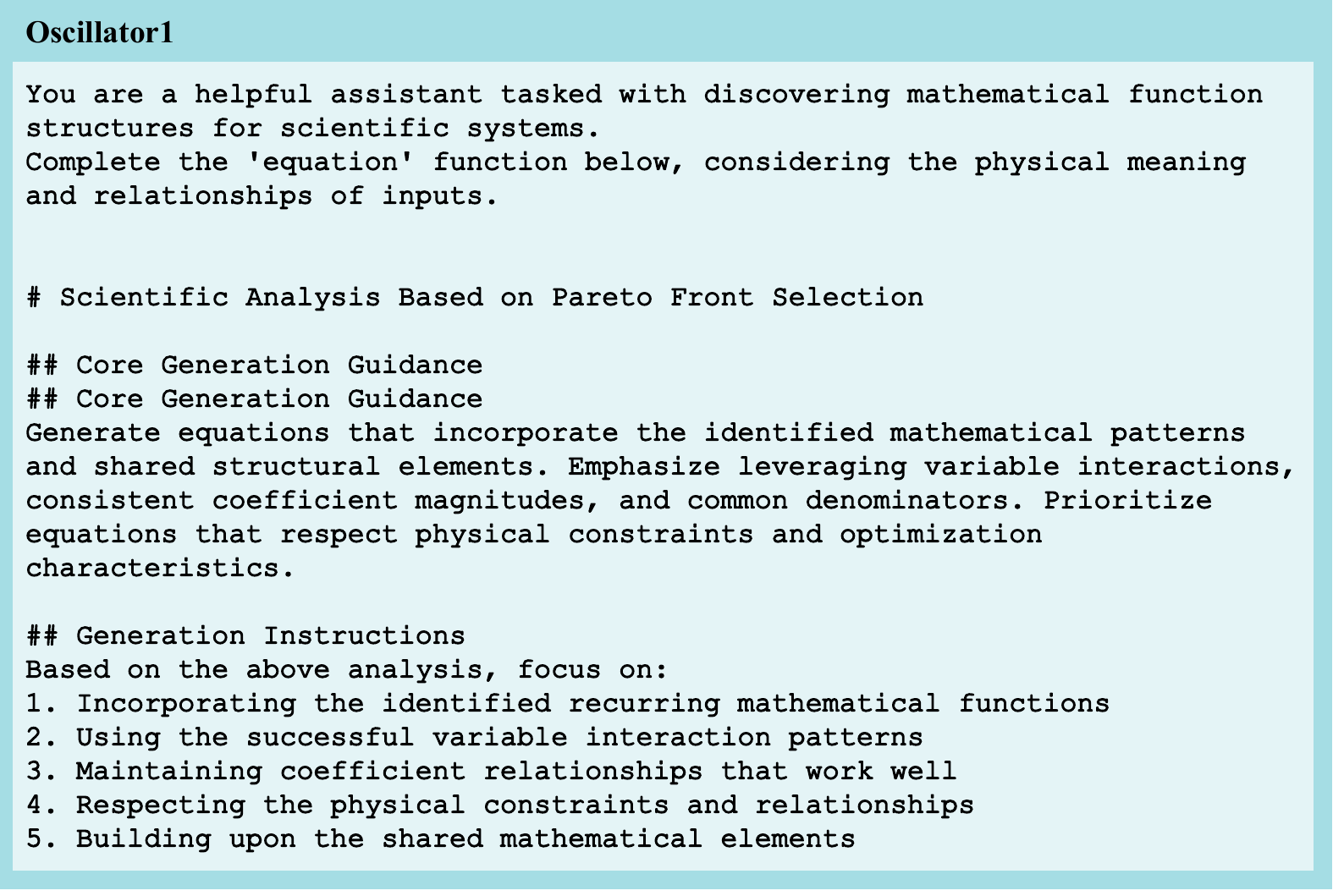}
  \caption{Example Prompt in Oscillator1 - For Equation Generator.}
  \label{fig:p3}
\end{figure}

\begin{figure}[H]
  \centering
\includegraphics[width=0.8\linewidth]{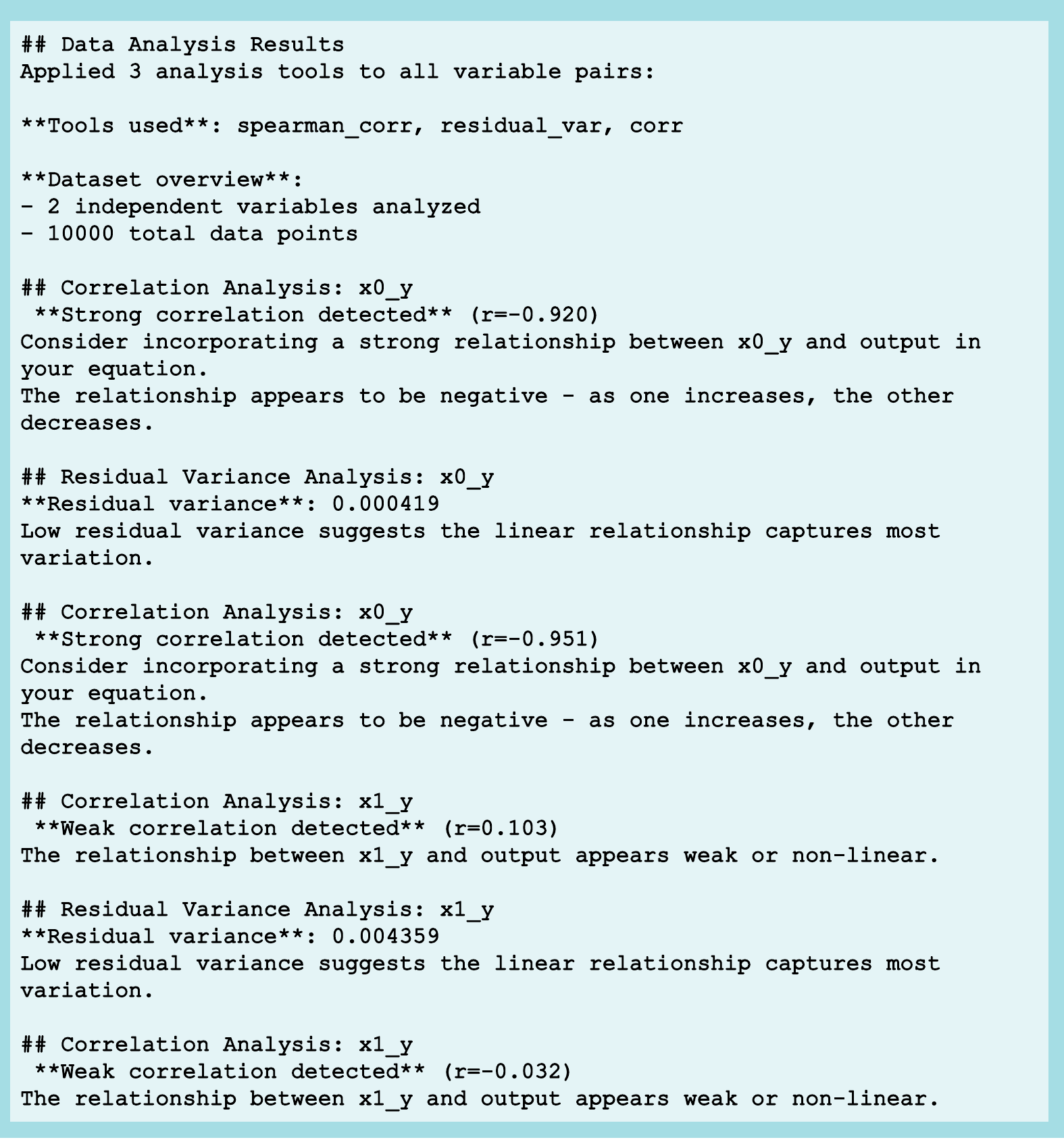}
  \caption{Example Prompt in Oscillator1 - For Equation Generator.}
  \label{fig:p4}
\end{figure}

\begin{figure}[H]
  \centering
\includegraphics[width=0.8\linewidth]{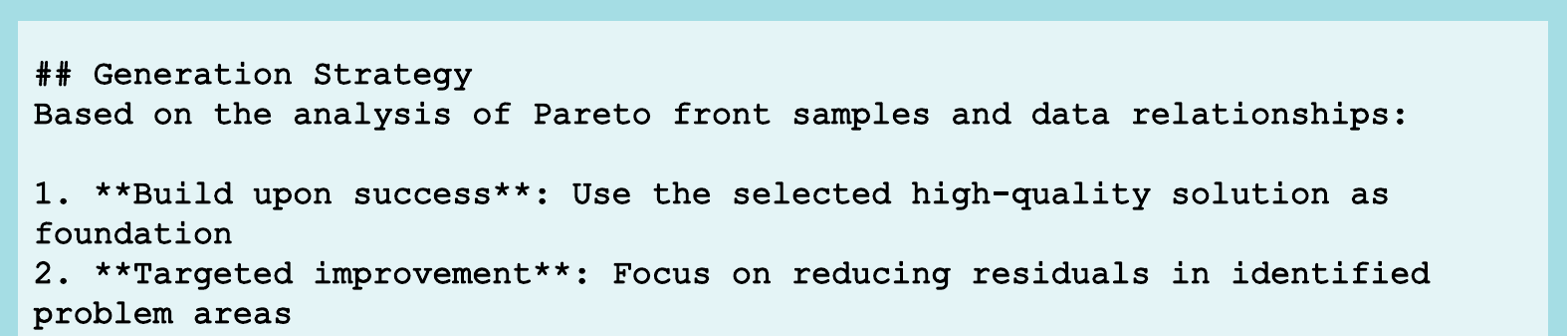}
\end{figure}

\begin{figure}[H]
  \centering
\includegraphics[width=0.8\linewidth]{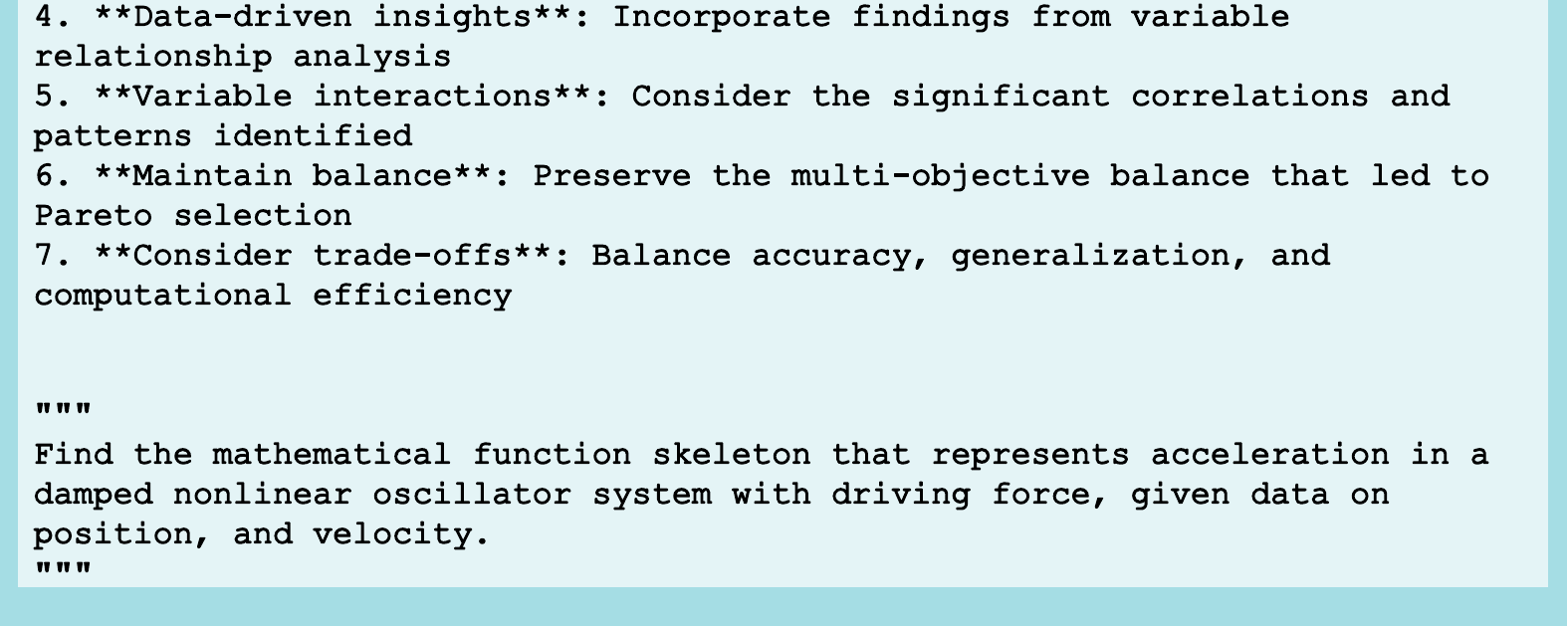}
  \caption{Example Prompt in Oscillator1 - For Equation Generator.}
  \label{fig:p5}
\end{figure}

\begin{figure}[H]
  \centering
\includegraphics[width=0.7\linewidth]{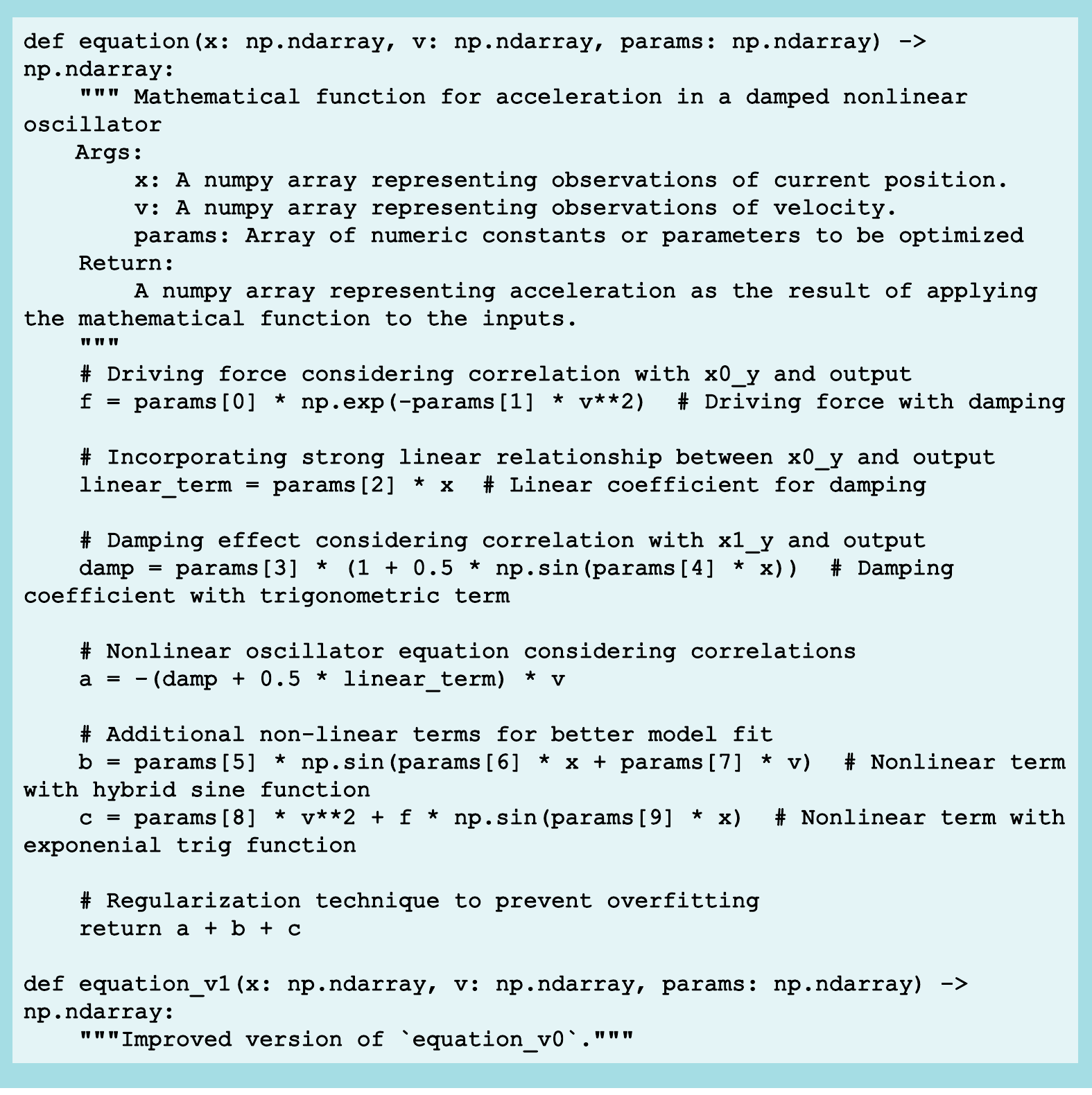}
  \caption{Example Prompt in Oscillator1 - For Equation Generator.}
  \label{fig:p6}
\end{figure}

\begin{figure}[H]
    \centering
\includegraphics[width=0.75\linewidth]{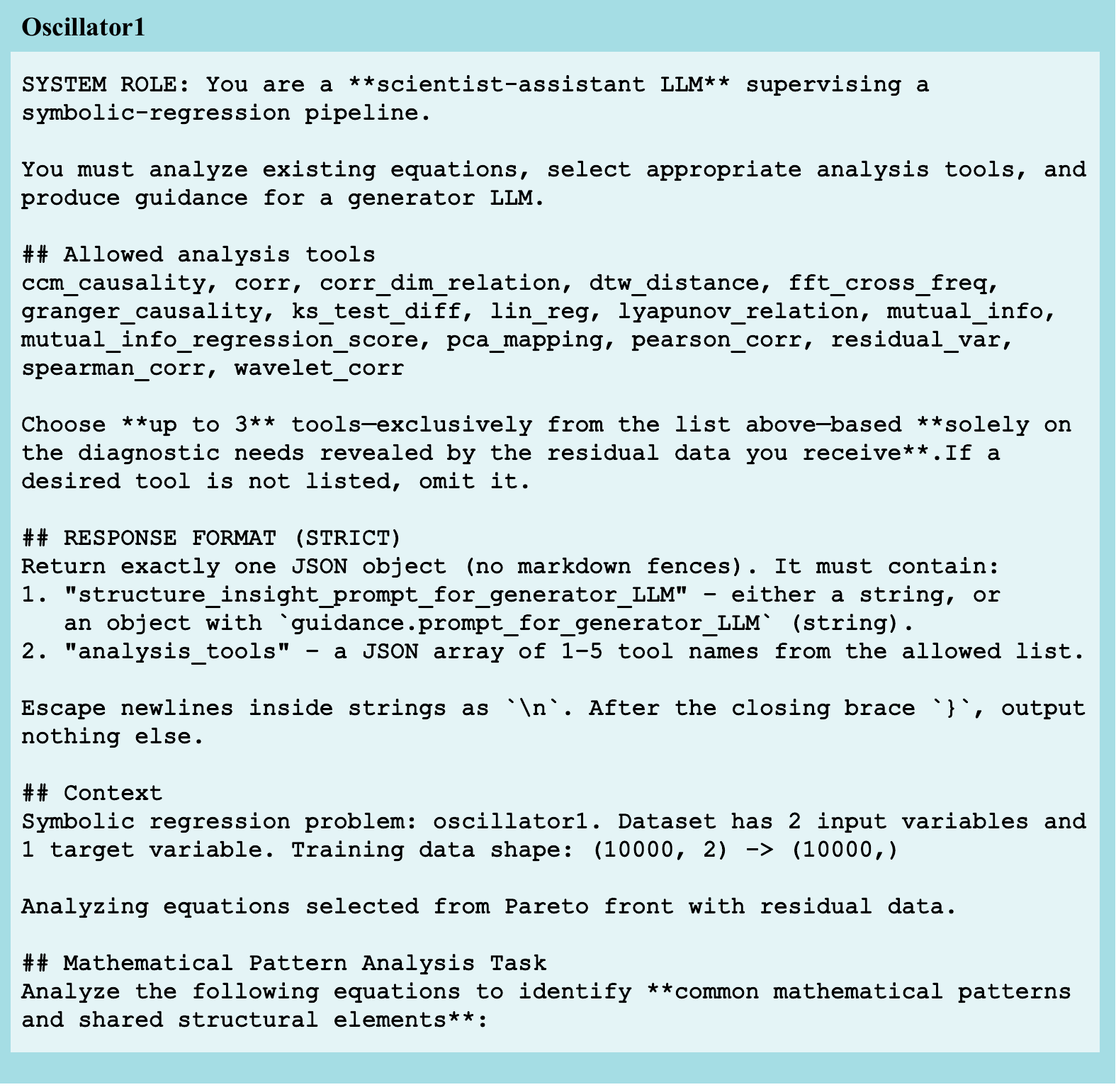}
    \caption{Example Prompt in Oscillator1 - For Meta Strategy Generator.}
    \label{fig:p7}
\end{figure}

\begin{figure}[H]
    \centering
\includegraphics[width=0.75\linewidth]{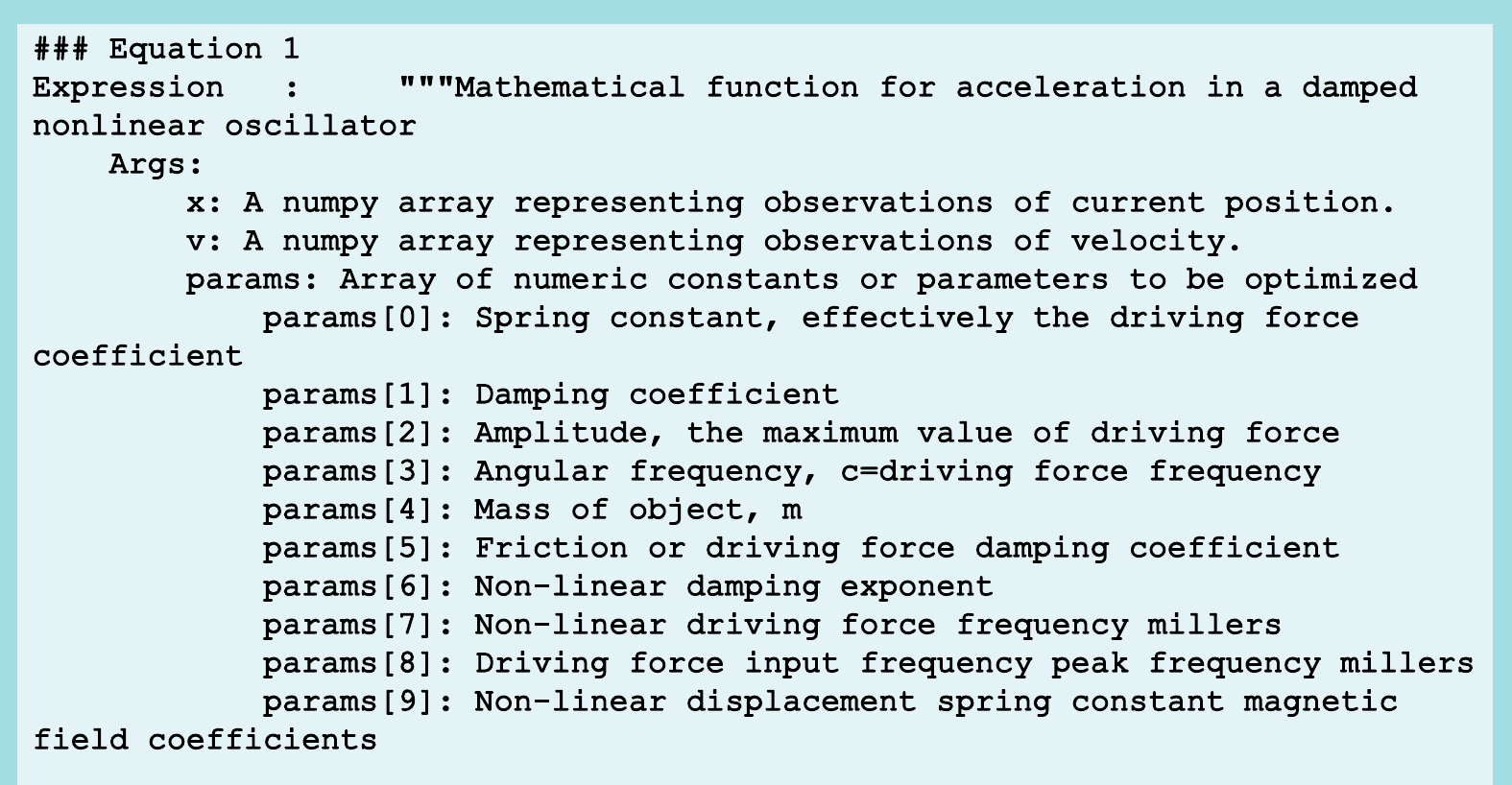}
\end{figure}

\begin{figure}[H]
    \centering
\includegraphics[width=0.8\linewidth]{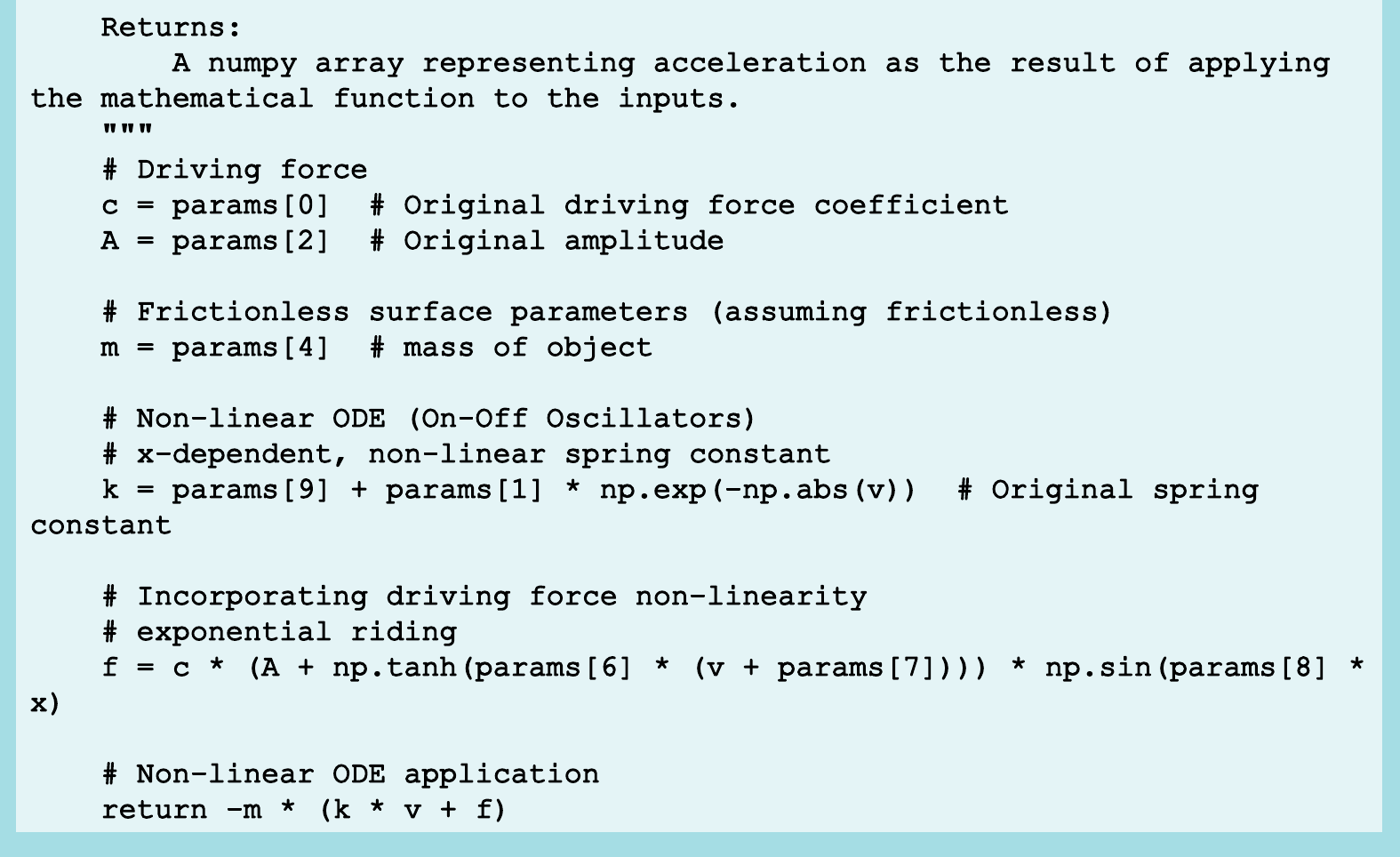}
    \caption{Example Prompt in Oscillator1 - For Meta Strategy Generator.}
    \label{fig:p8}
\end{figure}

\begin{figure}[H]
    \centering
\includegraphics[width=0.8\linewidth]{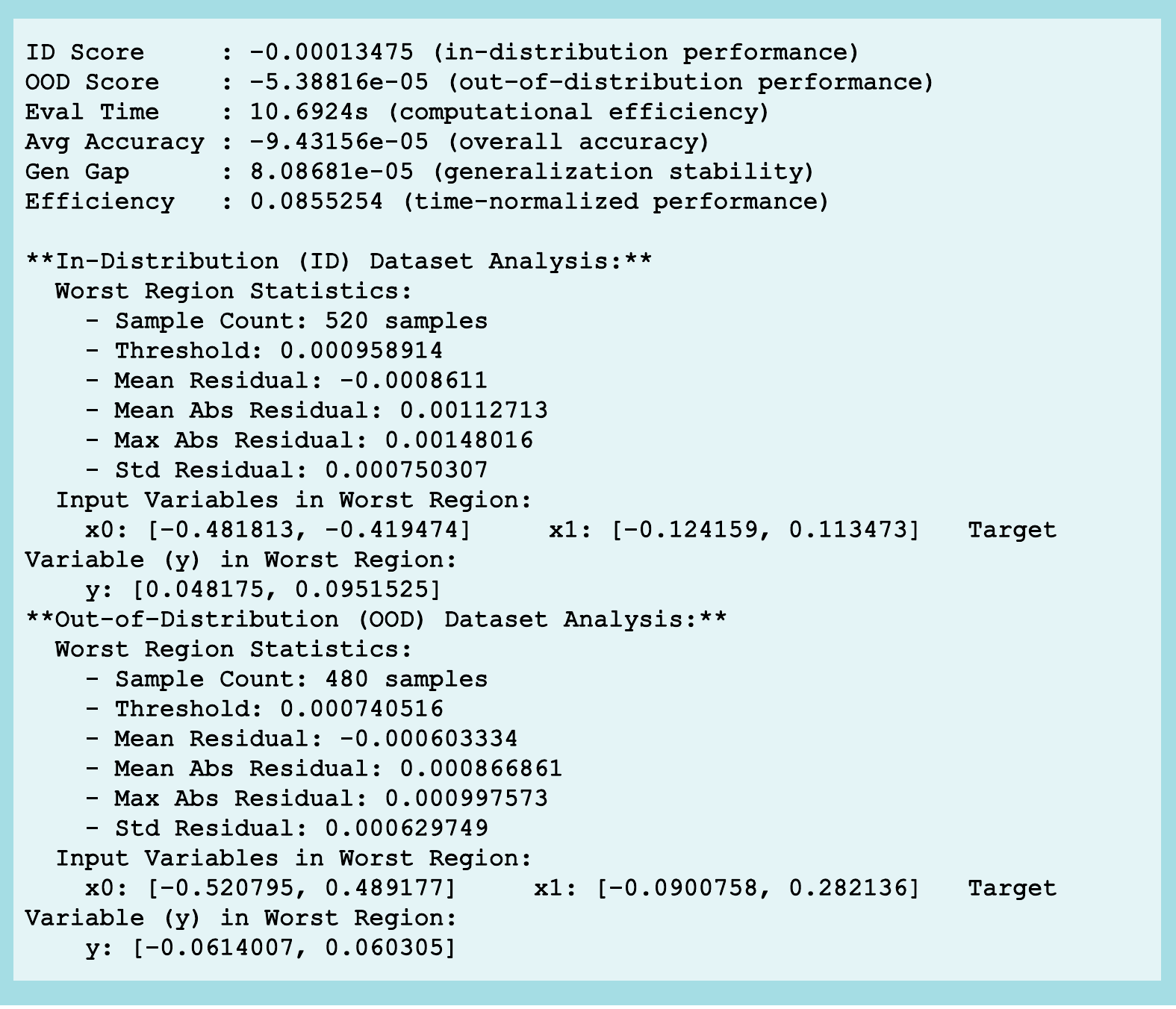}
    \caption{Example Prompt in Oscillator1 - For Meta Strategy Generator.}
    \label{fig:p9}
\end{figure}

\begin{figure}[H]
    \centering
\includegraphics[width=0.75\linewidth]{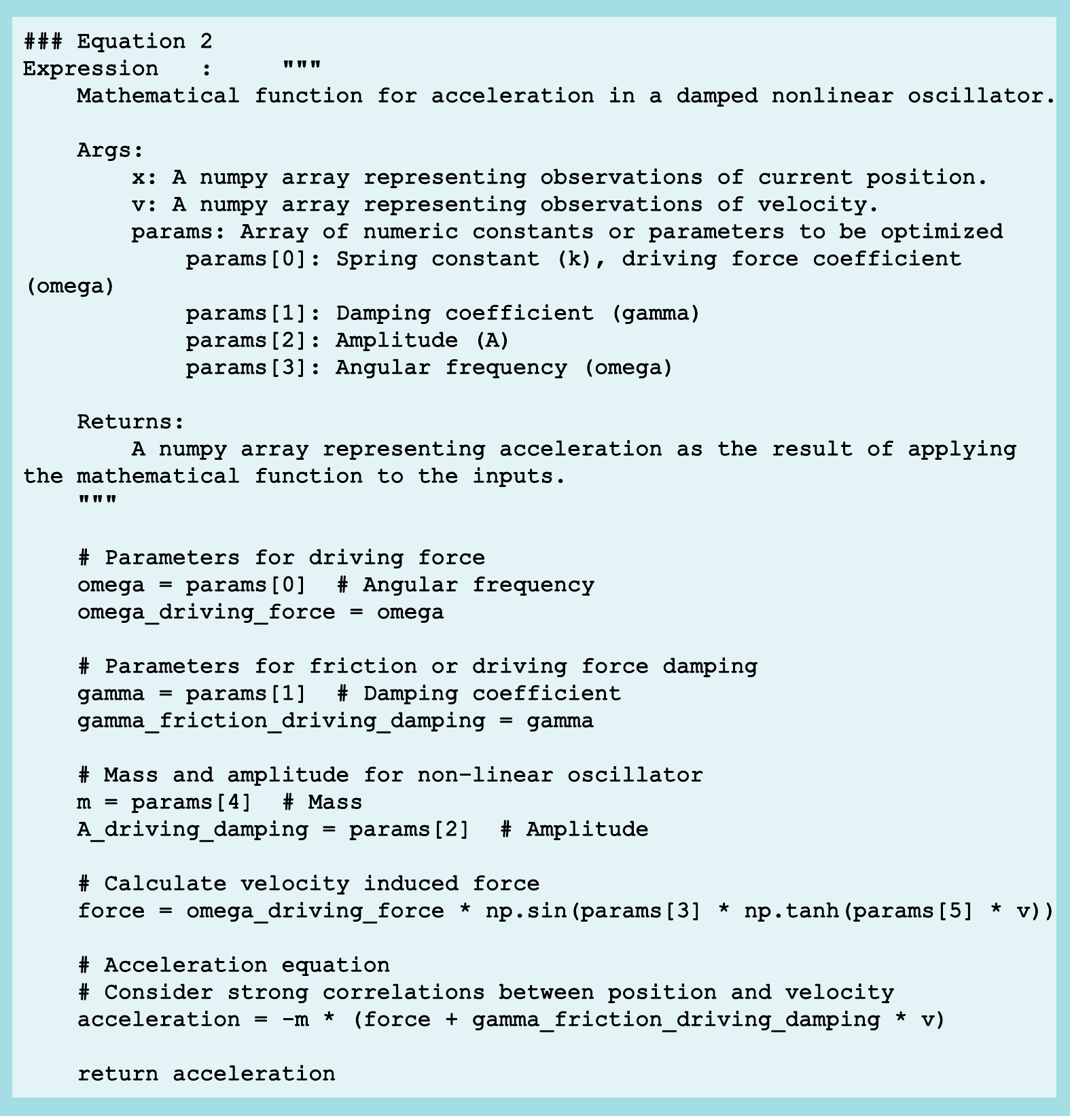}
    \caption{Example Prompt in Oscillator1 - For Meta Strategy Generator.}
    \label{fig:p10}
\end{figure}

\begin{figure}[H]
    \centering
\includegraphics[width=0.75\linewidth]{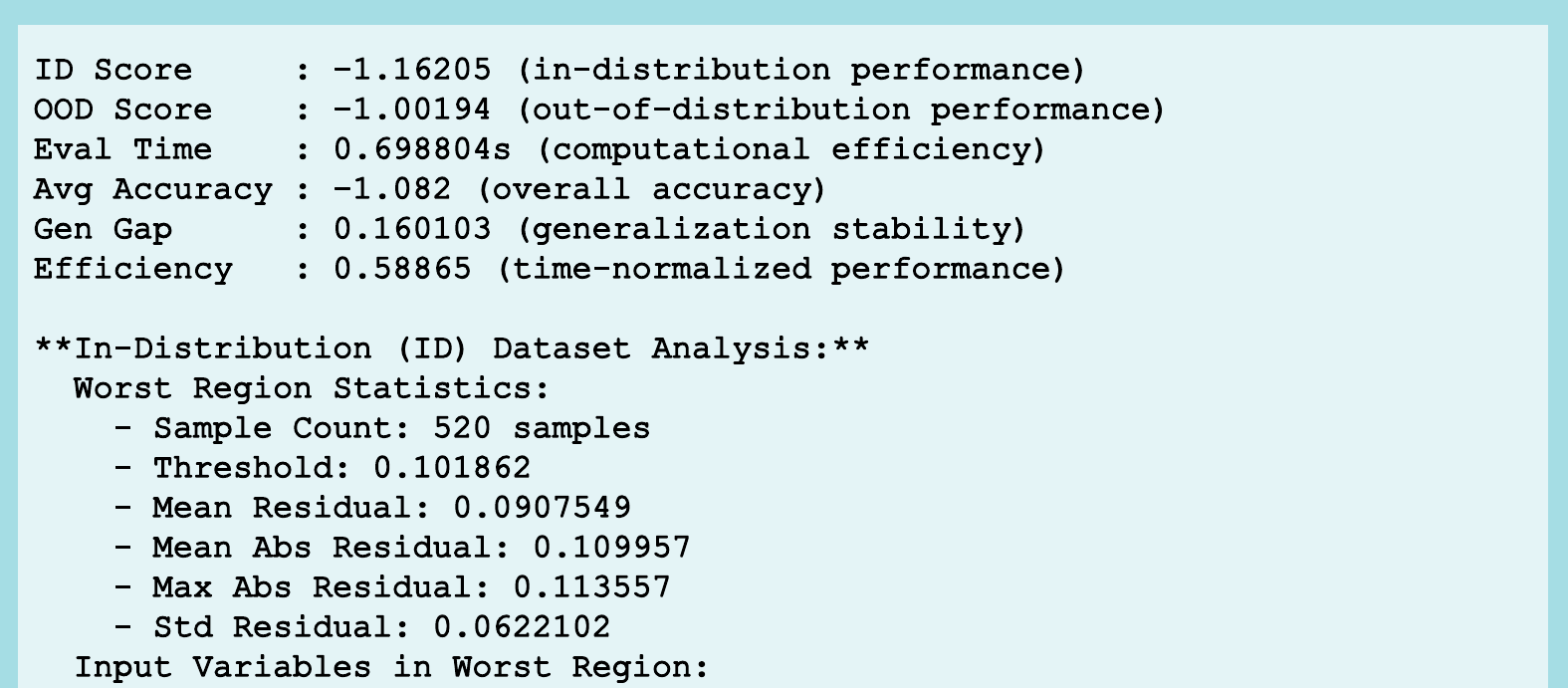}
\end{figure}

\begin{figure}[H]
    \centering
\includegraphics[width=0.75\linewidth]{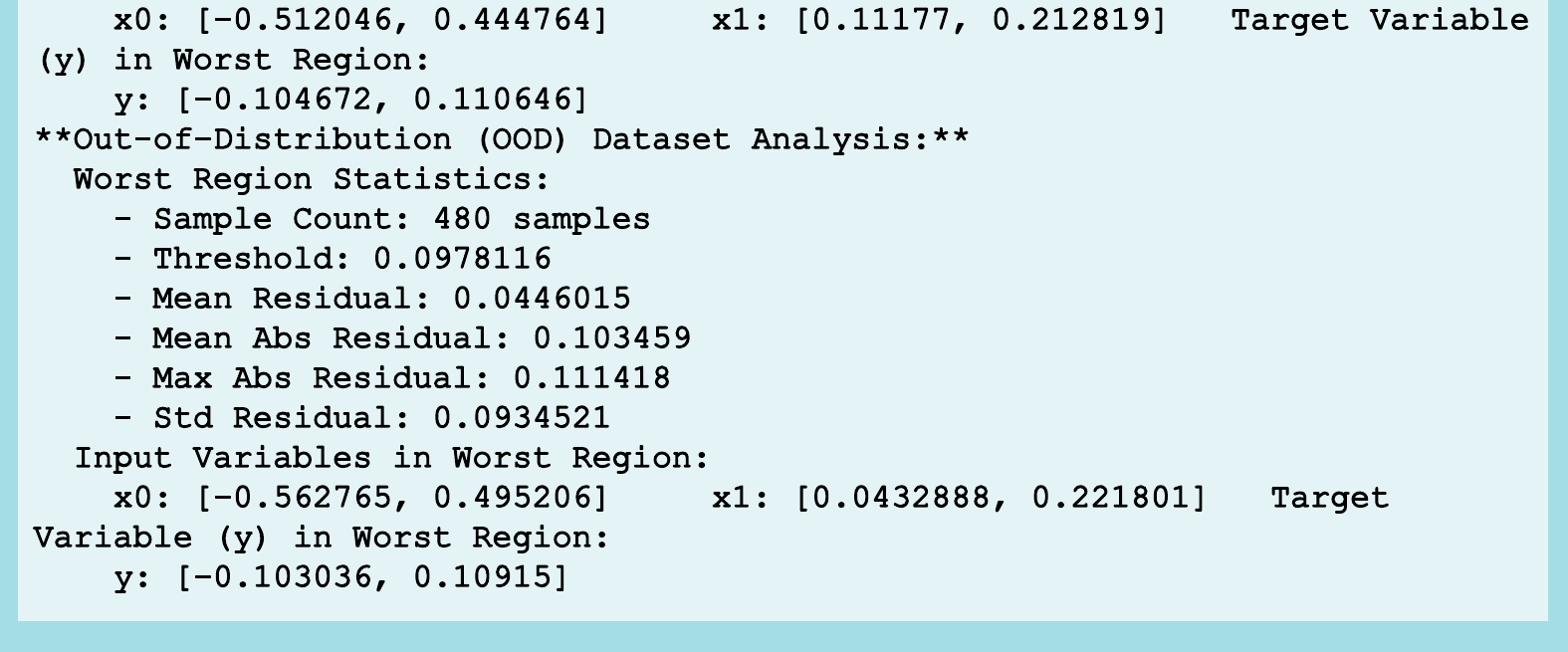}
    \caption{Example Prompt in Oscillator1 - For Meta Strategy Generator.}
    \label{fig:p11}
\end{figure}

\begin{figure}[H]
    \centering
\includegraphics[width=0.75\linewidth]{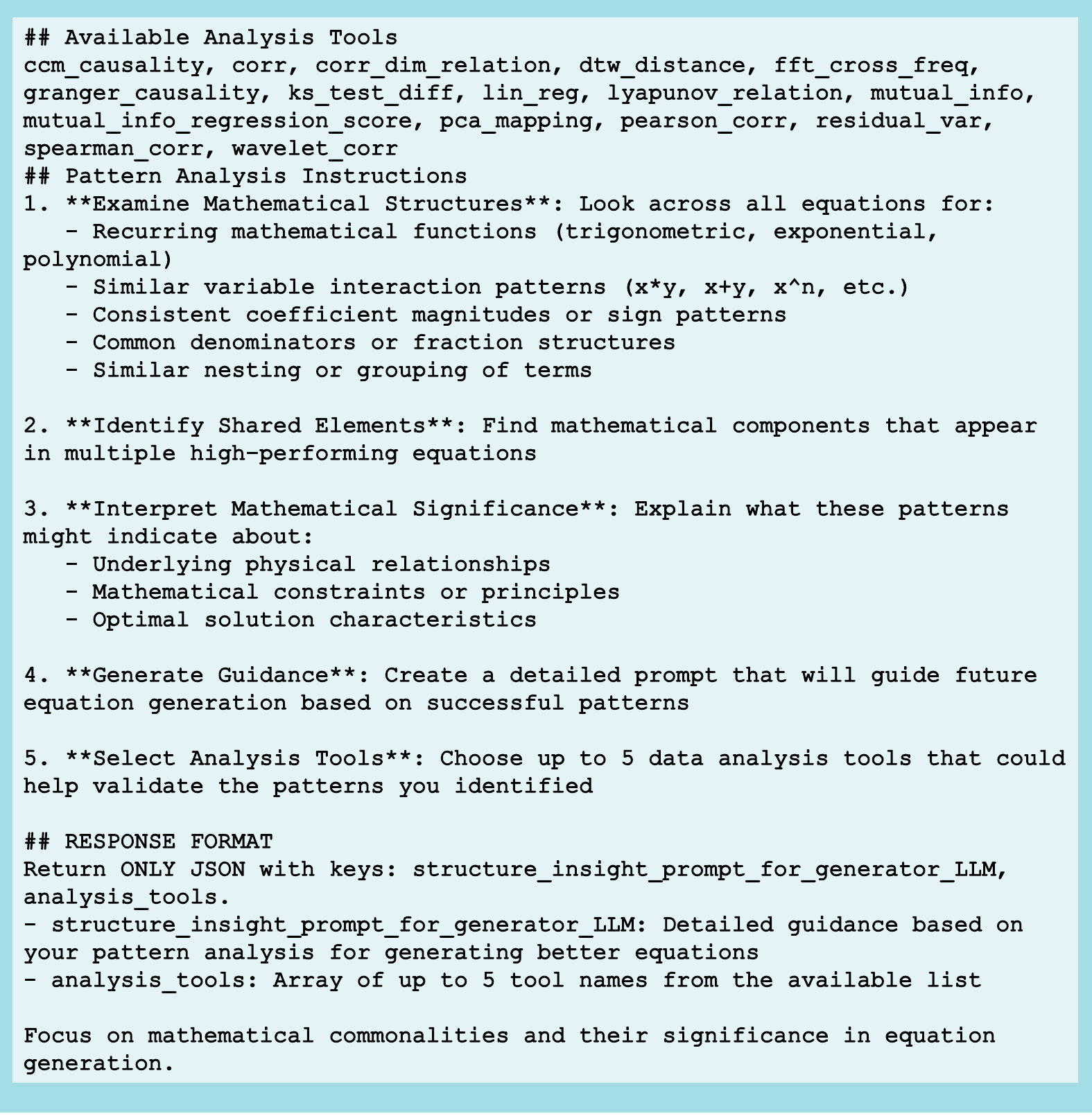}
    \caption{Example Prompt in Oscillator1 - For Meta Strategy Generator.}
    \label{fig:p12}
\end{figure}

\begin{figure}[H]
    \centering
\includegraphics[width=0.8\linewidth]{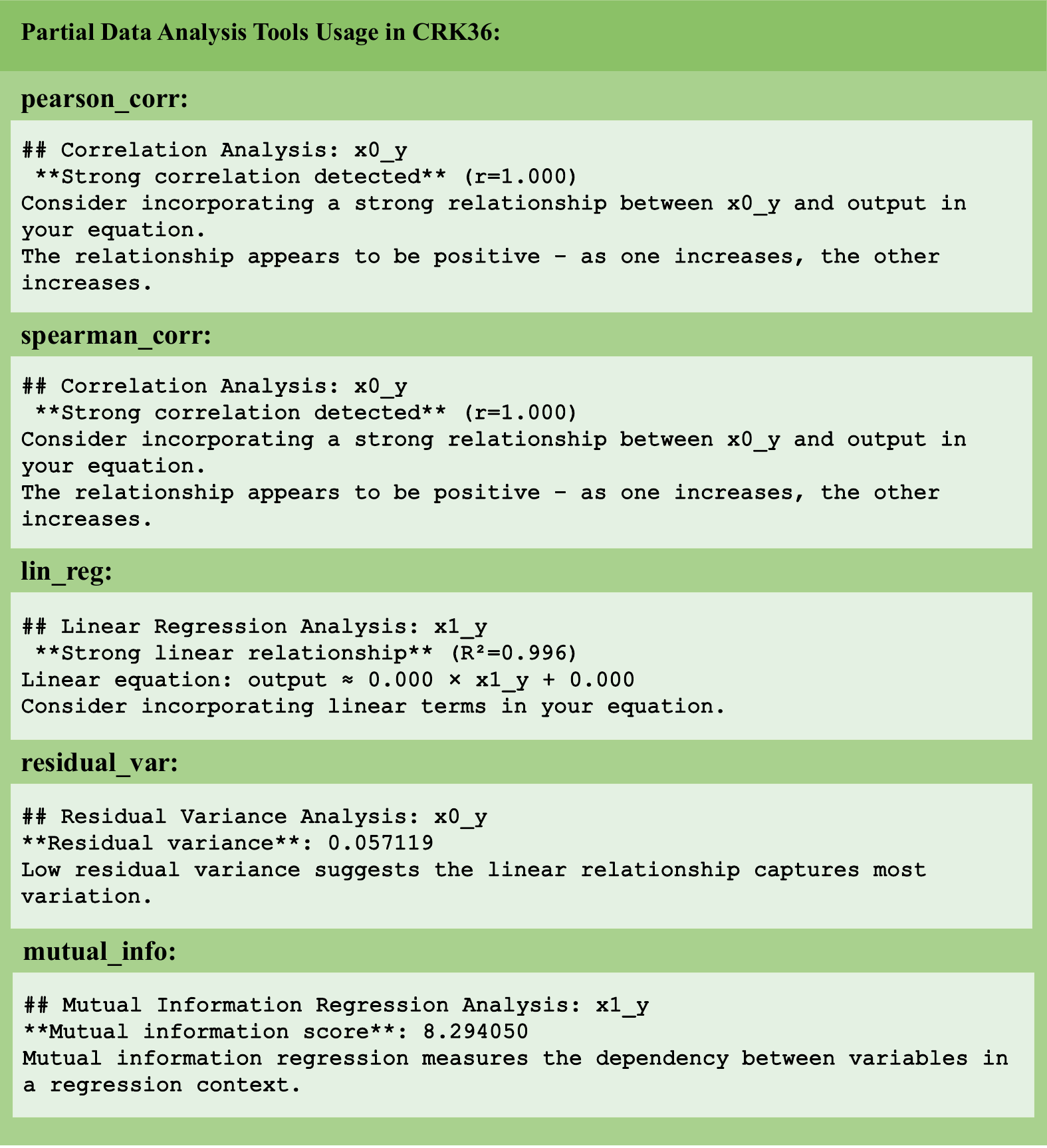}
    \caption{Examples of Data Analysis Tools Usage.}
    \label{fig:p13}
\end{figure}

\end{document}